\pdfoutput=1

\documentclass[11pt]{article}

\usepackage{acl}

\usepackage{times}
\usepackage{latexsym}

\usepackage[T1]{fontenc}

\usepackage[utf8]{inputenc}

\usepackage{microtype}

\usepackage{inconsolata}

\usepackage{booktabs} 
\usepackage{amsfonts}
\usepackage[T1]{fontenc}
\usepackage{adjustbox}
\usepackage{smiles}
\usepackage{bbm, dsfont}
\usepackage[normalem]{ulem}

\usepackage[utf8]{inputenc}
\usepackage[ruled,vlined]{algorithm2e}

\newcommand{\ours}{\textsc{Patron}}
\newcommand{\ptr}{\textsc{Ptr}}
\definecolor{chocolate}{rgb}{0.7, 0.3, 0.2}

\newcommand{\std}[1]{\scriptsize{ $\pm$ #1}}
\newcommand{\blue}[1]{\textcolor{blue}{#1}}
\newcommand{\bsl}[1]{\underline{\textbf{#1}}}

\definecolor{light-gray}{gray}{0.9}

\title{Cold-Start Data Selection for Few-shot Language Model Fine-tuning: \\ A Prompt-Based Uncertainty Propagation Approach}

\author{Yue Yu$^1$ \quad Rongzhi Zhang$^1$ \quad Ran Xu$^2$ \quad Jieyu Zhang$^3$ \\ \bf \quad Jiaming Shen$^4$  \quad Chao Zhang$^1$ \\  
$^1$ Georgia Institute of Technology  \quad  $^2$ Emory University  \\ $^3$ University of Washington \quad $^4$ Google \\
\texttt{\{yueyu, rongzhi.zhang, chaozhang\}@gatech.edu}, \ \texttt{\{ran.xu\}@emory.edu}, \\ \texttt{jieyuz2@cs.washington.edu}, \  \texttt{jmshen@google.com}
}

\begin{document}
\maketitle
\begin{abstract}

Large Language Models have exhibited remarkable few-shot performance; however, this performance can be sensitive to the selection of few-shot instances. We introduce \textit{\ours}, a prompt-based data selection method for fine-tuning pre-trained language models under cold-start scenarios, where no initial labeled data are available. In \textit{\ours}, we develop (1) a prompt-based uncertainty propagation approach for estimating the importance of data points, and (2) a partition-then-rewrite (PTR) strategy to enhance sample diversity when requesting annotations. Experiments conducted on six text classification datasets demonstrate that \textit{\ours} surpasses the most competitive cold-start data selection baselines by up to 6.9\%. Furthermore, with only 128 labels, \textit{\ours} achieves 91.0\% and 92.1\% of the fully supervised performance based on conventional fine-tuning and prompt-based learning, respectively. Our implementation of \textit{\ours} is available at \url{https://github.com/yueyu1030/Patron}. 

\end{abstract}

\section{Introduction}
\begin{figure}[t]
    \vspace{-2mm}
        \centering
        \hspace{-4mm}
        \subfigure[Fine-tuning]{
    \includegraphics[width=0.243\textwidth]{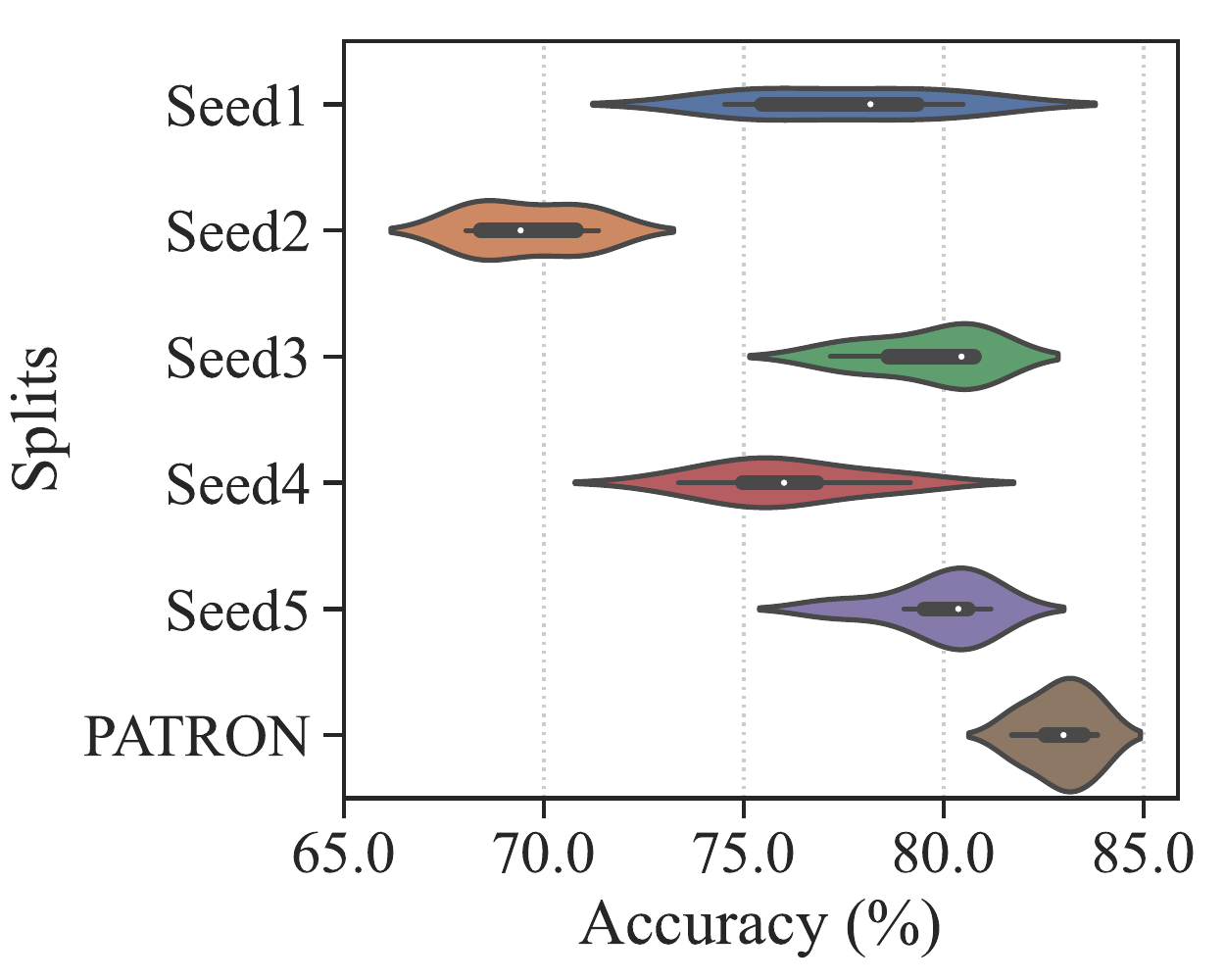}
            \label{fig:ablation_badge}
        }\hfill  \hspace{-4mm}
        \subfigure[Prompt-based Learning]{
            \includegraphics[width=0.243\textwidth]{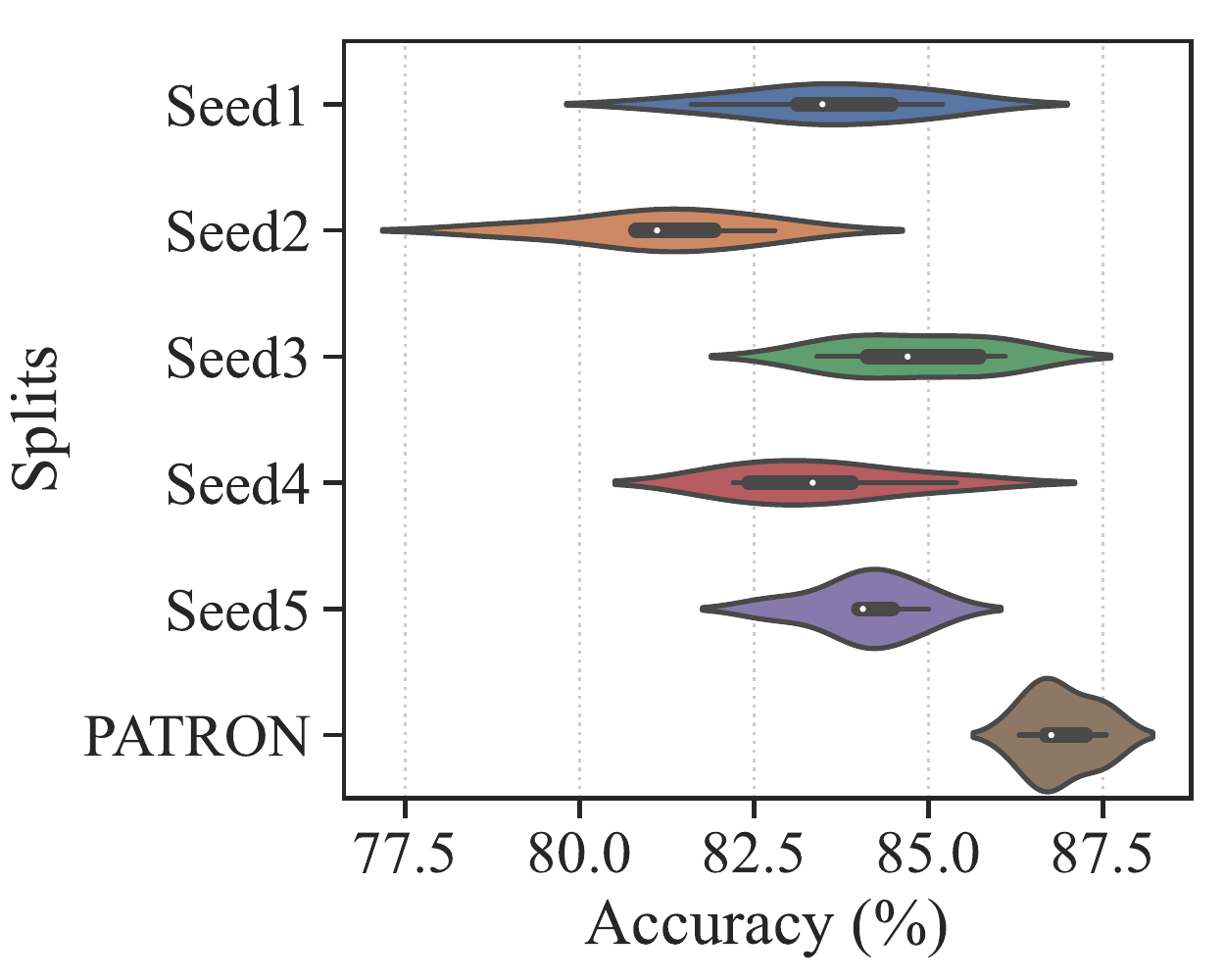}
            \label{fig:ablation_components}
        }
        \vspace{-1.5ex}
        \caption{
        The performance with large variances of vanilla fine-tuning and prompt-based learning on 5 random samplings, compared with better performance with low variances of {\ours} (our proposed selection strategy) on \textsl{AG News}~\cite{zhang2015character} with 32 labels. 
        } \label{fig:init}
        \vspace{-1ex}
\end{figure}

Pre-trained language models (PLMs)~\cite{bert,liu2019roberta,raffel2020exploring} have achieved competitive performance with limited labeled data~\cite{gao2021making,schick-schutze-2021-exploiting,schick-schutze-2021-just} for  many natural language processing (NLP) tasks. 
However, there still exists a non-negligible gap between the performance of few-shot and fully-supervised PLMs. 
Besides, when the task-specific data for fine-tuning is small,
the performance of PLMs can have high variance~\cite{bragg2021flex}. 
As illustrated in Figure~\ref{fig:init}, when fine-tuning RoBERTa-base~\cite{liu2019roberta} on different subsets of \textit{AG News} dataset with 32 labels, the performance on the test set varies up to 10\% for vanilla fine-tuning and 5\% for prompt-based learning~\cite{gao2021making}. 
Such large variations demonstrate the crucial need for strategical selection of training data to improve PLMs' performance  under low-data regimes.

To solicit training data intelligently,   
\emph{active learning} (AL)~\cite{settles2011theories} has been proposed to adaptively annotate unlabeled data~\cite{badge,ein-etal-2020-active,zhang-plank-2021-cartography-active,margatina2021active,margatina2022importance}.  
Despite their efficacy, most of these works assume there are hundreds, or even thousands of labels in the initial stage, and query similarly significant amounts of labeled data in each AL round. 
In practice, however, we usually do not have any startup labels to initialize the AL process, and the labeling budget can also be limited.  
This hinders the application of such techniques, as they often rely on a well-trained model with decent uncertainty~\cite{margatina2021active}, or gradient estimations~\cite{badge} to perform well. 

To facilitate training instance selection on such a challenging low-data regime, 
 \textit{cold-start} data selection (also known as cold-start AL~\cite{yuan-etal-2020-cold}) has been proposed, where we have only unlabeled data and \emph{zero} initial labels, and need to design acquisition functions to effectively query samples for PLM fine-tuning with \emph{few labels} only.  

However, cold-start data selection can be nontrivial for PLMs.  
Due to the absence of labeled data, the estimated uncertainty for unlabeled data from the PLM can be \emph{biased} over classes~\cite{zhao2021calibrate}. 
As a result, uncertainty-based approaches can underperform even the random selection strategy~\cite{hacohen2022active}.
Moreover, cold-start data selection requires greater care to ensure the sample diversity compared to the traditional AL, as fine-tuning PLMs on few redundant data will lead to poor generalization.
Existing approaches often first cluster the whole unlabeled data, and then greedily select samples from each cluster with the predefined heuristics~\cite{muller2022active}, 
which fails to control the distance between selected samples and thus cannot fully promote sample diversity.
In addition, under cold-start scenarios, it is critical to harness the knowledge from PLMs for sample selection. 
While there are several methods that leverage pre-trained  embeddings~\cite{hacohen2022active,chang-etal-2021-training} or  masked language modeling (MLM) loss~\citep{yuan-etal-2020-cold} to assist data selection, 
the mismatch between pre-training  and fine-tuning tasks hurts their efficacy.  
To sum up, it is still challenging to design effective methods for cold-start  data selection with PLMs. 

We propose {\ours}\footnote{\textbf{\underline{P}}rompt-based d\textbf{\underline{at}}a selection fo\textbf{\underline{r}} few-sh\textbf{\underline{o}}t PLM fi\textbf{\underline{n}}e-tuning.}, a prompt-based data-selection strategy tailored for PLMs, to address the above challenges.   
To estimate model  uncertainty without access to any labeled data under the cold-start setting,
{\ours} leverages prompts~\cite{gao2021making}, which convert the classification task into a cloze-style task with  customized templates and verbalizers, to generate the task-aware pseudo labels for unlabeled data by predicting the surface name for the \texttt{[MASK]} token.
In this way, we also bridge the gap between pre-training and downsteam tasks, and distill task-specific knowledge from PLMs to facilitate data selection. 

However, one important issue for such pseudo labels is they can be inaccurate and biased even after calibration~\cite{zhao2021calibrate}.  
To remedy this, we further propose \emph{uncertainty propagation} to first measure the correlation between samples based on kernel similarity in the embedding space, and then propagate their prediction uncertainty to their neighbors. 
Thus, a sample will have higher propagated uncertainty only when the predictive uncertainty for both itself and its neighbors are high, indicating the model is less certain for the local region around this sample.

To select a batch of diverse samples, we go beyond  
existing techniques and propose a two stage method named  \emph{partition-then-rewrite}~({\ptr}), which is initially proposed for combinatorial optimization~\cite{chen2019learning}, to dynamically adjust the selected sample within each cluster.
Concretely, we first use K-Means clustering to partition the unlabeled data and select one sample from each cluster to initialize our solution. 
We then build a neighbor graph based on $k$-nearest-neighbor (kNN) to encode the neighborhood relationships among selected data and explicitly control the distances between them.  
After that, we add an additional regularization term to prevent the selected sample in each cluster from being too close to samples in its neighbor clusters.
We iterate the above process for several rounds to gradually refine our solution and  promote  diversity in data selection.


It is worth noting that {\ours} can be naturally applied to various setups including vanilla fine-tuning, prompt-based learning, semi-supervised learning and standard multi-round AL to improve the data efficiency for PLM fine-tuning. 
We summarize the key contributions of our work as follows:
(1) a cold-start data selection paradigm {\ours} for lifting the label scarcity issue for few-shot PLM fine-tuning; 
(2) an prompt-based uncertainty propagation approach to query most informative samples;
(3) a partition-then-rewrite (\ptr) strategy to balance between diversity and informativeness of queried samples during AL and (4) experiments on six datasets demonstrating {\ours} improves the label efficiency over existing baselines by 3.2\%--6.9\% on average.

\section{Related Work}
\vspace{-1mm}

\noindent \textbf{Few-shot Language Model Fine-tuning.}  
Our method is closely relevant to other label-efficient learning paradigms in NLP such as cold-start fine-tuning~\cite{zhang2020revisiting,shnarch-etal-2022-cluster}, prompt-based learning\footnote{In this work, we refer prompt-based learning to \emph{Fixed-prompt PLM Tuning} mentioned in \cite{liu2021pre}.}~\cite{gao2021making,schick-schutze-2021-exploiting,schick-schutze-2021-just,min-etal-2022-noisy,zhang2022prompt,hu2022knowledgeable}, semi-supervised learning~\cite{du-etal-2021-self,wang2021list,xie2020unsupervised,xu2023neighborhood} and many others,  
but---orthogonal to our approach---they assume a small set of labeled data is given and design better training strategies. 
Instead, we aim to select most valuable instances from the unlabeled corpus, which is orthogonal to and can be combined with the above methods to further enhance label efficiency, as shown in Section~\ref{sec:main} and \ref{sec:multi_round_al}.

\noindent \textbf{Training Data Selection.} 
Designing better strategies to selectively annotate training data is not a new research topic. One of the most important line of research lies in active learning~\citep{zhang2020seqmix,schroder2022revisiting,yu-etal-2022-actune}, which improves the label efficiency of deep NLP models. 
However, most of them need a large amount of clean labels to first train the model before data selections~\cite{badge,zhang-plank-2021-cartography-active}. 
Differently, we aim to facilitate training data selection with minimal supervision, where no initial labeled data is given. 

The idea of such cold-start data selection has been applied for image classification~\cite{wang2021unsupervised,hacohen2022active} and speech processing~\cite{park2022unsupervised}, but has not been fully explored for the NLP domain. 
For this setting, \citet{chang-etal-2021-training} focus on data selection from the embedding space, but fail to leverage the task-specific knowledge from PLMs.
\citet{yuan-etal-2020-cold} uses the MLM loss as a proxy for uncertainty measurement, and 
\citet{liu2021makes,su2022selective} study few-shot sample selection for billion-scale language models~\cite{brown2020language}, but mainly focus on in-context learning. 
Different from the above methods, we aim to leverage prompts to facilitate sample selection, and design additional techniques (\emph{i.e.}, uncertainty propagation and \ptr) to boost the performance of few-shot PLM fine-tuning.  

\vspace{-0.5ex}
\section{Backgrounds}
\vspace{-0.5ex}

\subsection{Problem Formulation}
\vspace{-0.5mm}
We study cold-start data selection for text classification with $c$ classes formulated as follows: Given a pool of unlabeled samples $\cD_u = \{x_j\}_{j=1}^U$ and an empty training set $\cD_{l} = \emptyset$, 
we aim to fine-tune a pre-trained language model $\cM$ denoted as 
$f(\cdot;\theta)$ 
under limited labeling budget $|B|$ interactively:  
In each round, we use an acquisition function  $\cF(\cdot)$  to
query $b$ samples denoted as $\cQ$ from $\cD_u$.
Next, the acquired samples are labeled and moved from $\cD_u$ to $\cD_l$. Then we fine-tune the pre-trained language model $f(\cdot;\theta)$ with  $\mathcal{D}_l$ to maximize the performance on downstream classification tasks. The above steps can either be one-round~\cite{chang-etal-2021-training,hacohen2022active} ($b=|B|$ in this case) or repeated for multiple rounds~\cite{yuan-etal-2020-cold} ($b=|B|/|\text{Rounds}|$) until reaching the budget $|B|$.

\subsection{Prompt-based Learning for PLMs}
Prompting methods have been proposed to bridge the gap between the pre-training  and fine-tuning stage via applying the cloze-style tasks to fine-tune PLMs~\cite{schick-schutze-2021-exploiting,schick-schutze-2021-just}. 
Formally, there are two key components in prompts: a predefined template $\cT$, and a verbalizer $\cV$.  For each input sample $x$, it will be wrapped with the template which contains a piece of
natural language text together with a \texttt{[MASK]} token before feeding into the PLM $\cM$. 
Then, the verbalizer $\cV$ is used to map the task labels $y$ to individual words $\cV(y)$ in the vocabulary. 
Take the binary sentiment classification as an example, for input sentence $x$, a template $\mathcal{T}$ could be \underline{$\mathcal{T}(x)=[ x$. It was [\texttt{MASK}].$]$}, and the verbalizer for the positive and negative sentiment can be ``\texttt{good}'' and ``\texttt{terrible}”, respectively. 

With the template and verbalizer, we can calculate the probability distribution over the label set $\cY$ via Mask Language Modeling (MLM) as 

\begin{small}
\begin{equation}
\setlength{\abovedisplayskip}{0.2pt}
\setlength{\belowdisplayskip}{0.2pt}
\begin{aligned}
p\left(y \mid x\right) &=p\left(\texttt{[MASK]}=\cV(y) \mid \cT(x) \right) \\
&=\frac{\exp \left(\bm{w}_{\mathcal{V}(y)}^T  \bm{h}_{\texttt{[MASK]}}\right)}{\sum_{y^{\prime} \in \mathcal{Y}} \exp \left(\bm{w}^T_{\mathcal{V}\left(y^{\prime}\right)}  \bm{h}_{\texttt{[MASK]}}\right)}
\end{aligned}\label{eq:prompt}
\end{equation}
\end{small}

\noindent where $\bm{h}_{\texttt{[MASK]}}$ is the hidden embedding of the \texttt{[MASK]} token and $\bm{w}_{\mathcal{V}(y)}$ denotes the embedding of the label word $\mathcal{V}(y)$ from $\cM$. 
As these tokens' embeddings have been optimized during pre-training with the MLM objective, the use of prompts narrows the gap between pre-training and fine-tuning. 
In other words, prompts serve as a source of prior knowledge when adapting PLMs to new tasks.

\vspace{-1mm}
\section{Methodology}
\vspace{-1mm}
\begin{figure*}
    \centering
    \includegraphics[scale=0.35]{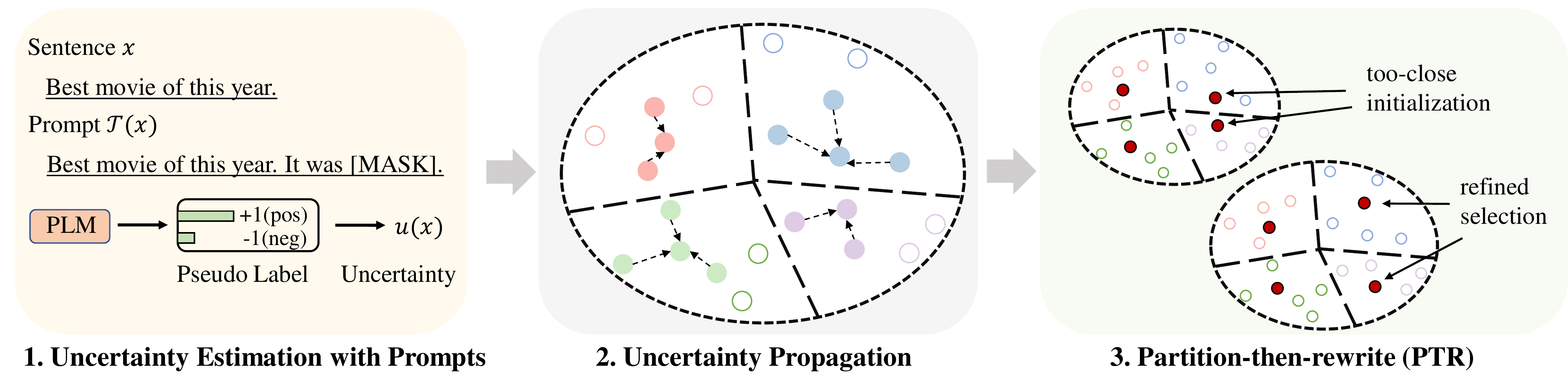}
    \vspace{-1ex}
    \caption{The illustration of the overall procedure for {\ours}.}
    \vspace{-1.5ex}
    \label{fig:my_label}
\end{figure*}
 In this section, we present our method, {\ours}, that exploits prompts for cold-start data selection. We first introduce how to leverage prompts for uncertainty estimation under cold-start scenarios. With the estimated uncertainty, we then propose two key designs, namely uncertainty propagation and partition-then-rewrite (\ptr) strategy to balance informativeness and  diversity for sample selection. 
\subsection{Uncertainty Estimation with Prompts}
\label{sec:uncertainty_estimation}
\vspace{-1mm}
We first describe how to estimate the uncertainty for unlabeled data to facilitate {\ours}.  Given the pre-trained language model (PLM) $\cM$ without labeled data, we leverage prompts to generate  pseudo labels\footnote{In this study, we use the manual prompts and verbalizers from existing works~\cite{hu2022knowledgeable,schick-schutze-2021-exploiting} due to their simplicity and competitive performance.} for uncertainty estimation.
According to Eq.~\ref{eq:prompt}, we are able to obtain the occurring probability for different label words on each sample $x$, based on the prediction of the \texttt{[MASK]} token. 

However, directly adopting this probability can be problematic as PLMs suffer from the mis-calibration issue~\cite{zhao2021calibrate,hu2022knowledgeable}, \ie, label words may have varying occurring frequencies, making some of them less likely to be predicted than the others. 
Thus, the prediction in Eq.~\ref{eq:prompt} and the estimated uncertainty can be biased.

Being aware of this, we adopt the method in \cite{hu2022knowledgeable} to calculate the \emph{contextualized prior} of the label words. 
We first construct a support set $\cS$ by choosing $k$ samples with highest $p(y_i|x)$ for each class $i$ as 
\begin{equation}
\setlength{\abovedisplayskip}{0.2pt}
\setlength{\belowdisplayskip}{0.2pt}
\cS = \bigcup\limits_{i \in \{1, 2, \ldots, c\}} \underset{x\in \cD_u}{\operatorname{Top-k}} \  p(y_i|x).
\label{eq:k_support}
\end{equation}
\noindent Then, the contextualized prior is approximated by
\begin{equation}
\setlength{\abovedisplayskip}{0.2pt}
\setlength{\belowdisplayskip}{0.2pt}
P(v) \approx \frac{1}{|{\cS}|} \sum_{x \in {\cS}} P_{\mathcal{M}}\left(\texttt{[MASK]} = v \mid x\right),
\end{equation}
which is used to calibrate the pseudo labels as
\begin{equation}
\setlength{\abovedisplayskip}{0.2pt}
\setlength{\belowdisplayskip}{0.2pt}
\hat{y_i} = \left({\frac{p(y_i|x)}{P(\cV(y_i))}}\right) /  \left({\sum_{j=1}^{C} \frac{p(y_j|x)}{P(\cV(y_j))}}\right).
\label{eq:calibration}
\end{equation}
After obtaining the pseudo labels, we use entropy~\cite{lewis1994entropy} as the measurement of uncertainty for each sample $x$ as 
\begin{equation}
\setlength{\abovedisplayskip}{0.2pt}
\setlength{\belowdisplayskip}{0.2pt}
u(x) = - \sum_{i=1}^{C} \hat{y_i} \log{\hat{y_i}}.
\label{eq:entropy}
\end{equation}

\subsection{Uncertainty Propagation for Data Utility Estimation}
\vspace{-1mm}
\label{sec:prop}

Although we have mitigated the bias for the prompt-based pseudo labels, such pseudo labels can still be inaccurate due to insufficient supervision under zero-shot settings.
Under this circumstance, directly using the uncertainty in Eq.~\ref{eq:entropy} for sample selection yields suboptimal results as it can be sensitive to outliers, which naturally have large model uncertainty but are less beneficial for model learning~\cite{karamcheti2021mind}. 

To remedy this issue, we leverage the kernel similarity in the embedding space to measure the correlation between data points and propagate the model uncertainty:  
for each data point $x$, we first calculate its embedding using SimCSE~\cite{gao-etal-2021-simcse}\footnote{Notably, we use the version of \url{princeton-nlp/unsup-simcse-roberta-base} as the encoder} as $\mathbf{z}=g(x;\theta)$, and calculate
$K$-nearest neighbors based on its Euclidean distance as 
$\cX_{\text{KNN}}(x)=\text{KNN}(x, \cD_u)$. Then, we choose the radial basis function (RBF)~\cite{scholkopf1997comparing} as the similarity metric for two data points $x_i$ and $x_j$, denoted as 
\begin{equation}
\setlength{\abovedisplayskip}{0.2pt}
\setlength{\belowdisplayskip}{0.2pt}
\kappa\left(x_{i}, x_{j}\right)=\exp \left(-\rho\left\|\mathbf{z}_{i}-\mathbf{z}_{j}\right\|_2^{2}\right),
\label{eq:prop}
\end{equation}
where $\mathbf{z}_{i}$ is the embedding of $x_i$ from the SimCSE, and $\rho$ is a hyper-parameter controlling the weight of propagation. 
Formally, the propagated uncertainty for $x$ can be represented as 
\begin{equation}
\setlength{\abovedisplayskip}{0.2pt}
\setlength{\belowdisplayskip}{0.2pt}
\hat{u}_{\text{prop}}(x) = u(x) + \frac{\sum_{x_i\in\cX_{\text{KNN}}(x)}\kappa(x, x_i)\cdot u(x_i)}{|\cX_{\text{KNN}}(x)|}.
\label{eq:prop_final}
\end{equation}
We highlight that only when the sample has higher uncertainty for both \emph{itself} and \emph{its neighbors} will result in higher propagated uncertainty, indicating the PLMs are uncertain about the surrounding regions around  the sample.
In this case, actively annotating such samples will be most beneficial for PLMs.

\noindent \subsection{Partition-then-rewrite ({\ptr}) for Diversity-Promoting Data Selection}  \vspace{-1mm}
Instead of querying one sample at a time, modern AL methods usually query a batch of samples to improve the query efficiency. In this case, querying samples without considering their correlations will lead to a redundant query set with limited performance gain~\cite{ein-etal-2020-active}.
We now present our {\ptr} strategy for diversity-promoting sample selection underpinned by
the estimated uncertainty.  

\noindent \textbf{Initialization of Selection with Partition.} As previous studies~\cite{aharoni2020unsupervised} revealed that PLMs implicitly learn sentence representations clustered by topics even without fine-tuning, we first employ K-Means clustering to partition the unlabeled pool $\cD_u$ into different clusters based on their embeddings and enforce the coverage over different topics of selected samples.
We follow existing works~\cite{chang-etal-2021-training,hacohen2022active} to set the number of clusters equal to $b$, denoted as $\cC_i$ ($1 \leq i \leq b$)\footnote{Here we use one-round AL for better  illustration. We provide the details for adapting {\ptr} to the multi-round AL setting in Appendix~\ref{app:pisa_multi}.}. 
We then use a greedy method to select one sample $q_i$ from $\cC_i$ to initialize the selected data pool $\cQ$ as 
\begin{equation}
\setlength{\abovedisplayskip}{0.5pt}
\setlength{\belowdisplayskip}{0.5pt}
q_i = \argmax_{x_j\in\cC_i}  \left(\hat{u}_{\text{prop}}(x_j) - \beta \left\|\mathbf{z}_j-\mathbf{\bar{z}}_i\right\|_2^{2}\right),
\label{eq:init}
\end{equation}
where $\mathbf{\bar{z}}_i=\frac{1}{|\cC_i|}{\sum_{x_j\in\cC_i}\mathbf{z}_j}$ is the centroid for the cluster $i$ and $\beta$ is a hyperparameter. 
In this way, we aim to select data points with higher propagated uncertainty while not being faraway with most of the data points to balance between the uncertainty and diversity.

\noindent \textbf{Sample Refinement with Rewriting.}   
Although the previous steps attempt to select the most informative samples within each cluster, they fail to model the relations among samples in different clusters. 
As a result, samples can still be very close to other selected samples in adjacent clusters,  leading to the limited overall diversity.
To tackle this issue, we build an additional KNN graph to retrieve the nearest query samples from other clusters as 
\begin{equation}
\setlength{\abovedisplayskip}{0.2pt}
\setlength{\belowdisplayskip}{0.2pt}
\cX_{\text{c-KNN},i} =  \text{KNN}(q_i, \cQ).
\label{eq:cknn}
\end{equation}
Note that we use c-KNN  to denote the cluster-level KNN to differentiate from the sample-level KNN in section \ref{sec:prop}. 
To update the selected pool $\cQ$, for cluster $i$, we add an additional regularization term to Eq.~\ref{eq:init} to prevent samples in adjacency clusters from being overly close:
\begin{equation}
\setlength{\abovedisplayskip}{1pt}
\setlength{\belowdisplayskip}{1pt}
\begin{aligned}
\tilde{q}_i = & \argmax_{x_j\in\cC_i} \  ( \hat{u}_{\text{prop}}(x_j) - \beta \left\|\mathbf{z}_j-\mathbf{\bar{z}}_i\right\|_2^{2} \\
& - \gamma \sum_{q_k \in \cX_{\text{c-knn}, i}}\left[m - \left\|\mathbf{z}_j-\mathbf{z}_k\right\|_2^{2}\right]_{+} ),
\end{aligned}
\label{eq:init_updated}
\end{equation}
where $\gamma$ is the weight for the penalty term, $m=0.5$ is the pre-defined margin,  $[\cdot]_{+}=\max(\cdot, 0)$ is the gating function.
To interpret the regularization term, we argue that when the distance between the selected samples in adjacency clusters is smaller than $m$, the regularization will be greater than 0 to discourage them from being selected together. 

We run the above rewriting steps several times until convergence to obtain the final set $\cQ=\{\tilde{q}_i\}_{i=1}^{b}$, which usually takes 2-3  iterations\footnote{The efficiency analysis of {\ours} is in Appendix~\ref{app:time}.}. The overall procedure of {\ours} is in Algorithm~\ref{alg:main}. 

\begin{algorithm}[t]
	\begin{small}
	\KwIn{Unlabeled samples $\mathcal{X}_u$; Pre-trained LM $\cM=f(\cdot; \theta)$, number of acquired samples $B$, the number of iterations $T$ ($T$=2 in this work).}
	// \textbf{Step 1}: \textit{Uncertainty Propagation for  Utility Estimation.} \\
    \textbf{1a}. Calculate uncertainty for samples $x\in\cX_u$ with prompts based on Eq.~\eqref{eq:entropy}. \\
    \textbf{1b}. Estimate uncertainty $\hat{u}_{\text{prop}}$ with Eq.~\eqref{eq:prop} and \eqref{eq:prop_final}. \\ 
	// \textbf{Step 2}: \textit{Predict-then-propagate (\ptr) for Diversity Promoting Selection.} \\
	\textbf{2a}. Run K-Means on $\cX_u$ with $k$=$B$ until convergence. \\
    \textbf{2b}. Select initial sample set $\cQ^{(0)}$ based on Eq.~\eqref{eq:init}. \\
	\For{$t = 1, 2, \cdots, T$}{
		{
			\textbf{2c}. Building the additional KNN graph to obtain $\cX_{\text{c-KNN}}$ with Eq.~\eqref{eq:cknn}. \\
			\textbf{2d}. Update $\cQ^{(t)}$ by optimizing the selected sample within each cluster  $\tilde{q}$ with Eq.~\eqref{eq:init_updated}. \\
		}
	}
	\KwOut{The final selected labeled data $\cQ^{(T)}$.}
	\end{small}
	\caption{Procedures of the data selection strategy \ours. }
	\label{alg:main}
\end{algorithm}

\section{Experiments}
\label{sec:exp}
\vspace{-1mm}
\subsection{Experiment Setup}
\label{exp_setup}
\vspace{-0.5mm}
\noindent\textbf{Datasets.}
 Following prior works~\cite{yuan-etal-2020-cold,schroder2022revisiting}, we use six text classification tasks in our experiments: \emph{IMDB}~\cite{imdb}, \emph{Yelp-full}~\cite{yelp-full}, \emph{AG News}~\cite{zhang2015character}, 
\emph{Yahoo! Answers}~\cite{zhang2015character}, 
\emph{DBPedia}~\cite{lehmann2015dbpedia}, and \emph{TREC}~\cite{li2002learning}. 
All the datasets are in English, and their
detailed statistics are shown in Table \ref{tab:dataset}. 
Besides, we use 3 additional datasets to evaluate the out-of-distribution (OOD) performance, the details are shown in Appendix~\ref{app:ood}.  The template and verbalizer for prompts are shown in table~\ref{app:dataset}.

\noindent\textbf{Evaluation Setup.} 
Following \cite{chang-etal-2021-training,chen2022zeroround},
we focus on \emph{one-round} data selection in our main experiments because  it can more faithfully reflect the performance of different strategies. 
We choose the labeling budget $|B|$ from \{32, 64, 128\} to simulate the few-shot scenario and align with existing works~\cite{muller2022active,shnarch-etal-2022-cluster}.
We also apply {\ours} for standard multi-round AL (see Section~\ref{sec:multi_round_al}).

\begin{table}[t]
    \centering
    \renewcommand\arraystretch{0.85}
\begin{adjustbox}{max width=0.485\textwidth}
\begin{tabular}{c|c|c|c|c}
\toprule
    \bf Dataset & \bf Label Type & \bf \#Class $c$ & \bf \#Unlabeled $|\cD_u|$ & \bf \#Test   \\\midrule
    IMDB & Sentiment  & 2 & 25k  & 25k \\
    Yelp-full & Sentiment & 5 & 40k  & 10k \\
    AG News & News Topic  & 4 & 120k  & 7.6k \\
    Yahoo! Ans. & QA Topic  & 10 & 300k  & 60k \\
    DBPedia & Wikipedia Topic & 14 & 420k & 70k \\ 
    TREC & Question & 6 & 5.4k  & 0.5k \\
    \bottomrule
\end{tabular}
\end{adjustbox}
 \caption{Dataset Statistics. For DBPedia and Yahoo! Answers, we randomly sample 30k sample from each class due to the limited computational resource.}
\vspace{-1ex}
\label{tab:dataset}
\end{table}

\begin{table}[t]
    \centering
    \renewcommand\arraystretch{0.85}
\begin{adjustbox}{max width=0.495\textwidth}
\begin{tabular}{c|c|c|c|c|c|c}
\toprule
    \bf IMDB & \bf Yelp-full & \bf AG News & \bf Yahoo! & \bf DBPedia & \bf TREC  & \bf Mean \\\midrule
    94.1 & 66.4 & 94.0  & 77.6 & 99.3 & 97.2 &  88.1\\
    \bottomrule
\end{tabular}
\end{adjustbox}
 \caption{Fully supervised performance on six datasets.}
\vspace{-2ex}
\label{tab:full_supervised}
\end{table}

\begin{table*}[!t]
\centering 
\renewcommand\arraystretch{0.8}
\fontsize{7.5}{9.5}\selectfont \setlength{\tabcolsep}{0.5em}
\resizebox{0.95\linewidth}{!}{%
\begin{tabular}{c|cc|c|cc|cc|ccc|c}
\toprule
\bf    Task     & $c$ & $|B|$  & \bf  Random & \bf Uncertainty & \bf CAL  & \bf BERT-KM &\bf  Coreset & \bf Margin-KM & \bf ALPS & \bf TPC & \bf {\ours~(Ours)} \\ \midrule
\multirow{3}{*}{IMDB} & \multirow{3}{*}{2} 
&    32  & 80.2\std{2.5} & 81.9\std{2.7} & 77.8\std{2.4} & 79.2\std{1.6} & 74.5\std{2.9} &  76.7\std{3.5} & 82.2\std{3.0} & \underline{82.8\std{2.2}} & \bf 85.5\std{1.5}$^{**}$ \\
& &  64   & 82.6\std{1.4} & {84.7\std{1.5}} & 81.2\std{3.4} & 84.9\std{1.5} & 82.8\std{2.5} &  84.0\std{2.0} & \underline{86.1\std{0.9}} & 84.0\std{0.9} & \bf 87.3\std{1.0}$^{**}$ \\
& & 128  & 86.6\std{1.7} & 87.1\std{0.7} & 87.9\std{0.9} & \underline{88.5\std{1.6}} & 87.8\std{0.8} &  88.2\std{1.0} & 87.5\std{0.8} & 88.1\std{1.4} & \bf 89.6\std{0.4}  $^{*}$ \\
\midrule
\multirow{3}{*}{Yelp-F} & \multirow{3}{*}{5} 
& 32     & 30.2\std{4.5} & 32.7\std{1.0} & \underline{36.6\std{1.6}} & 35.2\std{1.0} & 32.9\std{2.8}  & 32.7\std{0.4} & \bf{36.8\std{1.8}} & 32.6\std{1.5} & 35.9\std{1.6}  \ \ \  \\
& & 64   & \underline{42.5\std{1.7}} & 36.8\std{2.1} & 41.2\std{0.2} & 39.3\std{1.0} & 39.9\std{3.4} & 39.8\std{1.2}& 40.3\std{2.6}  & 39.7\std{1.8} & \bf 44.4\std{1.1} $^{*}$ \ \\
& & 128 & 47.7\std{2.1} & 41.3\std{1.9} & 45.7\std{1.3} & 46.4\std{1.3} & \underline{49.4\std{1.6}} & 47.1\std{1.2}  & 45.1\std{1.0} & 46.8\std{1.6} & \bf 51.2\std{0.8}$^{**}$ \\
\midrule

\multirow{3}{*}{AG News} & \multirow{3}{*}{4} 
& 32 & 73.7\std{4.6} & 	73.7\std{3.0} & 69.4\std{4.5}   & 79.1\std{2.7} & 78.6\std{1.6} &	 75.1\std{1.8} & 78.4\std{2.3}  & \underline{80.7\std{1.8}} & \bf 83.2\std{0.9}$^{**}$\\
& & 64  & 80.0\std{2.5} &  80.0\std{2.2} & 78.5\std{3.7} & {82.4\std{2.0}} & 82.0\std{1.5} &	 81.1\std{2.2} & 82.6\std{2.5}  & \underline{83.0\std{2.4}} & \bf 85.3\std{0.7}$^{**}$ \\
& & 128 & 84.5\std{1.7} & 82.5\std{0.8} & 81.3\std{0.9} & {85.6\std{0.8}} & 
85.2\std{0.6} &	 \underline{85.7\std{0.3}} & 84.3\std{1.7}  & \underline{85.7\std{0.3}} & \bf 87.0\std{0.6}$^{**}$ \\
\midrule
\multirow{3}{*}{Yahoo! Ans.} & \multirow{3}{*}{10} 
& 32     & 43.5\std{4.0} & 23.0\std{1.6} & 26.6\std{2.5} & 46.8\std{2.1} & 22.0\std{2.3} & 34.0\std{2.5} & \underline{47.7\std{2.3}} & 36.9\std{1.8} & \bf 56.8\std{1.0}$^{**}$ \\
& & 64   & 53.1\std{3.1} & 37.6\std{2.0} & 30.0\std{1.7} & 52.9\std{1.6} & 45.7\std{3.7} & 44.4\std{2.8} & \underline{55.3\std{1.8}} & 54.0\std{1.6} & \bf 61.9\std{0.7}$^{**}$ \\
& & 128  & 60.2\std{1.5} & 41.8\std{1.9} & 41.1\std{0.9} & \underline{61.3\std{1.0}} & 56.9\std{2.5} & 52.1\std{1.2} & 60.8\std{1.9} & 58.2\std{1.5} & \bf 65.1\std{0.6}$^{**}$ \\
\midrule
\multirow{3}{*}{DBPedia} & \multirow{3}{*}{14} 
& 32     & 67.1\std{3.2} & 18.9\std{2.4} & 14.6\std{1.5} & \underline{83.3\std{1.0}} & 64.0\std{2.8} & 55.1\std{2.2} & 77.5\std{4.0} & 78.2\std{1.8} & \bf 85.3\std{0.9}$^{**}$\\
& & 64  & 86.2\std{2.4} & 37.5\std{3.0} & 20.7\std{2.0} & \underline{92.7\std{0.9}} & 85.2\std{0.8} & 78.0\std{4.1} & 89.7\std{1.1} & 88.5\std{0.7} & \bf 93.6\std{0.4}$^{**}$\\
& & 128  & 95.0\std{1.5} & 47.5\std{2.3} & 26.8\std{1.4} & \underline{96.5\std{0.5}} & 89.4\std{1.5} & 85.6\std{1.9} & 95.7\std{0.4} & 95.7\std{0.6} & \bf 97.0\std{0.2} $^{*}$ \\
\midrule
\multirow{3}{*}{TREC} & \multirow{3}{*}{6} 
& 32 & 49.0\std{3.5} & 46.6\std{1.4} & 23.8\std{3.0} & 60.3\std{1.5} & 47.1\std{3.6} & 49.5\std{1.2} & \underline{60.5\std{3.7}} & 42.0\std{4.4} & \bf 64.0\std{1.2}$^{**}$ \\
 & & 64  & 69.1\std{3.4} & 59.8\std{4.2} & 28.8\std{3.6} & \underline{77.3\std{2.0}} & 75.7\std{3.0} & 63.0\std{2.5} & 73.0\std{2.0} & 72.6\std{2.1} & \bf 78.6\std{1.6}$^{**}$ \\
 & & 128  & 85.6\std{2.8} & 75.0\std{1.8} & 50.5\std{1.9} & \underline{87.7\std{1.5}} & 87.6\std{3.0} & 80.5\std{2.8} & 87.3\std{3.6} & 83.0\std{3.8} & \bf 91.1\std{0.8}$^{**}$ \\
\midrule
\multirow{3}{*}{Average} &  
& 32 &    57.2  & 46.1 & 41.5 & \underline{64.0} & 53.2 & 53.8 & 63.9 & 58.9 & \textbf{68.4}~\scriptsize{\blue{($\uparrow$ 4.4)}}  \\
& & 64   & 68.9  & 56.1 & 46.8 & \underline{71.6} & 68.5 & 65.1 & 71.2 & 70.3 &  \textbf{75.2}~\scriptsize{\blue{($\uparrow$ 3.6)}} \\
 & & 128  &  76.6 & 62.5 & 55.6 & \underline{77.6} & 76.1 & 73.2 & 76.8 & 76.3 & \textbf{80.2}~\scriptsize{\blue{($\uparrow$ 2.5)}} \\
\bottomrule
\end{tabular}
}
\caption{Main results of cold-start data selection on six datasets with \textbf{10 runs}. Here $c$ means the number of classes and $|B|$ is the number of acquired samples. We use accuracy as the metric, and the higher value indicates better performance. Since TREC is an imbalanced dataset, we report the F1 score in \textbf{Appendix~\ref{app:f1}}.
\textbf{Bold} and \underline{underline} indicate the best and second best results for each setting respectively. We use $\pm$ to indicate standard deviation and use */** to indicate statistical significant results according to student’s t-test at level 0.05/0.01. (Same as belows.)}
\label{tab:main}
\end{table*}

\begin{table*}[!t]
\centering 
\renewcommand\arraystretch{0.8}
\fontsize{7.5}{9.5}\selectfont \setlength{\tabcolsep}{0.5em}
\resizebox{0.95\linewidth}{!}{%
\begin{tabular}{c|cc|c|cc|cc|ccc|c}
\toprule
\bf    Task     & $c$& $|B|$ & \bf  Random & \bf Uncertainty & \bf CAL  & \bf BERT-KM &\bf  Coreset & \bf Margin-KM & \bf ALPS & \bf TPC & \bf {\ours~(Ours)} \\ \midrule

\multirow{3}{*}{IMDB} &\multirow{3}{*}{2} & 32 & 81.8\std2.5 &	82.4\std1.7&	79.6\std1.6&	81.7\std1.3&	85.5\std1.1&	\underline{86.0\std1.2}	&83.5\std2.6	&84.5\std0.9&\bf	86.5\std0.9 \\
& & 64 &  85.6\std{1.3}&	86.0\std{1.4}&	81.1\std{1.9}&	84.2\std{0.9}&	\underline{87.8\std{0.6}}	&87.6\std{0.7}	&84.4\std{1.6}&	85.8\std{1.2}&	\bf 88.8\std{0.8}$^{*}$ \\
&  & 128 & 87.7\std{0.4}&	88.4\std{0.5}&	83.0\std{2.0}&	88.5\std{0.8}&	88.9\std{0.5}&	\underline{89.1\std{0.4}}&	88.9\std{0.3}&	88.0\std{0.5}&	\bf 89.3\std{0.3}\\ \midrule

\multirow{3}{*}{Yelp-F} & \multirow{3}{*}{5} & 32 & 48.9\std1.3	& 46.6\std0.9	& 47.9\std0.6 &	45.5\std1.0 &	46.0\std1.5 &	47.5\std1.1 &	47.0\std1.0	& \underline{49.8\std0.5}	&\bf  50.5\std0.8$^{*}$ \\
&  & 64   & 51.0\std{0.8}	&49.9\std{0.8}&	49.4\std{1.1}&	51.9\std{0.5}&	48.8\std{1.2}&	52.6\std{0.6}&	\underline{52.8\std{0.5}} & 52.3\std{0.7} & \bf  53.6\std{0.3}$^{**}$\\
&  & 128   & 51.3\std{0.9}&	50.8\std{0.6}&	48.7\std{1.6}&	51.5\std{1.4}	&53.7\std{1.1}&	\underline{54.2\std{0.7}}&	51.7\std{0.5}&	51.0\std{0.7}&\bf	55.6\std{0.6}$^{**}$\\ \midrule

\multirow{3}{*}{AG News} & \multirow{3}{*}{4}  & 32  & 83.1\std1.2	& 82.8\std2.0	& 81.4\std1.0	& 84.9\std0.9	& 85.1\std1.5	& 84.6\std1.7 &	{84.2\std0.8} & 	\underline{85.6\std1.0} &\bf	86.8\std0.3$^{**}$ \\ 
& &  64 & 84.5\std{1.3} & 84.3\std{1.4} & 82.6\std{1.2}	&\underline{86.5\std{0.8}} &	86.4\std{1.3}	&85.9\std{0.7}&	86.2\std{0.5}&	85.6\std{0.5}&	\bf 87.4\std{0.6}$^{*}$ \\
& & 128 & 84.9\std{0.5} &	83.1\std{0.8}&	83.0\std{0.9}&	\underline{87.6\std{0.4}}&	87.5\std{0.3}&	87.1\std{0.4}&	87.5\std{0.4}&	87.0\std{0.6}&	\bf 87.8\std{0.3} \\ \midrule 

\multirow{3}{*}{Yahoo! Ans.} & \multirow{3}{*}{10}  &  32 & 58.5\std4.0 &	55.0\std3.0 &	54.0\std1.5 &	61.4\std1.8 &	55.3\std2.1 &	57.8\std2.6 &\underline{61.9\std0.9} &	57.0\std1.6 &\bf	63.2\std1.2$^{*}$ \\
& & 64 & 62.2\std{1.0}&	60.4\std{0.7}&	58.6\std{1.3}&	62.8\std{0.7}&	59.5\std{0.7}&	58.8\std{1.2}&	\underline{63.3\std{0.8}}&	60.8\std{0.7}&	\bf 66.2\std{0.3}$^{**}$ \\
& & 128 & 64.7\std{1.3}	& 63.0\std{1.2}	& 60.1\std{1.8}& 	65.4\std{1.2}& 	62.7\std{1.0}& 	65.4\std{0.7}& 	65.9\std{0.7}& 	\underline{66.2\std{0.6}}& \bf	67.6\std{0.5}$^{**}$\\
\midrule 

\multirow{3}{*}{DBPedia} & \multirow{3}{*}{14}  &  32 & 89.1\std3.0 &	77.9\std2.8 &	58.9\std1.3 &	94.1\std1.4 &	92.0\std0.6 &	90.6\std0.7 &	91.2\std2.8 &\underline{94.3\std0.5} &\bf 95.4\std0.4$^{**}$\\
& & 64 & 95.5\std{1.2}&	86.3\std{1.0}&	63.5\std{1.7}&	95.8\std{0.7}&	\underline{96.1\std{0.4}} &	95.5\std{0.6}&	95.4\std{0.7}&	95.6\std{0.5}&	\bf 96.9\std{0.2}$^{**}$ \\ 
& & 128&96.0\std{0.6}&	87.8\std{0.7}&	78.1\std{2.0}	&\underline{97.2\std{0.2}}&	96.4\std{0.5}&	96.6\std{0.4}&	96.8\std{0.3}&	97.0\std{0.3}&	\bf 97.4\std{0.1}$^{*}$ \\ \midrule

\multirow{3}{*}{TREC} & \multirow{3}{*}{6}  & 32 &  69.4\std2.8 &	66.4\std3.5 &	41.6\std2.5 &	68.1\std2.3 &	61.0\std4.6 &	64.8\std2.7 &	\underline{72.1\std2.3} &	59.5\std3.3 &\bf 76.1\std1.1$^{**}$ \\
& & 64 & 75.4\std{1.4}&	68.0\std{2.3}	&49.8\std{1.5}&	78.8\std{2.0} &	78.6\std{1.3}&	74.2\std{1.4}&	\underline{80.6\std{0.9}} &	77.8\std{1.5}&	\bf 81.9\std{1.3}$^{*}$ \\
& & 128 & 85.0\std{2.1}&	78.8\std{2.0}&	67.2\std{2.7}	&85.6\std{1.8}&	84.2\std{2.4}&	78.0\std{1.9}&		\underline{86.5\std{2.0}}&	80.6\std{1.4}	& \bf 88.9\std{1.0}$^{**}$ \\
\midrule
\multirow{3}{*}{Average} &  & 32& 
71.9 &	68.6&	60.4	&72.6&	71.0	&71.9	& \underline{73.2} &	71.8& \textbf{76.5}~\scriptsize{\blue{($\uparrow$ 3.3)}} \\
 &  & 64& 
75.7 &	72.5&	64.2	&76.7&	69.5	&75.7	& \underline{77.1} &	76.3& \textbf{79.5}~\scriptsize{\blue{($\uparrow$ 2.4)}} \\
 &   & 128 & 
78.2 &	75.3&	70.0	&79.3&	78.9	&78.4	& \underline{79.5} &	78.3& \textbf{81.1}~\scriptsize{\blue{($\uparrow$ 1.6)}} \\
\bottomrule
\end{tabular}
}
\caption{Experimental result for prompt-based learning~\cite{gao2021making} on six datasets with \textbf{10 runs}. 
}
\label{tab:lmbff_result_all}
\end{table*}

\noindent\textbf{Baselines.}  We include both traditional AL baselines (Random, Uncertainty, CAL, Coreset) as well as recently-proposed cold-start AL baselines (BERT-KM, Margin-KM, ALPS, TPC). 
%

\noindent $\diamond$  \bsl{Random}: It acquires annotations randomly with the annotation budget.   \\
\noindent $\diamond$ \bsl{Uncertainty}~\cite{schroder2022revisiting}: It acquires annotations on samples with the highest uncertainty in Eq.~\ref{eq:entropy} after calibration. We use \textsc{Entropy}~\cite{lewis1994entropy} as the uncertainty estimate\footnote{We do not find clear performance gains for other metrics.}. \\
\noindent $\diamond$ \bsl{CAL}~\cite{margatina2021active}: It selects samples based on the KL divergence between the prediction of itself and that of its neighbors. \\
\noindent $\diamond$ \bsl{Coreset}~\cite{coreset}: It selects samples such that the largest distance between a data point and its nearest center is minimized. \\
\noindent $\diamond$ \bsl{BERT-KM}~\cite{chang-etal-2021-training}: It first uses K-Means to cluster pre-trained embeddings and then selects one example from each cluster that is closest to the center of the cluster. \\
\noindent $\diamond$ \bsl{Margin-KM}~\cite{muller2022active}: It utilizes K-Means clustering to group pre-trained embeddings, followed by the selection of samples with the minimum margin between the two most likely probabilities from each cluster. Unlike {\ptr}, Margin-KM does not explicitly control the distance among selected samples, nor does it employ uncertainty propagation to improve data utility estimation. \\
\noindent $\diamond$ \bsl{ALPS}~\cite{yuan-etal-2020-cold}: It uses the masked language model (MLM) loss of BERT to generate surprisal embeddings to query samples. \\
\noindent $\diamond$ \bsl{TPC}~\cite{hacohen2022active}: It is the most recent method for CSAL, which first 
calculates the density for each data point, and then selects those with the highest density from each cluster.

\noindent\textbf{Implementation Details.}
We choose RoBERTa-base~\cite{liu2019roberta} from the Hugging Face codebase~\cite{huggingface} as the backbone for all the compared methods.
For prompt-based learning, we use OpenPrompt~\cite{openprompt} as the codebase. 
More details are in Appendix \ref{appendix:impl}.

\noindent\textbf{Hyperparameters.}
The hyperparameters settings are given in Appendix~\ref{appendix:hyper} and~\ref{appendix:ours}.

%

\subsection{Main Results}
\label{sec:main}
\vspace{-1mm}
Table~\ref{tab:main} reports the performance of {\ours} and the
baselines under different budgets $|B|$ on \textbf{10} runs. We have the following observations:

\noindent $\diamond$ Compared with the baselines, {\ours} achieves the best overall performance on the six datasets, with an average gain of 3.2\%-6.9\% over the strongest baselines under different annotation budgets. 
Moreover, with 128 labels only (<0.5\% of total labeled data), {\ours} obtains 91.0\% of the fully supervised performance on the average of six datasets. 
It is also worth noting that {\ours} also lead to more \emph{stable} results --- it achieves lower standard deviations when compared with baselines on 14 of 18 cases.   
These results justify the  benefits of {\ours} in cold-start setting. 

\noindent $\diamond$ We observe the performance gains are more significant for datasets with larger number of classes (\emph{e.g.} TREC, Yahoo!).
This observation further strengthens the superiority of {\ours} in resolving label scarcity issue brought by cold-start setting, because for datasets with more classes, each class would have less labeled data given a fixed budget.

\noindent $\diamond$ Similar to the findings in \cite{hacohen2022active}, pure uncertainty-based AL methods (\emph{e.g.} CAL) do not perform well under cold-start settings. The reason is two-fold: (1) these methods focus on choosing `hard samples' without considering the sample diversity, leading to imbalanced label distribution for acquired samples; (2) they do not consider the potential bias in uncertainty estimation.  

\noindent $\diamond$ Diversity-based methods (\emph{e.g.} BERT-KM, ALPS)   generally achieve performance gains over the uncertainty-based strategy. Intriguingly, we find that directly using K-Means performs better than other hybrid approaches with more complicated operations (\emph{e.g.} TPC, ALPS) for data selection, especially for datasets with larger number of classes. 
This is because these complex methods often ignore the diversity for selected samples in adjacent clusters and therefore underperform {\ours}.

\subsection{Adapting {\ours} to Other Settings}
\label{sec:multi_round_al}
\begin{figure*}[t]
    \vspace{-2mm}
    \hspace{-2mm}
    \hfill
    \hspace{-2mm}
        \centering
         \subfigure[AG News]{
            \includegraphics[width=0.32\textwidth]{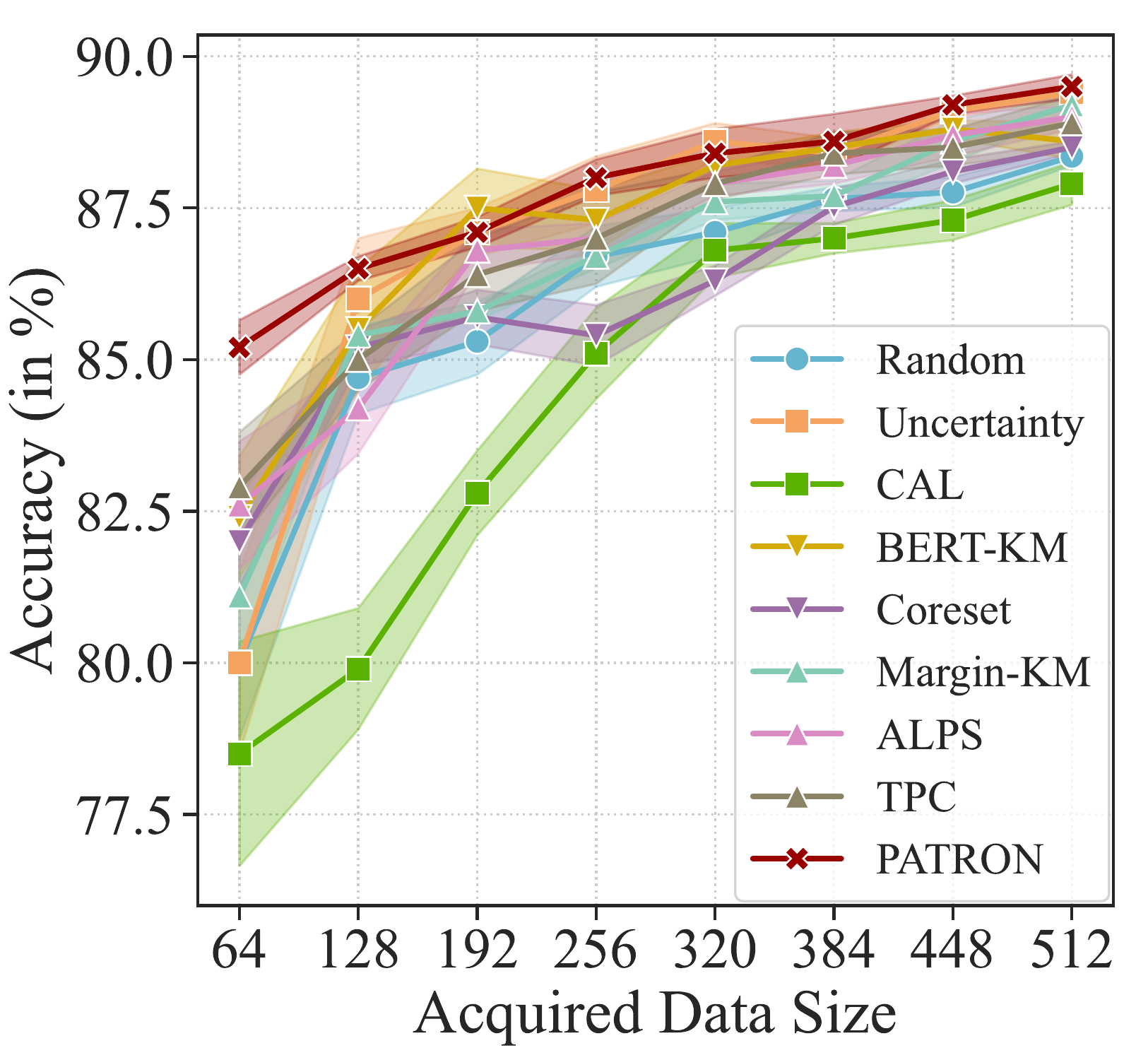}
            \label{fig:agnews}
            }\hspace{-2mm}
        \subfigure[IMDB]{
    \includegraphics[width=0.32\textwidth]{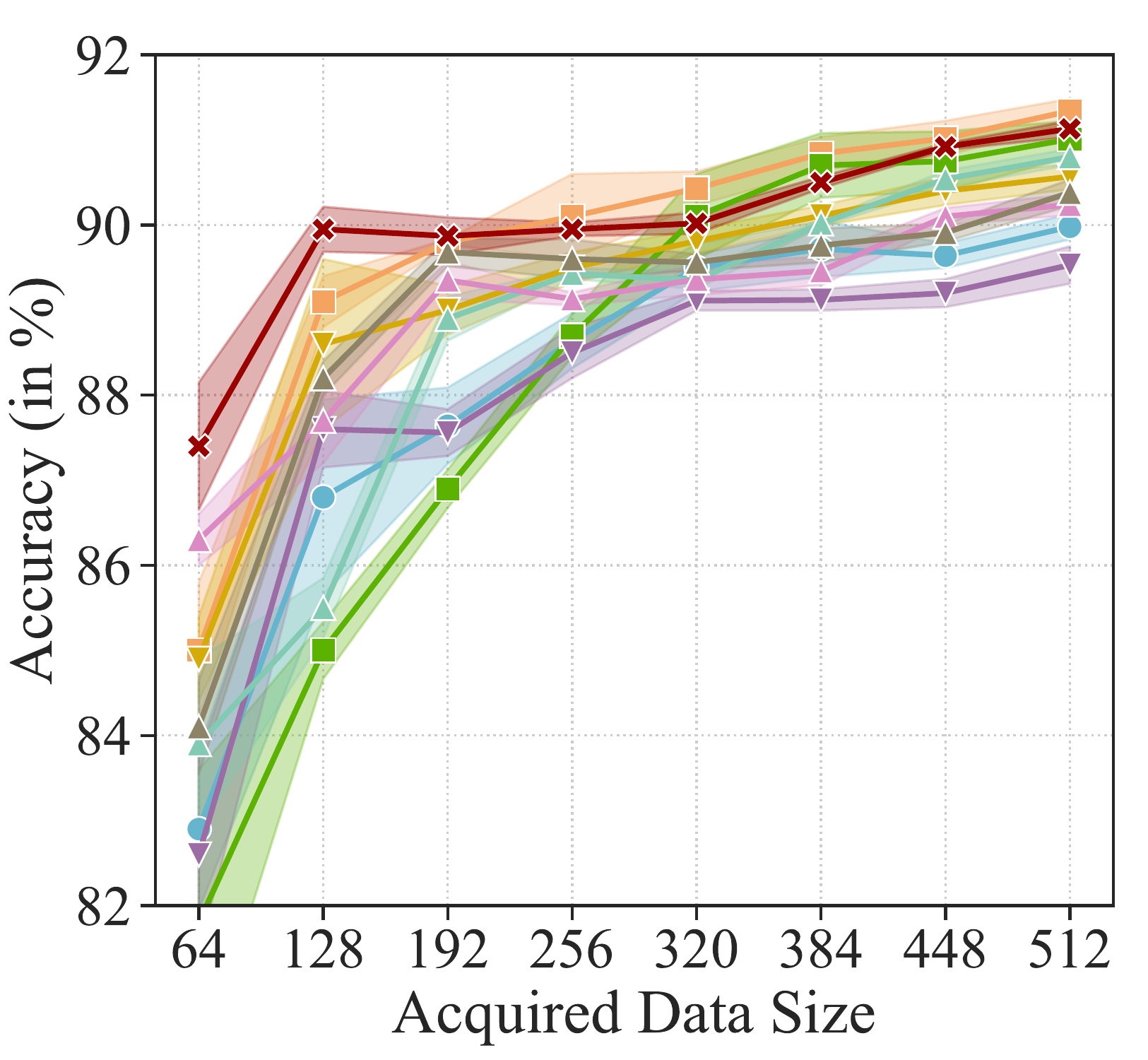}
            \label{fig:imdb}
        }\hspace{-2mm}
        \subfigure[Yelp-full]{
            \includegraphics[width=0.32\textwidth]{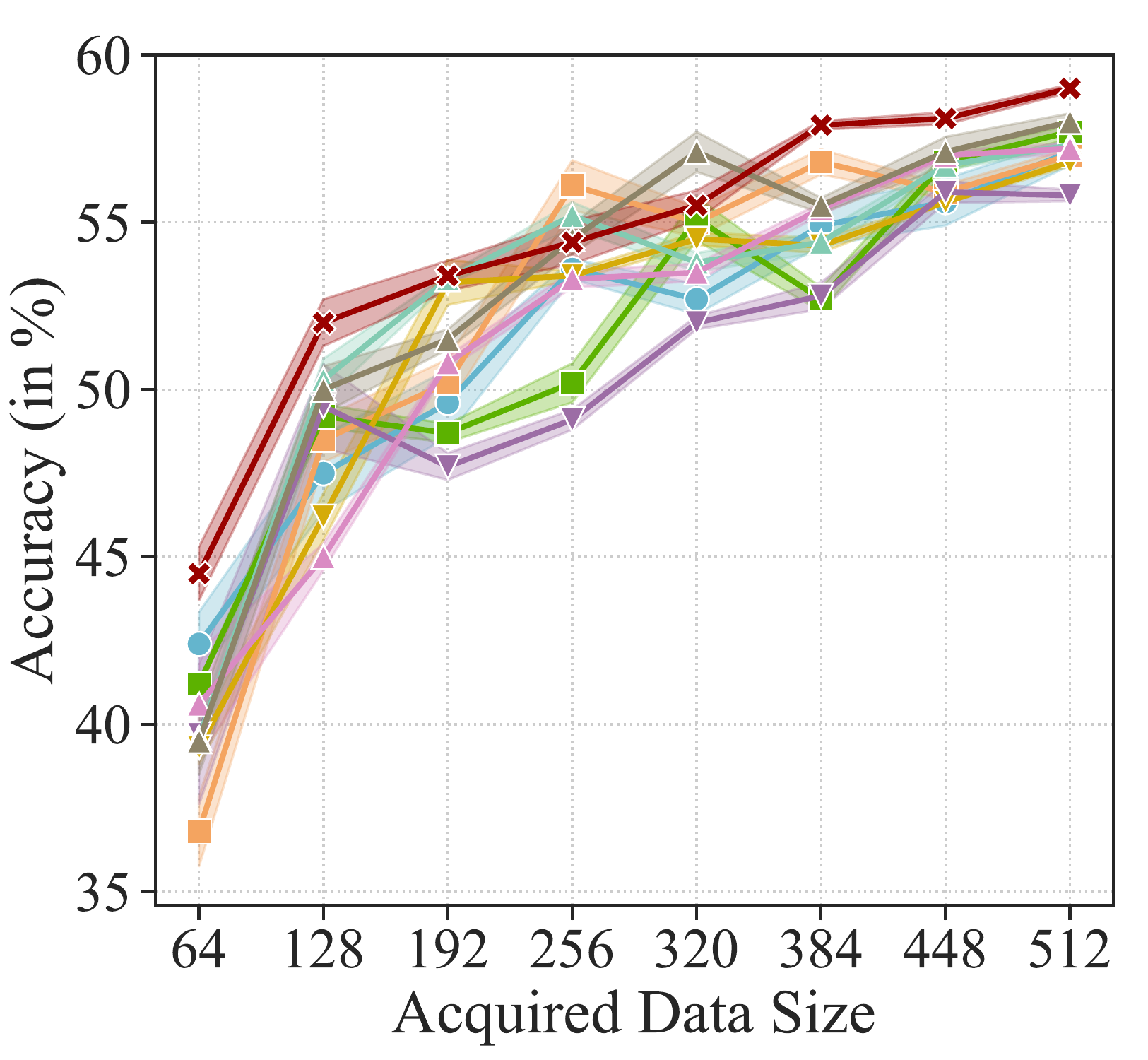}
            \label{fig:yelp_full}
        }
        \vspace{-1ex}
         \subfigure[TREC]{
    \includegraphics[width=0.32\textwidth]{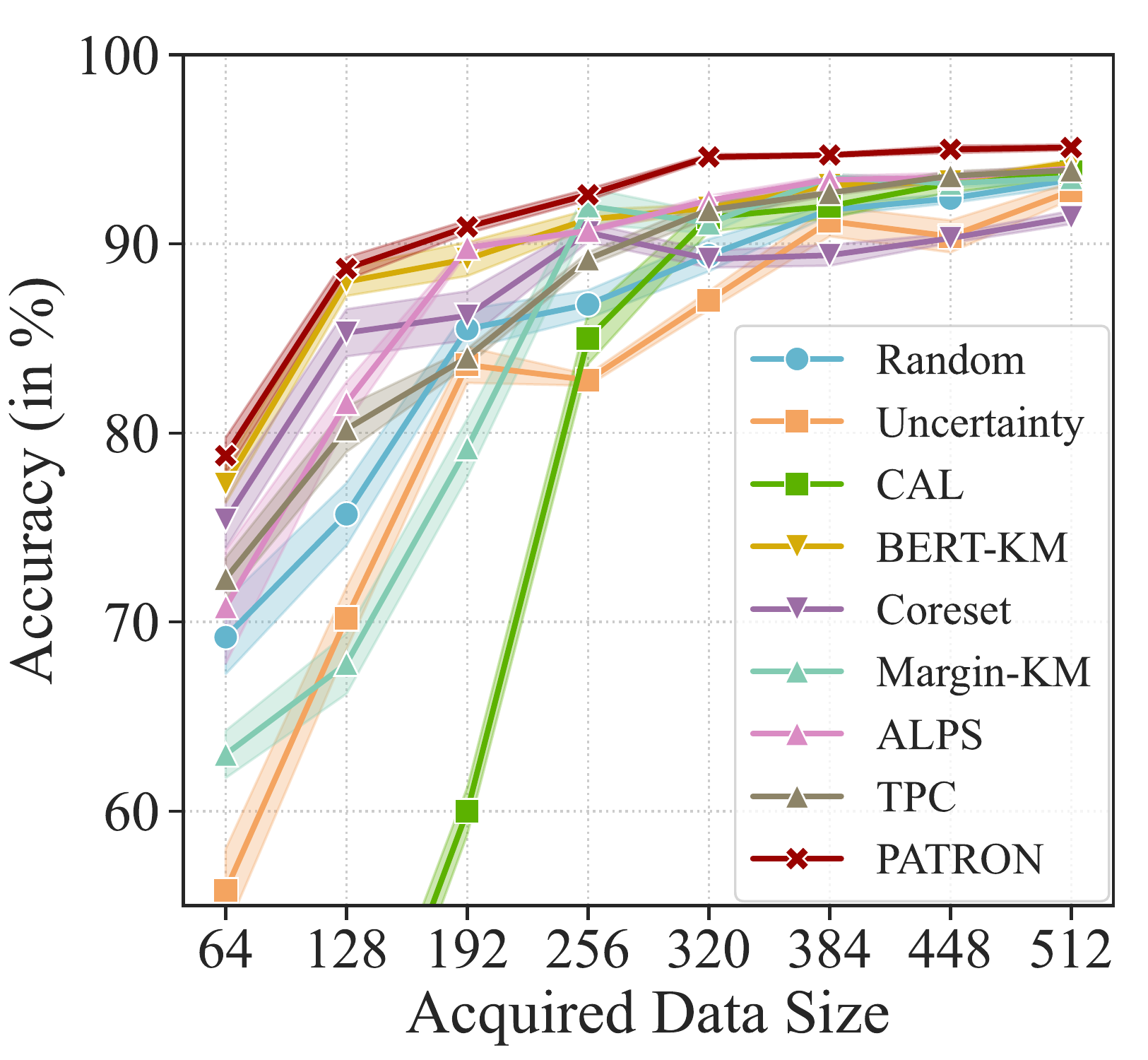}
            \label{fig:trec_al}
        }
        \hspace{-2mm}
        \subfigure[Yahoo!]{
            \includegraphics[width=0.32\textwidth]{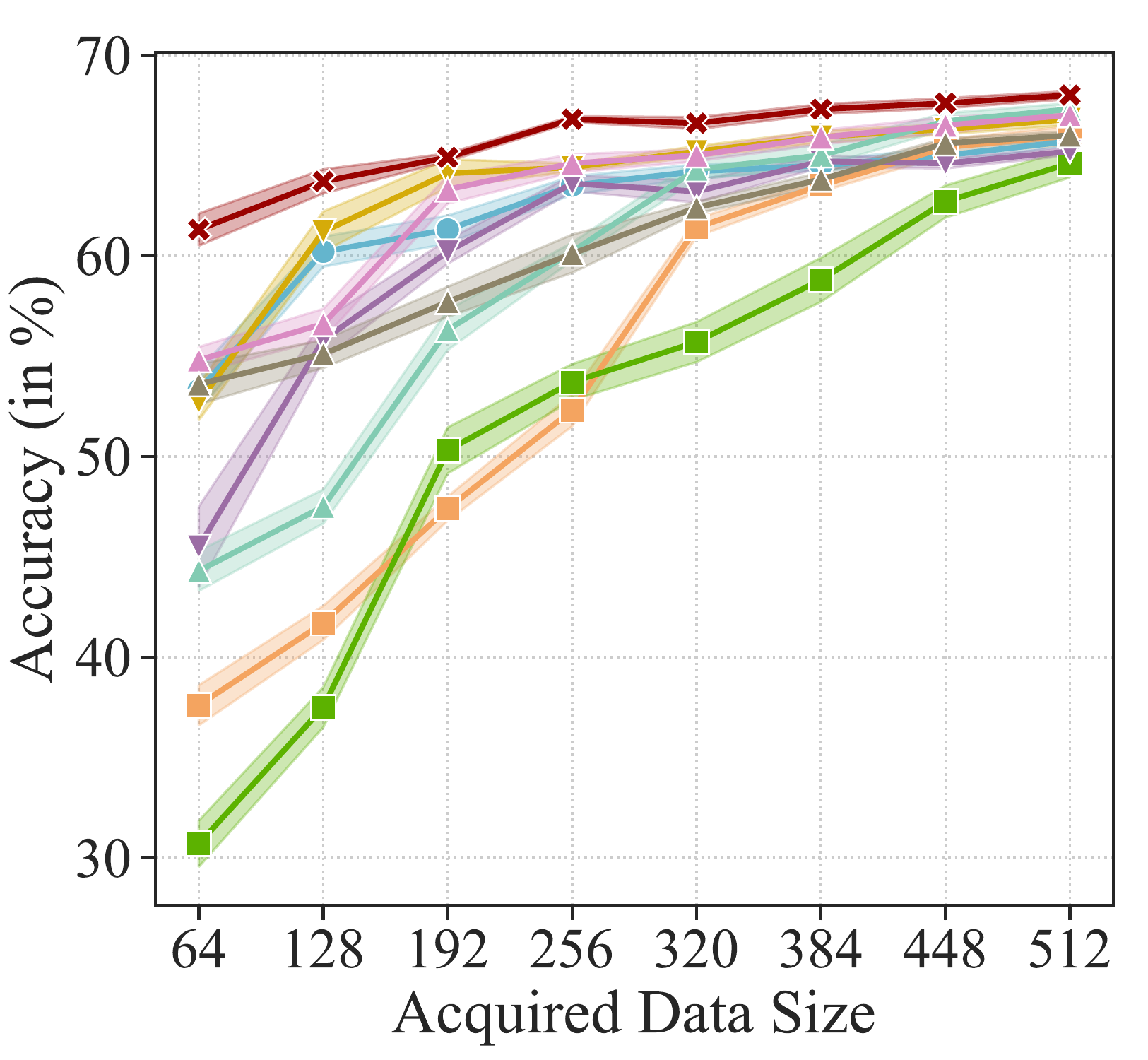}
            \label{fig:yahoo}
        }\hspace{-2mm}
        \subfigure[DBPedia]{
            \includegraphics[width=0.32\textwidth]{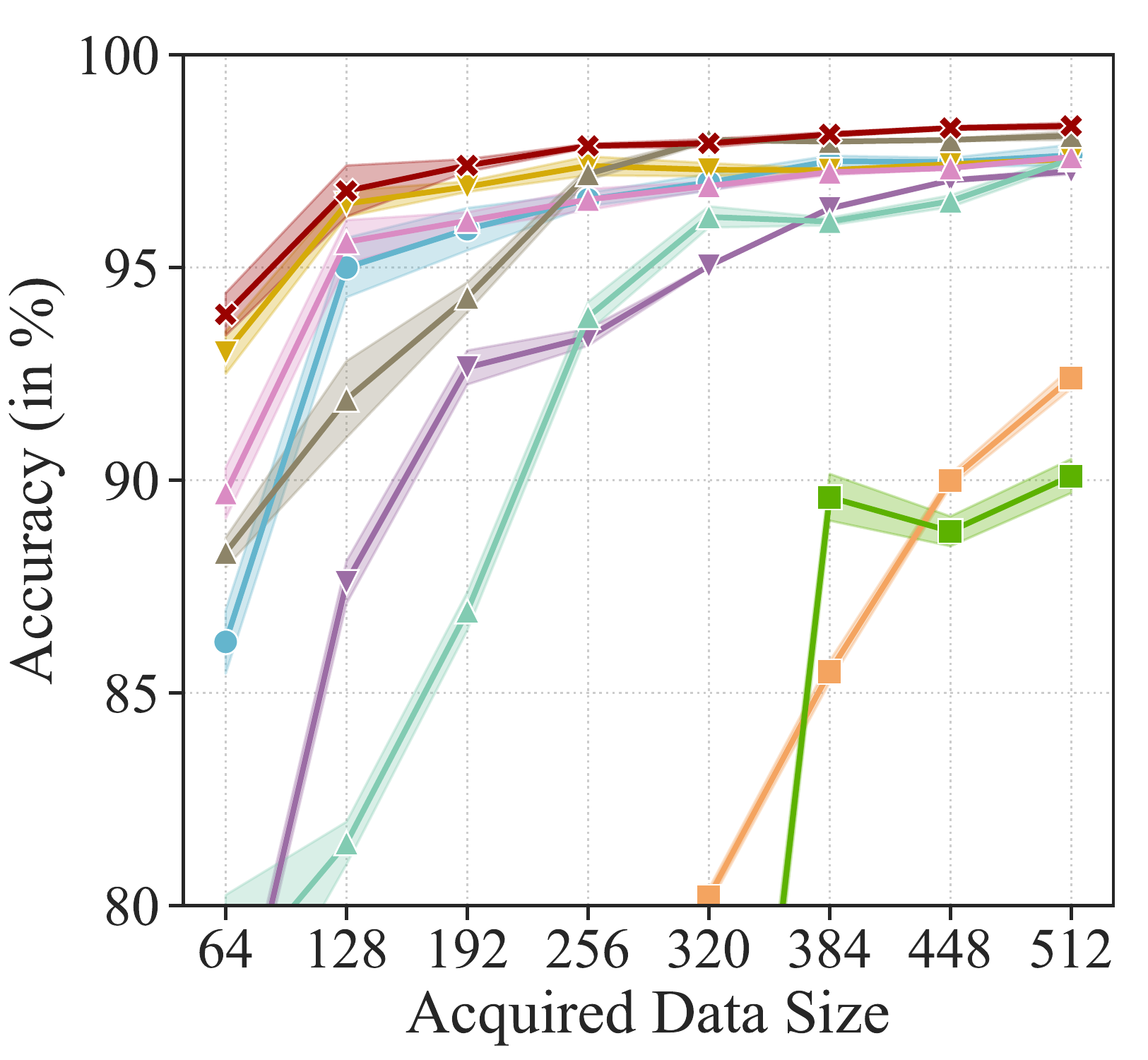}
            \label{fig:dbpedia_full}
        }\hspace{-2mm}
        \vspace{-1ex}
        \caption{The comparision of {\ours} with other baselines under  standard multi-round AL setting. }
        \label{fig:multi_round_al}
         \vspace{-1ex}
\end{figure*}

Here, we adapt {\ours} to other related settings to demonstrate its general applicability.

\noindent \textbf{Multi-round Low-budget Active Learning.} {\ours} can also be applied in standard multi-round active learning.
We study an AL setting where the labeling budget is set to $512$ and the queries to $64$ labels in each round ($8$ rounds in total).
More details are in Appendix~\ref{app:multi_round_al}.
Figure~\ref{fig:multi_round_al} shows the result of {\ours} and the baselines on 4 datasets.
From the results, we observe that {\ours} also achieves competitive performance when compared with baselines.
One exception is the IMDB dataset, where uncertainty-based methods outperform {\ours} when the annotation size is larger than 256.
This phenomenon indicates that when the labels are abundant and the cold-start issue is mitigated, uncertainty-based methods can be employed to further enhance the performance~\cite{yuan-etal-2020-cold}.
In this case, we can design \emph{hybrid strategies} to combine {\ours} and uncertainty-based methods for acquiring labeled data.

\noindent \textbf{Prompt-based Few-shot Learning.} 
Prompt-based Learning~\cite{liu2021pre} is another popular approach to promote the data efficiency for PLMs. 
To demonstrate the compatibility of {\ours} with prompt-based learning, we leverage the same prompt as the pseudo label generation part (Sec. ~\ref{sec:prop}), and use the same pipeline as LM-BFF~\cite{gao2021making} to fine-tune the PLM. 
Table~\ref{tab:lmbff_result_all} shows the result of  few-shot prompt-based learning using \{32, 64, 128\} samples. 
From the result, we find that LM-BFF performs better than vanilla fine-tuning with 12.5\% gain on average, which makes further improvements difficult. 
However, {\ours} still outperforms the best baseline by 2.0\%--4.5\%. 
We remark that {\ours} is naturally suitable for prompt-based learning, as we leverage the uncertainty derived from prompt-based predictions to assist data selection. 

\noindent \textbf{Semi-supervised Learning.}
When there are large amounts of unlabeled data, Semi-supervised Learning (SSL) methods can be used to improve AL performance.
Here, we choose two representative SSL methods: unsupervised data augmentation (UDA)~\cite{xie2020unsupervised} and self-training (ST)~\cite{du-etal-2021-self,yu-etal-2021-fine}\footnote{Details are in Appendix~\ref{app:ssl}.}.
Table~\ref{tab:ssl} exhibits the results for {\ours} and baselines. 
Notably, when the selection strategy is sub-optimal,  directly adopting SSL approaches cannot bring additional performance gains.
This is because the PLM fine-tuned on those samples is likely to produce incorrect pseudo labels. 
As a result, such incorrect labeled samples will hurt the final performance. 
In contrast, we observe that {\ours} leads to better performance for PLMs than baselines, which indicates the potentials of combining {\ours} with SSL approaches.

\begin{table}[!tb]
\centering 
\renewcommand\arraystretch{0.8}
\fontsize{7.5}{9.5}\selectfont 
\setlength{\tabcolsep}{0.5em}
\resizebox{0.95\linewidth}{!}{%
\begin{tabular}{l|ll|ll}
\toprule
\bf    Dataset ($\rightarrow$)    &  \multicolumn{2}{c|}{\bf AG News}  & \multicolumn{2}{c}{\bf TREC}  \\ \specialrule{.4pt}{1pt}{1pt}
\bf    Method  ($\downarrow$)   &  \multicolumn{1}{c}{UDA} & \multicolumn{1}{c|}{ST}  & \multicolumn{1}{c}{UDA} & \multicolumn{1}{c}{ST}  \\
\specialrule{.4pt}{1pt}{1pt}
Random & 78.0\std2.1	& 82.9\std1.5& 56.5\std3.0&	56.0\std2.5 \\
Uncertainty & 74.5\std1.6	&71.9\std2.0 &51.6\std1.5&	44.2\std2.3
 \\ 
CAL & 71.0\std2.0&	66.8\std2.7 &23.5\std2.1&	22.4\std2.1 \\\specialrule{.4pt}{1pt}{1pt}
BERT-KM&83.4\std1.0	&85.2\std1.1 &\underline{68.4\std1.6}	&67.2\std2.1 \\
Coreset &82.1\std1.0	&\underline{85.4\std0.6} &51.1\std2.0	&48.0\std2.4 \\\specialrule{.4pt}{1pt}{1pt}
Margin-KM &77.1\std1.2	&83.1\std1.4 &54.4\std1.8	&50.5\std1.6\\
ALPS &82.7\std0.8	&84.5\std0.8 &\underline{68.8\std1.6}&	\underline{71.0\std1.2}\\
TPC&\underline{83.8\std0.5}	&\underline{85.5\std0.4} &48.0\std1.9&	48.8\std2.1
 \\\specialrule{.4pt}{1pt}{1pt}
{\ours}&\bf 84.9\std0.5	&\bf 86.4\std0.3 &\bf 71.7\std1.0 &\bf	73.6\std0.5
 \\
\bottomrule
\end{tabular}
}
\caption{Experimental results for combining two semi-supervised learning: unsupervised data augmentation (UDA) and self-training (ST) with different data selection strategies on 2 datasets with the budget of 32 labels.}
\label{tab:ssl}
\vspace{-3ex}
\end{table}


\subsection{Out-of-Distribution (OOD) Evaluation}
\label{app:ood_full}
\begin{table*}[t]
\centering 
\renewcommand\arraystretch{0.8}
\fontsize{7.5}{9.5}\selectfont 
\setlength{\tabcolsep}{0.5em}
\resizebox{0.95\linewidth}{!}{%
\begin{tabular}{l|ccc|ccc|ccc}
\toprule
\multicolumn{1}{c|}{\multirow{2}{*}{\bf Datasets}}   &    \multicolumn{1}{c|}{SST-2} & \multicolumn{1}{c|}{IMDB} & \multicolumn{1}{c|}{IMDB} & \multicolumn{1}{c|}{SST-2} & \multicolumn{1}{c|}{IMDB} & \multicolumn{1}{c|}{IMDB} & \multicolumn{1}{c|}{SST-2} & \multicolumn{1}{c|}{IMDB} & \multicolumn{1}{c}{IMDB} \\ 
& \multicolumn{1}{c|}{Test} & \multicolumn{1}{c|}{Contrast} & \multicolumn{1}{c|}{Counterfactual} &
\multicolumn{1}{c|}{Test} & \multicolumn{1}{c|}{Contrast} & \multicolumn{1}{c|}{Counterfactual} & \multicolumn{1}{c|}{Test} & \multicolumn{1}{c|}{Contrast} & \multicolumn{1}{c}{Counterfactual} \\ \midrule 
\multicolumn{1}{c|}{\bf Budget $|B|$} &\multicolumn{3}{c|}{32} & \multicolumn{3}{c|}{64} & \multicolumn{3}{c}{128}  \\
\midrule
Random & 76.2\std2.4 &	76.1\std4.0	& 80.5\std4.7 & 80.0\std1.2	& 77.0\std1.1& 	80.8\std2.0 & 83.0\std2.1	& 83.8\std1.2	& 	87.9\std1.6
\\
Uncertainty & 78.0\std2.3& 	66.0\std4.0	 & 69.9\std3.1
 & 80.0\std1.5	& 75.5\std0.4&	82.6\std2.9
 & 83.6\std2.3	& 	81.6\std1.0	& 	85.6\std0.8
 \\ 
CAL         & 76.2\std3.1 &	76.5\std2.9	 & 77.6\std3.2
& 77.5\std3.5	& 76.7\std3.9&	78.7\std3.8
& 78.3\std3.4	& 	85.4\std0.9		& \underline{90.8\std0.8}
\\\midrule
BERT-KM     & 76.9\std1.3 &	75.6\std2.0	 & 81.2\std2.0  & \bf 81.5\std1.4	& 82.3\std4.2&	85.8\std4.4
&  84.6\std3.0	& 	86.2\std1.4	& 	90.3\std0.5
\\
Coreset     & 71.6\std2.0 &	60.7\std3.4 &	63.7\std4.3
& 79.6\std3.4	& 66.3\std5.5&	66.6\std4.4 
& \underline{82.2\std2.5}	& 	80.5\std2.6	& 	83.7\std3.6
\\\midrule
Margin-KM   & 71.5\std3.4 &	61.2\std3.0 &	57.5\std2.4 & 80.0\std3.0	&   74.9\std1.6 &	79.3\std2.5
& 80.9\std3.5	& 	\underline{86.8\std2.0}	& 	90.1\std2.3
\\
ALPS        & \underline{78.5\std1.9} &	\underline{78.5\std2.7} &	\underline{81.8\std2.4} & 77.8\std2.8	& \underline{83.1\std1.8} &	\underline{87.5\std1.5}
& 83.0\std3.2	& 	84.4\std1.5	& 	89.1\std1.4
\\
TPC         & 77.8\std3.8 &	72.1\std5.0 &	76.9\std6.1 & \underline{81.0\std0.9}	& 74.2\std1.2&	77.1\std2.2
& 79.3\std3.1	& 	83.0\std2.2	& 	87.5\std2.6
\\\midrule
{\ours}     & \bf 81.3\std2.6 &\bf 	81.9\std2.3 &\bf 	85.3\std2.1 
&  80.8\std2.7 &\bf 	84.7\std1.8	 &\bf  88.9\std1.0 
& \bf  85.9\std2.0	& \bf 	87.0\std1.5	& \bf 	92.2\std1.3
\\
\bottomrule
\end{tabular}
}
\caption{Full results of the evaluation on OOD tasks for IMDB datasets.}
\label{tab:ood_full}
\vspace{-1ex}
\end{table*}


We conduct Out-of-Distribution (OOD) evaluation to verify whether the methods can robustly select representative samples for the task instead of overfitting one specific dataset.
We use IMDB dataset as a source domain for data selection and fine-tuning, and then directly evaluate the fine-tuned model on 3 out-of-domain datasets (see Appendix~\ref{app:ood} for details): SST-2~\cite{socher-etal-2013-recursive}, IMDB Contrast Set (IMDB-CS)~\cite{gardner2020evaluating}, and IMDB Counterfactually Augmented Dataset (IMDB-CAD)~\cite{Kaushik2020Learning}.

As shown in Table~\ref{tab:ood_full}, diversity-based approaches also perform better than uncertainty-based methods on OOD tasks, due to the better coverage of the selected samples. 
However, {\ours} still outperforms these baselines by 3.2\% on average. The performance gains
illustrate that {\ours} can  discover informative samples to truly enable the PLM to capture task-specific linguistic knowledge instead of spurious features and improve 
the PLM's generalization ability under limited budget.

\begin{figure}[t]
    \centering
    \includegraphics[width=0.35\textwidth]{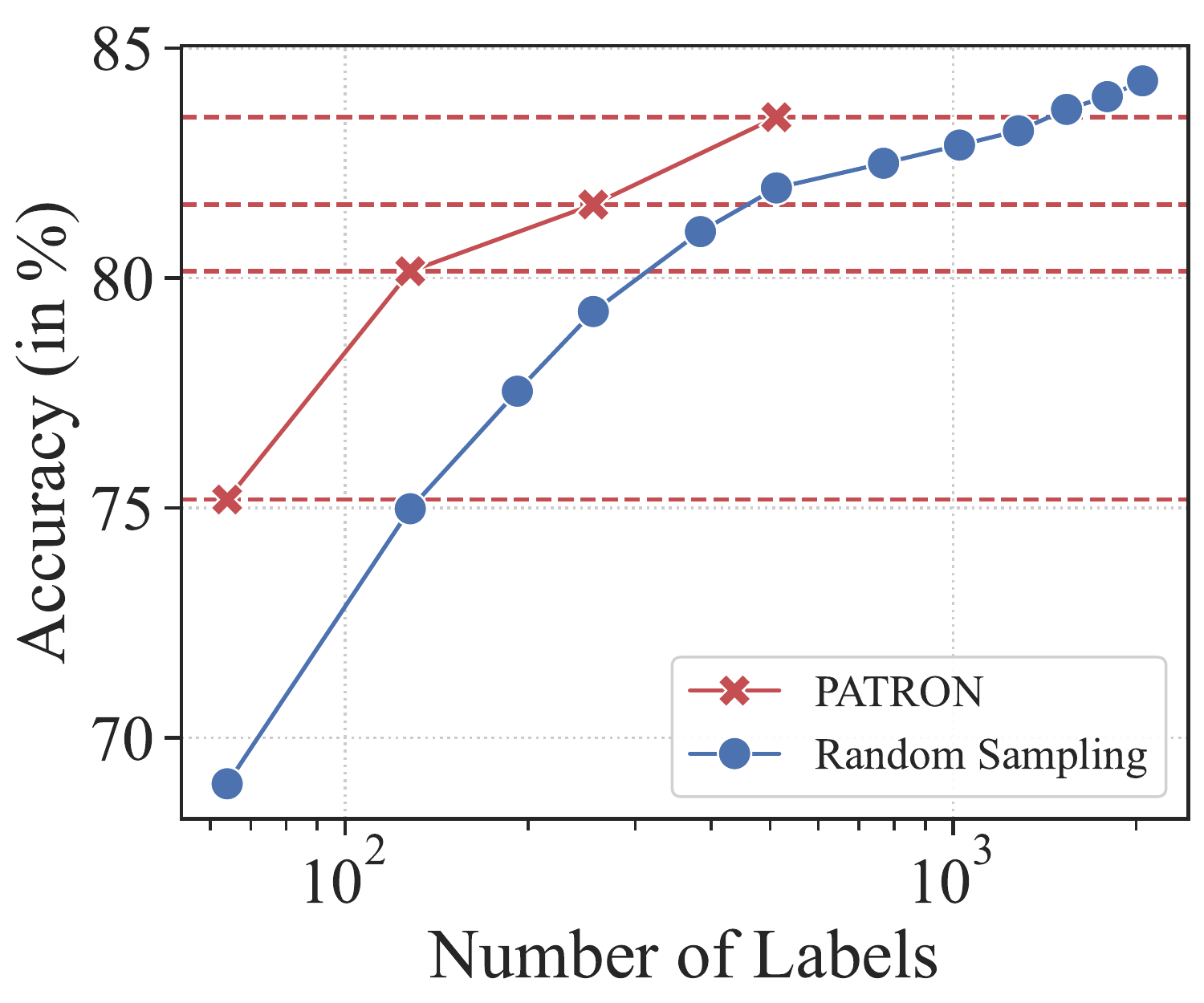}
    \vspace{-1ex}
    \caption{The comparison of {\ours} and random sampling with various volume of labeled data.}
    \vspace{-1ex}
    \label{fig:hybrid}
\end{figure}

\subsection{Label Efficiency Analysis}
\label{sec:label_efficiency}
Fig.~\ref{fig:hybrid} demonstrate the average performance on six datasets with different volume of labeled data selected via random sampling and {\ours}.
The label efficiency curve for each dataset is shown in Fig.~\ref{fig:efficiency_app}. 
We notice that {\ours} largely alleviates the label scarsity bottleneck: with 128 labels as the budget, {\ours} achieves better performance with 2X labels. Furthermore, after collecting 512 labels with multi-round AL (Sec~\ref{sec:multi_round_al}), {\ours} achieves 95\% of the fully-supervised performance on average, which is comparable with the performance using 3X labels based on random sampling. 
These results clearly justify that {\ours} is capable of promoting the label efficiency of PLMs.

\begin{figure*}[!t]
        \centering
         \subfigure[Ablation Study]{
            \includegraphics[width=0.24\textwidth]{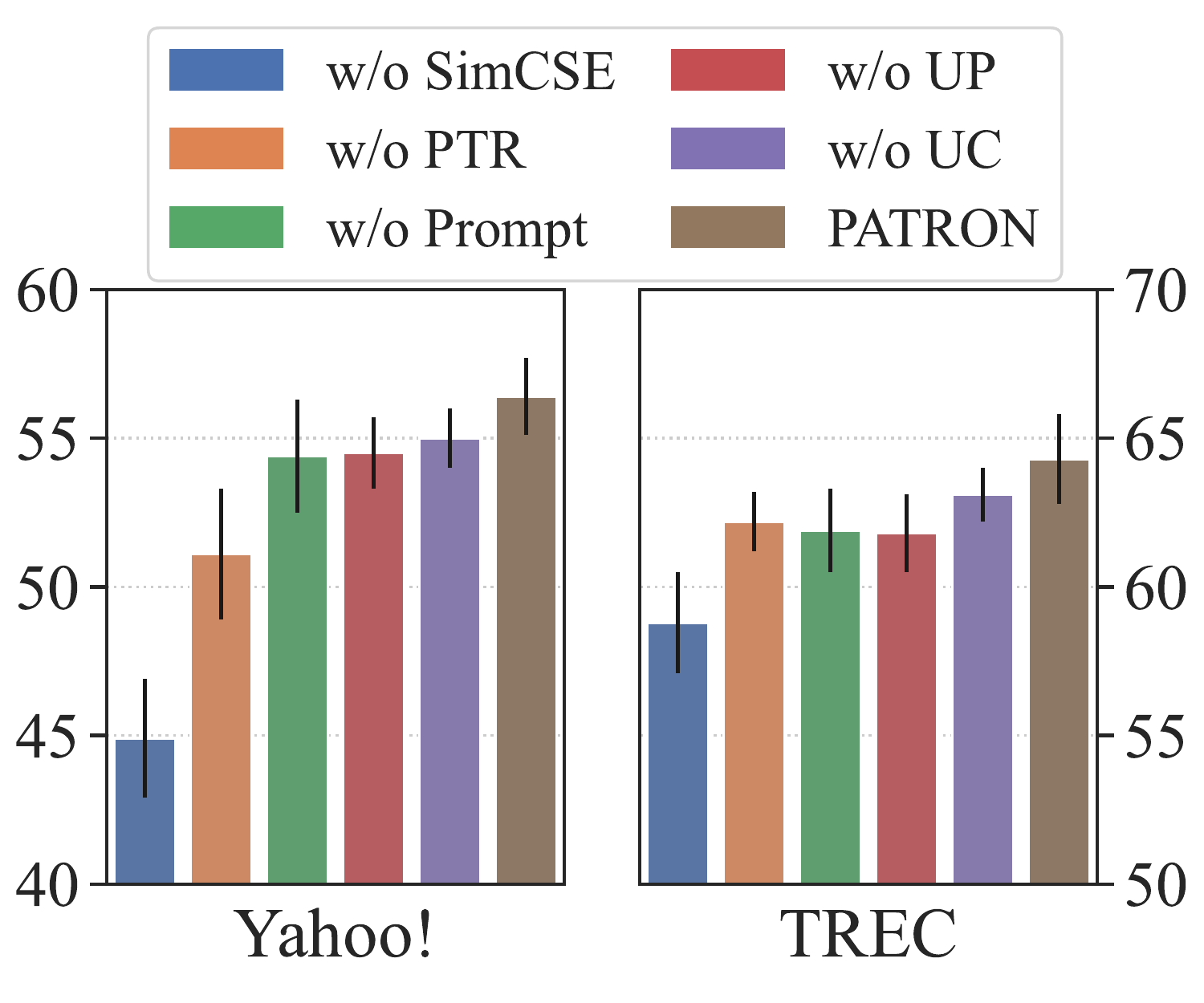}
            \label{fig:abla}
        }\hspace{-2mm}
        \subfigure[{Effect of $\rho$.}]{
            \includegraphics[width=0.24\textwidth]{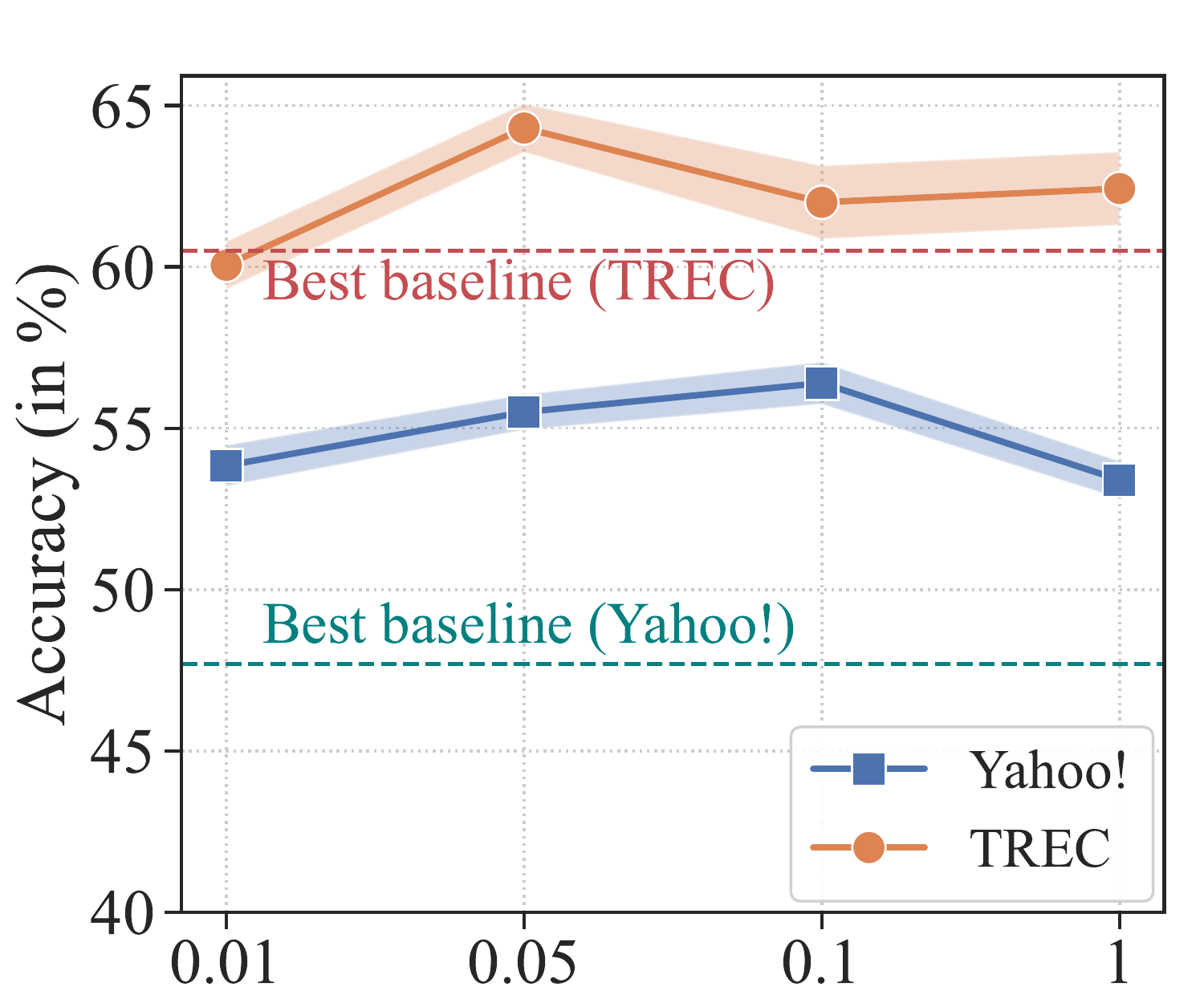}
            \label{fig:para_rho}
        }\hspace{-2.mm}
        \subfigure[{Effect of $\beta$.}]{
            \includegraphics[width=0.24\textwidth]{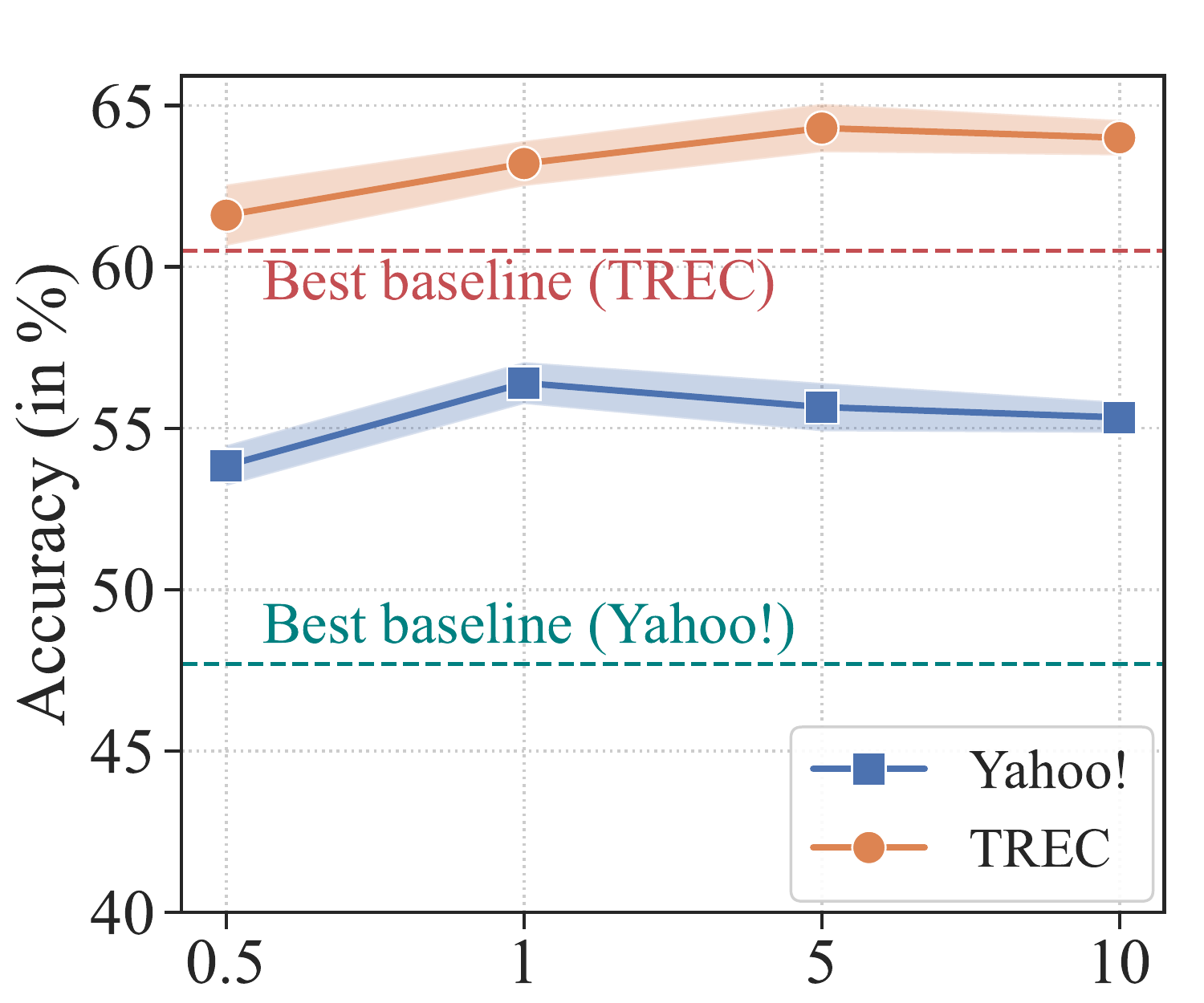}
            \label{fig:para_beta}
        }\hspace{-2.mm}
        \subfigure[Effect of $\gamma$.]{
            \includegraphics[width=0.24\textwidth]{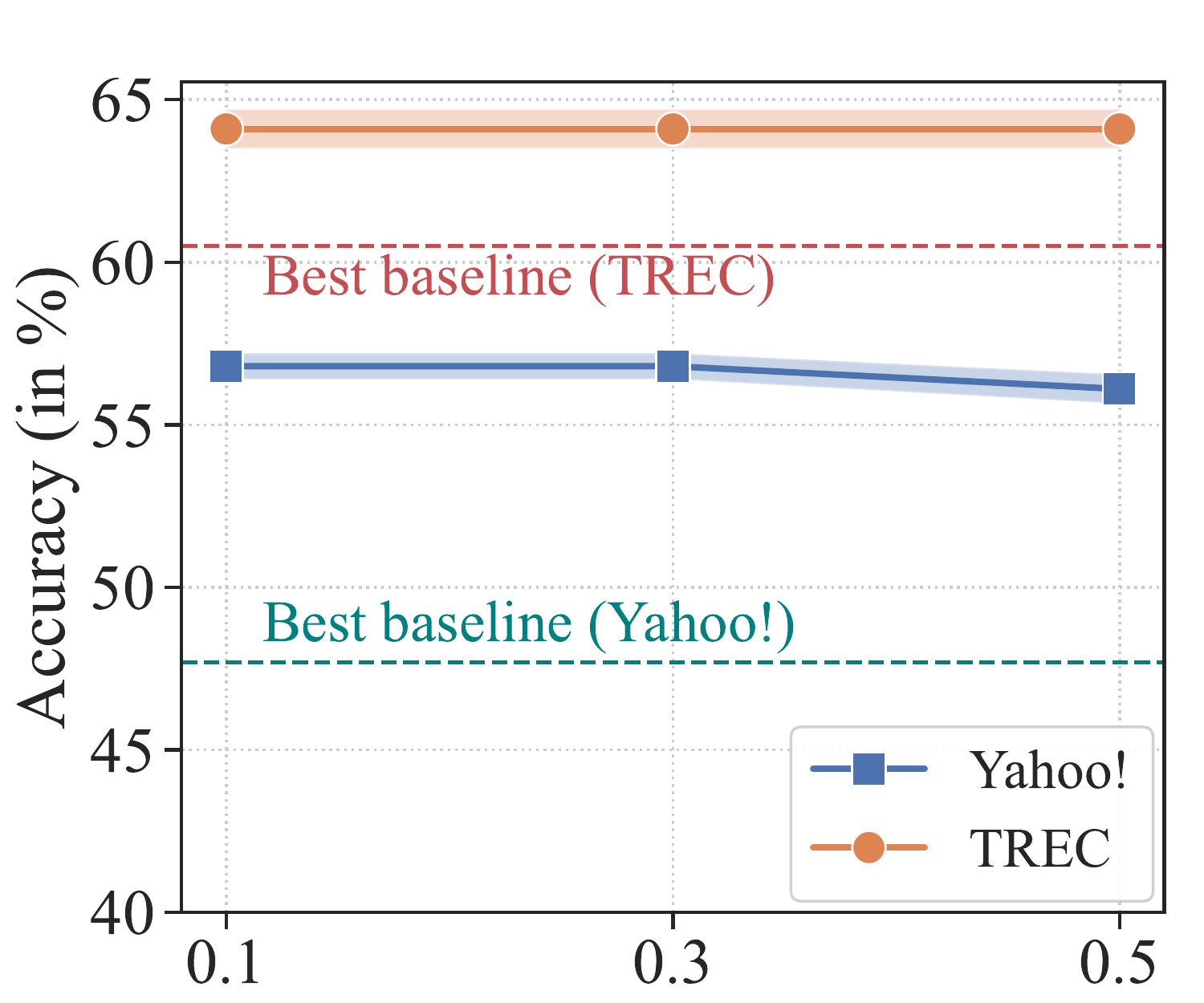}
            \label{fig:para_gamma}
        }
        \vspace{-1ex}
        \caption{Ablation and Hyper-parameter Study. }
        \label{fig:abla_main}
    \vskip -0.1in
\end{figure*}

\subsection{Ablation Study}
\label{sec:ablation}
We study the effects of different components of {\ours}, including the prompt-based uncertainty calibration in Eq.~\ref{eq:calibration}  and propagation in Eq.~\ref{eq:prop_final} (Prompt, UC and UP respectively), the feature encoder (SimCSE)\footnote{For {\ours} w/o Prompt, we use the same value $1$ to substitute the uncertainty in Eq.~\ref{eq:entropy}. 
For {\ours} w/o SimCSE, we use the RoBERTa-base  to generate document embeddings.}, as well as the {\ptr} strategy. 
We evaluated on the TREC and Yahoo! datasets with 32 labels as the budget. 
The results in  Fig.~\ref{fig:abla} show that all these components contribute to the final performance of \ours. 
We find that the usage of SimCSE brings considerable performance gains, as the embeddings generated via RoBERTa-base suffer from the \emph{degeneration} issue~\cite{li-etal-2020-sentence} and become less discriminative.   
Besides, the usage of prompts, as well as the two dedicated designs, namely uncertainty calibration and propagation enable us to complement the SimCSE embeddings with the prompt-based pseudo labels and improve the performance significantly. Lastly, {\ptr} helps regularize the distance among selected samples, which is also beneficial for AL.

\subsection{{\ours} is Robust to Hyperparameters} 
{\ours} introduces three additional hyperparameters, namely $\rho$ in Eq.~\ref{eq:prop}, $\beta$ in Eq.~\ref{eq:init} and $\gamma$ in Eq.~\ref{eq:init_updated}. 
To investigate their effects on the final performance,
Figure~ \ref{fig:para_rho}--\ref{fig:para_gamma} show the effects of them in {\ours} on two datasets with 32 labels as the budget. The results on other datasets are deferred to Figure~\ref{fig:hyper_app} in Appendix~\ref{app:param}. 

In general, the model is \emph{robust} to them as the {\ours} outperforms the baselines in most cases with different sets of hyperparameters. 
We also notice that the performance is not sensitive to $\gamma$.  
Besides, the performance first increases then decreases for both $\rho$ and $\beta$. 
For $\rho$, setting it too large makes the propagated uncertainty too small, and setting it too small makes the influence of neighbor samples too strong and hurt data utility estimation. 
For $\beta$, the sampled data is less informative with a too large $\beta$, while being too close from others during initialization with a too small $\beta$. 
In practice, we can always search for optimal hyperparameters in an \emph{iterative} manner: in each iteration, we only tune one hyperparameter and keep others fixed. 
In addition, there are automatically hyperparameter optimization tools~\cite{optuna} that can efficiently find the optimal hyperparameters.

To sum up, although {\ours} includes additional hyperparameters, it will not introduce much difficulty for hyperparameter tuning. Instead, these  hyperparameters improve the modeling flexibility of {\ours} to adapt to different tasks.

\section{Discussion}

\noindent \textbf{Connection to Weakly-supervised Learning.} Our method can also be considered as \emph{weakly-supervised} data selection, where only class-indicating keywords as well templates are provided.  
Although such formulations have been adopted for NLP tasks~\cite{yelp-full,meng-etal-2020-text,hu2022knowledgeable} (see \citet{zhang2022survey} for a detailed survey), how to effectively leverage such weak supervision signals for data selection has not been widely explored in NLP. 
In this study, we tackle this research problem to facilitate few-shot PLM fine-tuning, and demonstrate such task-specific weak supervision is beneficial for  downstream tasks.

\noindent \textbf{Data Selection under Low and High Budget.} 
In this study, we mainly focus on \emph{cold-start} setting to select data without any labeled data. 
This is different from traditional AL pipelines, and we do not claim {\ours} outperforms AL methods under high-budget scenarios. 
However, experiments show our method shines under low-budget setting, and {\ours} can also be leveraged in earlier rounds of standard AL to improve the label efficiency.

\section{Conclusion}
We develop \ours, a prompt-based  data selection  method for pre-trained language models (PLMs) under cold-start scenarios.
By leveraging prompts, we can distill the task-specific knowledge from the frozen PLM to guide data acquisition. 
Moreover, we develop two techniques, namely uncertainty propagation and predict-then-rewrite (\ptr) to achieve both sample representativeness and diversity. 
The experiments on six text classification tasks demonstrate the superiority of {\ours} against baselines for few-shot PLM fine-tuning.  

\section*{Limitations} In this work, we focus on designing strategies for PLMs with the MLM-style pre-training objective and do not account for other types of pre-trained language models such as discriminative PLMs~\cite{Clark2020ELECTRA,shen-etal-2021-training}. 
However, as there are recent works that aim to design prompts for discriminative PLMs~\cite{yao2022prompt,xia2022prompting}, {\ours} can be potentially combined with them to improve the data efficiency. 

We are also aware that there exists advanced few-shot fine-tuning techniques for PLMs recently~\cite[\emph{inter alia}]{hu2022knowledgeable,tam-etal-2021-improving,zhang2022differentiable}. 
We argue that {\ours} does not rely on a specific fine-tuning method, and can be combined with them to further improve the performance. 

\section*{Acknowledgements}

We would like to thank the anonymous reviewers from the ACL Rolling Review for their feedbacks. This work was supported in part by NSF IIS-2008334, IIS-2106961, CAREER IIS-2144338, and ONR MURI N00014-17-1-2656.

\bibliography{anthology,dataset}

\begin{thebibliography}{69}
\expandafter\ifx\csname natexlab\endcsname\relax\def\natexlab#1{#1}\fi

\bibitem[{Aharoni and Goldberg(2020)}]{aharoni2020unsupervised}
Roee Aharoni and Yoav Goldberg. 2020.
\newblock \href {https://doi.org/10.18653/v1/2020.acl-main.692} {Unsupervised
  domain clusters in pretrained language models}.
\newblock In \emph{Proceedings of the 58th Annual Meeting of the Association
  for Computational Linguistics}, pages 7747--7763, Online. Association for
  Computational Linguistics.

\bibitem[{Akiba et~al.(2019)Akiba, Sano, Yanase, Ohta, and Koyama}]{optuna}
Takuya Akiba, Shotaro Sano, Toshihiko Yanase, Takeru Ohta, and Masanori Koyama.
  2019.
\newblock \href {https://doi.org/10.1145/3292500.3330701} {Optuna: A
  next-generation hyperparameter optimization framework}.
\newblock In \emph{Proceedings of the 25th ACM SIGKDD International Conference
  on Knowledge Discovery \& Data Mining}, KDD '19, page 2623–2631.

\bibitem[{Ash et~al.(2020)Ash, Zhang, Krishnamurthy, Langford, and
  Agarwal}]{badge}
Jordan~T. Ash, Chicheng Zhang, Akshay Krishnamurthy, John Langford, and Alekh
  Agarwal. 2020.
\newblock \href {https://openreview.net/forum?id=ryghZJBKPS} {Deep batch active
  learning by diverse, uncertain gradient lower bounds}.
\newblock In \emph{International Conference on Learning Representations}.

\bibitem[{Bragg et~al.(2021)Bragg, Cohan, Lo, and Beltagy}]{bragg2021flex}
Jonathan Bragg, Arman Cohan, Kyle Lo, and Iz~Beltagy. 2021.
\newblock Flex: Unifying evaluation for few-shot nlp.
\newblock \emph{Advances in Neural Information Processing Systems}, 34.

\bibitem[{Brown et~al.(2020)Brown, Mann, Ryder, Subbiah, Kaplan, Dhariwal,
  Neelakantan, Shyam, Sastry, Askell et~al.}]{brown2020language}
Tom Brown, Benjamin Mann, Nick Ryder, Melanie Subbiah, Jared~D Kaplan, Prafulla
  Dhariwal, Arvind Neelakantan, Pranav Shyam, Girish Sastry, Amanda Askell,
  et~al. 2020.
\newblock Language models are few-shot learners.
\newblock \emph{Advances in neural information processing systems},
  33:1877--1901.

\bibitem[{Chang et~al.(2021)Chang, Shen, Yeh, and
  Demberg}]{chang-etal-2021-training}
Ernie Chang, Xiaoyu Shen, Hui-Syuan Yeh, and Vera Demberg. 2021.
\newblock \href {https://doi.org/10.18653/v1/2021.acl-short.2} {On training
  instance selection for few-shot neural text generation}.
\newblock In \emph{Proceedings of the 59th Annual Meeting of the Association
  for Computational Linguistics and the 11th International Joint Conference on
  Natural Language Processing (Volume 2: Short Papers)}, pages 8--13, Online.
  Association for Computational Linguistics.

\bibitem[{Chen et~al.(2021)Chen, Wang, and Jia}]{chen2022zeroround}
Si~Chen, Tianhao Wang, and Ruoxi Jia. 2021.
\newblock Zero-round active learning.
\newblock \emph{arXiv preprint arXiv:2107.06703}.

\bibitem[{Chen and Tian(2019)}]{chen2019learning}
Xinyun Chen and Yuandong Tian. 2019.
\newblock Learning to perform local rewriting for combinatorial optimization.
\newblock \emph{Advances in Neural Information Processing Systems}, 32.

\bibitem[{Clark et~al.(2020)Clark, Luong, Le, and Manning}]{Clark2020ELECTRA}
Kevin Clark, Minh-Thang Luong, Quoc~V. Le, and Christopher~D. Manning. 2020.
\newblock \href {https://openreview.net/forum?id=r1xMH1BtvB} {Electra:
  Pre-training text encoders as discriminators rather than generators}.
\newblock In \emph{International Conference on Learning Representations}.

\bibitem[{Devlin et~al.(2019)Devlin, Chang, Lee, and Toutanova}]{bert}
Jacob Devlin, Ming-Wei Chang, Kenton Lee, and Kristina Toutanova. 2019.
\newblock \href {https://doi.org/10.18653/v1/N19-1423} {{BERT}: Pre-training of
  deep bidirectional transformers for language understanding}.
\newblock In \emph{Proceedings of the 2019 Conference of the North {A}merican
  Chapter of the Association for Computational Linguistics: Human Language
  Technologies, Volume 1 (Long and Short Papers)}, pages 4171--4186,
  Minneapolis, Minnesota. Association for Computational Linguistics.

\bibitem[{Ding et~al.(2022)Ding, Hu, Zhao, Chen, Liu, Zheng, and
  Sun}]{openprompt}
Ning Ding, Shengding Hu, Weilin Zhao, Yulin Chen, Zhiyuan Liu, Haitao Zheng,
  and Maosong Sun. 2022.
\newblock \href {https://doi.org/10.18653/v1/2022.acl-demo.10} {{O}pen{P}rompt:
  An open-source framework for prompt-learning}.
\newblock In \emph{Proceedings of the 60th Annual Meeting of the Association
  for Computational Linguistics: System Demonstrations}, pages 105--113,
  Dublin, Ireland. Association for Computational Linguistics.

\bibitem[{Du et~al.(2021)Du, Grave, Gunel, Chaudhary, Celebi, Auli, Stoyanov,
  and Conneau}]{du-etal-2021-self}
Jingfei Du, Edouard Grave, Beliz Gunel, Vishrav Chaudhary, Onur Celebi, Michael
  Auli, Veselin Stoyanov, and Alexis Conneau. 2021.
\newblock \href {https://doi.org/10.18653/v1/2021.naacl-main.426}
  {Self-training improves pre-training for natural language understanding}.
\newblock In \emph{Proceedings of the 2021 Conference of the North American
  Chapter of the Association for Computational Linguistics: Human Language
  Technologies}, pages 5408--5418. Association for Computational Linguistics.

\bibitem[{Ein-Dor et~al.(2020)Ein-Dor, Halfon, Gera, Shnarch, Dankin, Choshen,
  Danilevsky, Aharonov, Katz, and Slonim}]{ein-etal-2020-active}
Liat Ein-Dor, Alon Halfon, Ariel Gera, Eyal Shnarch, Lena Dankin, Leshem
  Choshen, Marina Danilevsky, Ranit Aharonov, Yoav Katz, and Noam Slonim. 2020.
\newblock \href {https://doi.org/10.18653/v1/2020.emnlp-main.638} {{A}ctive
  {L}earning for {BERT}: {A}n {E}mpirical {S}tudy}.
\newblock In \emph{Proceedings of the 2020 Conference on Empirical Methods in
  Natural Language Processing (EMNLP)}, pages 7949--7962. Association for
  Computational Linguistics.

\bibitem[{Gao et~al.(2021{\natexlab{a}})Gao, Fisch, and Chen}]{gao2021making}
Tianyu Gao, Adam Fisch, and Danqi Chen. 2021{\natexlab{a}}.
\newblock \href {https://doi.org/10.18653/v1/2021.acl-long.295} {Making
  pre-trained language models better few-shot learners}.
\newblock In \emph{Proceedings of the 59th Annual Meeting of the Association
  for Computational Linguistics and the 11th International Joint Conference on
  Natural Language Processing (Volume 1: Long Papers)}, pages 3816--3830,
  Online. Association for Computational Linguistics.

\bibitem[{Gao et~al.(2021{\natexlab{b}})Gao, Yao, and
  Chen}]{gao-etal-2021-simcse}
Tianyu Gao, Xingcheng Yao, and Danqi Chen. 2021{\natexlab{b}}.
\newblock \href {https://doi.org/10.18653/v1/2021.emnlp-main.552} {{S}im{CSE}:
  Simple contrastive learning of sentence embeddings}.
\newblock In \emph{Proceedings of the 2021 Conference on Empirical Methods in
  Natural Language Processing}, pages 6894--6910, Online and Punta Cana,
  Dominican Republic. Association for Computational Linguistics.

\bibitem[{Gardner et~al.(2020)Gardner, Artzi, Basmov, Berant, Bogin, Chen,
  Dasigi, Dua et~al.}]{gardner2020evaluating}
Matt Gardner, Yoav Artzi, Victoria Basmov, Jonathan Berant, Ben Bogin, Sihao
  Chen, Pradeep Dasigi, Dheeru Dua, et~al. 2020.
\newblock \href {https://doi.org/10.18653/v1/2020.findings-emnlp.117}
  {Evaluating models{'} local decision boundaries via contrast sets}.
\newblock In \emph{Findings of the Association for Computational Linguistics:
  EMNLP 2020}, pages 1307--1323, Online. Association for Computational
  Linguistics.

\bibitem[{Hacohen et~al.(2022)Hacohen, Dekel, and
  Weinshall}]{hacohen2022active}
Guy Hacohen, Avihu Dekel, and Daphna Weinshall. 2022.
\newblock \href {https://proceedings.mlr.press/v162/hacohen22a.html} {Active
  learning on a budget: Opposite strategies suit high and low budgets}.
\newblock In \emph{Proceedings of the 39th International Conference on Machine
  Learning}, Proceedings of Machine Learning Research, pages 8175--8195. PMLR.

\bibitem[{Hu et~al.(2019)Hu, Lipton, Anandkumar, and Ramanan}]{hu2018active}
Peiyun Hu, Zack Lipton, Anima Anandkumar, and Deva Ramanan. 2019.
\newblock \href {https://openreview.net/forum?id=HJfSEnRqKQ} {Active learning
  with partial feedback}.
\newblock In \emph{International Conference on Learning Representations}.

\bibitem[{Hu et~al.(2022)Hu, Ding, Wang, Liu, Wang, Li, Wu, and
  Sun}]{hu2022knowledgeable}
Shengding Hu, Ning Ding, Huadong Wang, Zhiyuan Liu, Jingang Wang, Juanzi Li,
  Wei Wu, and Maosong Sun. 2022.
\newblock \href {https://aclanthology.org/2022.acl-long.158} {Knowledgeable
  prompt-tuning: Incorporating knowledge into prompt verbalizer for text
  classification}.
\newblock In \emph{Proceedings of the 60th Annual Meeting of the Association
  for Computational Linguistics (Volume 1: Long Papers)}, pages 2225--2240,
  Dublin, Ireland. Association for Computational Linguistics.

\bibitem[{Johnson et~al.(2019)Johnson, Douze, and J{\'e}gou}]{FAISS}
Jeff Johnson, Matthijs Douze, and Herv{\'e} J{\'e}gou. 2019.
\newblock Billion-scale similarity search with gpus.
\newblock \emph{IEEE Transactions on Big Data}.

\bibitem[{Karamcheti et~al.(2021)Karamcheti, Krishna, Fei-Fei, and
  Manning}]{karamcheti2021mind}
Siddharth Karamcheti, Ranjay Krishna, Li~Fei-Fei, and Christopher Manning.
  2021.
\newblock \href {https://doi.org/10.18653/v1/2021.acl-long.564} {Mind your
  outliers! investigating the negative impact of outliers on active learning
  for visual question answering}.
\newblock In \emph{Proceedings of the 59th Annual Meeting of the Association
  for Computational Linguistics and the 11th International Joint Conference on
  Natural Language Processing (Volume 1: Long Papers)}, pages 7265--7281,
  Online. Association for Computational Linguistics.

\bibitem[{Kaushik et~al.(2020)Kaushik, Hovy, and Lipton}]{Kaushik2020Learning}
Divyansh Kaushik, Eduard Hovy, and Zachary Lipton. 2020.
\newblock \href {https://openreview.net/forum?id=Sklgs0NFvr} {Learning the
  difference that makes a difference with counterfactually-augmented data}.
\newblock In \emph{International Conference on Learning Representations}.

\bibitem[{Lehmann et~al.(2015)Lehmann, Isele, Jakob, Jentzsch, Kontokostas,
  Mendes, Hellmann, Morsey, Van~Kleef, Auer et~al.}]{lehmann2015dbpedia}
Jens Lehmann, Robert Isele, Max Jakob, Anja Jentzsch, Dimitris Kontokostas,
  Pablo~N Mendes, Sebastian Hellmann, Mohamed Morsey, Patrick Van~Kleef,
  S{\"o}ren Auer, et~al. 2015.
\newblock Dbpedia--a large-scale, multilingual knowledge base extracted from
  wikipedia.
\newblock \emph{Semantic web}, 6(2):167--195.

\bibitem[{Lewis and Gale(1994)}]{lewis1994entropy}
David~D Lewis and William~A Gale. 1994.
\newblock A sequential algorithm for training text classifiers.
\newblock In \emph{Proceedings of the 17th annual international ACM SIGIR
  conference on Research and development in information retrieval}, pages
  3--12.

\bibitem[{Li et~al.(2020)Li, Zhou, He, Wang, Yang, and
  Li}]{li-etal-2020-sentence}
Bohan Li, Hao Zhou, Junxian He, Mingxuan Wang, Yiming Yang, and Lei Li. 2020.
\newblock \href {https://doi.org/10.18653/v1/2020.emnlp-main.733} {On the
  sentence embeddings from pre-trained language models}.
\newblock In \emph{Proceedings of the 2020 Conference on Empirical Methods in
  Natural Language Processing (EMNLP)}, pages 9119--9130, Online. Association
  for Computational Linguistics.

\bibitem[{Li and Roth(2002)}]{li2002learning}
Xin Li and Dan Roth. 2002.
\newblock \href {https://aclanthology.org/C02-1150} {Learning question
  classifiers}.
\newblock In \emph{The 19th International Conference on Computational
  Linguistics}.

\bibitem[{Liu et~al.(2021{\natexlab{a}})Liu, Shen, Zhang, Dolan, Carin, and
  Chen}]{liu2021makes}
Jiachang Liu, Dinghan Shen, Yizhe Zhang, Bill Dolan, Lawrence Carin, and Weizhu
  Chen. 2021{\natexlab{a}}.
\newblock What makes good in-context examples for gpt-$3$?
\newblock \emph{arXiv preprint arXiv:2101.06804}.

\bibitem[{Liu et~al.(2021{\natexlab{b}})Liu, Yuan, Fu, Jiang, Hayashi, and
  Neubig}]{liu2021pre}
Pengfei Liu, Weizhe Yuan, Jinlan Fu, Zhengbao Jiang, Hiroaki Hayashi, and
  Graham Neubig. 2021{\natexlab{b}}.
\newblock Pre-train, prompt, and predict: A systematic survey of prompting
  methods in natural language processing.
\newblock \emph{arXiv preprint arXiv:2107.13586}.

\bibitem[{Liu et~al.(2019)Liu, Ott, Goyal, Du, Joshi, Chen, Levy, Lewis,
  Zettlemoyer, and Stoyanov}]{liu2019roberta}
Yinhan Liu, Myle Ott, Naman Goyal, Jingfei Du, Mandar Joshi, Danqi Chen, Omer
  Levy, Mike Lewis, Luke Zettlemoyer, and Veselin Stoyanov. 2019.
\newblock Roberta: A robustly optimized bert pretraining approach.
\newblock \emph{arXiv preprint arXiv:1907.11692}.

\bibitem[{Loshchilov and Hutter(2019)}]{loshchilov2018adamw}
Ilya Loshchilov and Frank Hutter. 2019.
\newblock \href {https://openreview.net/forum?id=Bkg6RiCqY7} {Decoupled weight
  decay regularization}.
\newblock In \emph{International Conference on Learning Representations}.

\bibitem[{Maas et~al.(2011)Maas, Daly, Pham, Huang, Ng, and Potts}]{imdb}
Andrew~L. Maas, Raymond~E. Daly, Peter~T. Pham, Dan Huang, Andrew~Y. Ng, and
  Christopher Potts. 2011.
\newblock \href {https://aclanthology.org/P11-1015} {Learning word vectors for
  sentiment analysis}.
\newblock In \emph{Proceedings of the 49th Annual Meeting of the Association
  for Computational Linguistics: Human Language Technologies}, pages 142--150,
  Portland, Oregon, USA. Association for Computational Linguistics.

\bibitem[{Margatina et~al.(2022)Margatina, Barrault, and
  Aletras}]{margatina2022importance}
Katerina Margatina, Loic Barrault, and Nikolaos Aletras. 2022.
\newblock \href {https://aclanthology.org/2022.acl-short.93} {On the importance
  of effectively adapting pretrained language models for active learning}.
\newblock In \emph{Proceedings of the 60th Annual Meeting of the Association
  for Computational Linguistics (Volume 2: Short Papers)}, pages 825--836,
  Dublin, Ireland. Association for Computational Linguistics.

\bibitem[{Margatina et~al.(2021)Margatina, Vernikos, Barrault, and
  Aletras}]{margatina2021active}
Katerina Margatina, Giorgos Vernikos, Lo{\"\i}c Barrault, and Nikolaos Aletras.
  2021.
\newblock \href {https://aclanthology.org/2021.emnlp-main.51} {Active learning
  by acquiring contrastive examples}.
\newblock In \emph{Proceedings of the 2021 Conference on Empirical Methods in
  Natural Language Processing}, pages 650--663, Online and Punta Cana,
  Dominican Republic. Association for Computational Linguistics.

\bibitem[{Meng et~al.(2019)Meng, Shen, Zhang, and Han}]{yelp-full}
Yu~Meng, Jiaming Shen, Chao Zhang, and Jiawei Han. 2019.
\newblock Weakly-supervised hierarchical text classification.
\newblock In \emph{Proceedings of the AAAI conference on artificial
  intelligence}, volume~33, pages 6826--6833.

\bibitem[{Meng et~al.(2020)Meng, Zhang, Huang, Xiong, Ji, Zhang, and
  Han}]{meng-etal-2020-text}
Yu~Meng, Yunyi Zhang, Jiaxin Huang, Chenyan Xiong, Heng Ji, Chao Zhang, and
  Jiawei Han. 2020.
\newblock \href {https://doi.org/10.18653/v1/2020.emnlp-main.724} {Text
  classification using label names only: A language model self-training
  approach}.
\newblock In \emph{Proceedings of the 2020 Conference on Empirical Methods in
  Natural Language Processing (EMNLP)}, pages 9006--9017. Association for
  Computational Linguistics.

\bibitem[{Min et~al.(2022)Min, Lewis, Hajishirzi, and
  Zettlemoyer}]{min-etal-2022-noisy}
Sewon Min, Mike Lewis, Hannaneh Hajishirzi, and Luke Zettlemoyer. 2022.
\newblock \href {https://doi.org/10.18653/v1/2022.acl-long.365} {Noisy channel
  language model prompting for few-shot text classification}.
\newblock In \emph{Proceedings of the 60th Annual Meeting of the Association
  for Computational Linguistics (Volume 1: Long Papers)}, pages 5316--5330,
  Dublin, Ireland. Association for Computational Linguistics.

\bibitem[{M{\"u}ller et~al.(2022)M{\"u}ller, P{\'e}rez-Torr{\'o}, Basile, and
  Franco-Salvador}]{muller2022active}
Thomas M{\"u}ller, Guillermo P{\'e}rez-Torr{\'o}, Angelo Basile, and Marc
  Franco-Salvador. 2022.
\newblock Active few-shot learning with fasl.
\newblock \emph{arXiv preprint arXiv:2204.09347}.

\bibitem[{Park et~al.(2022)Park, Ahmad, and Hain}]{park2022unsupervised}
Chanho Park, Rehan Ahmad, and Thomas Hain. 2022.
\newblock Unsupervised data selection for speech recognition with contrastive
  loss ratios.
\newblock In \emph{ICASSP 2022-2022 IEEE International Conference on Acoustics,
  Speech and Signal Processing (ICASSP)}, pages 8587--8591. IEEE.

\bibitem[{Raffel et~al.(2020)Raffel, Shazeer, Roberts, Lee, Narang, Matena,
  Zhou, Li, and Liu}]{raffel2020exploring}
Colin Raffel, Noam Shazeer, Adam Roberts, Katherine Lee, Sharan Narang, Michael
  Matena, Yanqi Zhou, Wei Li, and Peter~J Liu. 2020.
\newblock Exploring the limits of transfer learning with a unified text-to-text
  transformer.
\newblock \emph{Journal of Machine Learning Research}, 21:1--67.

\bibitem[{Schick and
  Sch{\"u}tze(2021{\natexlab{a}})}]{schick-schutze-2021-exploiting}
Timo Schick and Hinrich Sch{\"u}tze. 2021{\natexlab{a}}.
\newblock \href {https://doi.org/10.18653/v1/2021.eacl-main.20} {Exploiting
  cloze-questions for few-shot text classification and natural language
  inference}.
\newblock In \emph{Proceedings of the 16th Conference of the European Chapter
  of the Association for Computational Linguistics: Main Volume}, pages
  255--269, Online. Association for Computational Linguistics.

\bibitem[{Schick and
  Sch{\"u}tze(2021{\natexlab{b}})}]{schick-schutze-2021-just}
Timo Schick and Hinrich Sch{\"u}tze. 2021{\natexlab{b}}.
\newblock \href {https://doi.org/10.18653/v1/2021.naacl-main.185} {It{'}s not
  just size that matters: Small language models are also few-shot learners}.
\newblock In \emph{Proceedings of the 2021 Conference of the North American
  Chapter of the Association for Computational Linguistics: Human Language
  Technologies}, pages 2339--2352, Online. Association for Computational
  Linguistics.

\bibitem[{Scholkopf et~al.(1997)Scholkopf, Sung, Burges, Girosi, Niyogi,
  Poggio, and Vapnik}]{scholkopf1997comparing}
Bernhard Scholkopf, Kah-Kay Sung, Christopher~JC Burges, Federico Girosi,
  Partha Niyogi, Tomaso Poggio, and Vladimir Vapnik. 1997.
\newblock Comparing support vector machines with gaussian kernels to radial
  basis function classifiers.
\newblock \emph{IEEE transactions on Signal Processing}, 45(11):2758--2765.

\bibitem[{Schr{\"o}der et~al.(2022)Schr{\"o}der, Niekler, and
  Potthast}]{schroder2022revisiting}
Christopher Schr{\"o}der, Andreas Niekler, and Martin Potthast. 2022.
\newblock \href {https://doi.org/10.18653/v1/2022.findings-acl.172} {Revisiting
  uncertainty-based query strategies for active learning with transformers}.
\newblock In \emph{Findings of the Association for Computational Linguistics:
  ACL 2022}, pages 2194--2203, Dublin, Ireland. Association for Computational
  Linguistics.

\bibitem[{Sener and Savarese(2018)}]{coreset}
Ozan Sener and Silvio Savarese. 2018.
\newblock \href {https://openreview.net/forum?id=H1aIuk-RW} {Active learning
  for convolutional neural networks: A core-set approach}.
\newblock In \emph{International Conference on Learning Representations}.

\bibitem[{Settles(2011)}]{settles2011theories}
Burr Settles. 2011.
\newblock From theories to queries: Active learning in practice.
\newblock In \emph{Active Learning and Experimental Design workshop}, pages
  1--18. JMLR Workshop and Conference Proceedings.

\bibitem[{Shen et~al.(2021)Shen, Liu, Liu, Yu, and
  Han}]{shen-etal-2021-training}
Jiaming Shen, Jialu Liu, Tianqi Liu, Cong Yu, and Jiawei Han. 2021.
\newblock \href {https://doi.org/10.18653/v1/2021.findings-acl.219} {Training
  {ELECTRA} augmented with multi-word selection}.
\newblock In \emph{Findings of the Association for Computational Linguistics:
  ACL-IJCNLP 2021}, pages 2475--2486, Online. Association for Computational
  Linguistics.

\bibitem[{Shnarch et~al.(2022)Shnarch, Gera, Halfon, Dankin, Choshen, Aharonov,
  and Slonim}]{shnarch-etal-2022-cluster}
Eyal Shnarch, Ariel Gera, Alon Halfon, Lena Dankin, Leshem Choshen, Ranit
  Aharonov, and Noam Slonim. 2022.
\newblock \href {https://aclanthology.org/2022.acl-long.526} {Cluster {\&}
  tune: {B}oost cold start performance in text classification}.
\newblock In \emph{Proceedings of the 60th Annual Meeting of the Association
  for Computational Linguistics (Volume 1: Long Papers)}, pages 7639--7653,
  Dublin, Ireland. Association for Computational Linguistics.

\bibitem[{Socher et~al.(2013)Socher, Perelygin, Wu, Chuang, Manning, Ng, and
  Potts}]{socher-etal-2013-recursive}
Richard Socher, Alex Perelygin, Jean Wu, Jason Chuang, Christopher~D. Manning,
  Andrew Ng, and Christopher Potts. 2013.
\newblock \href {https://aclanthology.org/D13-1170} {Recursive deep models for
  semantic compositionality over a sentiment treebank}.
\newblock In \emph{Proceedings of the 2013 Conference on Empirical Methods in
  Natural Language Processing}, pages 1631--1642. Association for Computational
  Linguistics.

\bibitem[{Su et~al.(2022)Su, Kasai, Wu, Shi, Wang, Xin, Zhang, Ostendorf,
  Zettlemoyer, Smith et~al.}]{su2022selective}
Hongjin Su, Jungo Kasai, Chen~Henry Wu, Weijia Shi, Tianlu Wang, Jiayi Xin, Rui
  Zhang, Mari Ostendorf, Luke Zettlemoyer, Noah~A Smith, et~al. 2022.
\newblock Selective annotation makes language models better few-shot learners.
\newblock \emph{arXiv preprint arXiv:2209.01975}.

\bibitem[{Tam et~al.(2021)Tam, R.~Menon, Bansal, Srivastava, and
  Raffel}]{tam-etal-2021-improving}
Derek Tam, Rakesh R.~Menon, Mohit Bansal, Shashank Srivastava, and Colin
  Raffel. 2021.
\newblock \href {https://doi.org/10.18653/v1/2021.emnlp-main.407} {Improving
  and simplifying pattern exploiting training}.
\newblock In \emph{Proceedings of the 2021 Conference on Empirical Methods in
  Natural Language Processing}, pages 4980--4991, Online and Punta Cana,
  Dominican Republic. Association for Computational Linguistics.

\bibitem[{Vinh et~al.(2010)Vinh, Epps, and Bailey}]{ari}
Nguyen~Xuan Vinh, Julien Epps, and James Bailey. 2010.
\newblock Information theoretic measures for clusterings comparison: Variants,
  properties, normalization and correction for chance.
\newblock \emph{The Journal of Machine Learning Research}, 11:2837--2854.

\bibitem[{Wang et~al.(2021{\natexlab{a}})Wang, Lian, and
  Yu}]{wang2021unsupervised}
Xudong Wang, Long Lian, and Stella~X Yu. 2021{\natexlab{a}}.
\newblock Unsupervised data selection for data-centric semi-supervised
  learning.
\newblock \emph{arXiv preprint arXiv:2110.03006}.

\bibitem[{Wang et~al.(2021{\natexlab{b}})Wang, Mukherjee, Liu, Gao, Awadallah,
  and Gao}]{wang2021list}
Yaqing Wang, Subhabrata Mukherjee, Xiaodong Liu, Jing Gao, Ahmed~Hassan
  Awadallah, and Jianfeng Gao. 2021{\natexlab{b}}.
\newblock List: Lite self-training makes efficient few-shot learners.
\newblock \emph{arXiv preprint arXiv:2110.06274}.

\bibitem[{Wolf et~al.(2020)Wolf, Debut, Sanh, Chaumond, Delangue, Moi, Cistac,
  Rault et~al.}]{huggingface}
Thomas Wolf, Lysandre Debut, Victor Sanh, Julien Chaumond, Clement Delangue,
  Anthony Moi, Pierric Cistac, Tim Rault, et~al. 2020.
\newblock \href {https://doi.org/10.18653/v1/2020.emnlp-demos.6} {Transformers:
  State-of-the-art natural language processing}.
\newblock In \emph{Proceedings of the 2020 Conference on Empirical Methods in
  Natural Language Processing: System Demonstrations}, pages 38--45, Online.
  Association for Computational Linguistics.

\bibitem[{Xia et~al.(2022)Xia, Artetxe, Du, Chen, and
  Stoyanov}]{xia2022prompting}
Mengzhou Xia, Mikel Artetxe, Jingfei Du, Danqi Chen, and Ves Stoyanov. 2022.
\newblock Prompting electra: Few-shot learning with discriminative pre-trained
  models.
\newblock \emph{arXiv preprint arXiv:2205.15223}.

\bibitem[{Xie et~al.(2020)Xie, Dai, Hovy, Luong, and Le}]{xie2020unsupervised}
Qizhe Xie, Zihang Dai, Eduard Hovy, Thang Luong, and Quoc Le. 2020.
\newblock Unsupervised data augmentation for consistency training.
\newblock \emph{Advances in Neural Information Processing Systems}, 33.

\bibitem[{Xu et~al.(2023)Xu, Yu, Cui, Kan, Zhu, Ho, Zhang, and
  Yang}]{xu2023neighborhood}
Ran Xu, Yue Yu, Hejie Cui, Xuan Kan, Yanqiao Zhu, Joyce~C. Ho, Chao Zhang, and
  Carl Yang. 2023.
\newblock Neighborhood-regularized self-training for learning with few labels.
\newblock In \emph{Proceedings of the Thirty-Seventh AAAI Conference on
  Artificial Intelligence}.

\bibitem[{Yao et~al.(2022)Yao, Dong, Zhang, Zhang, Xie, Liu, Lin, Sun, and
  Wang}]{yao2022prompt}
Yuan Yao, Bowen Dong, Ao~Zhang, Zhengyan Zhang, Ruobing Xie, Zhiyuan Liu, Leyu
  Lin, Maosong Sun, and Jianyong Wang. 2022.
\newblock \href {https://doi.org/10.18653/v1/2022.findings-acl.273} {Prompt
  tuning for discriminative pre-trained language models}.
\newblock In \emph{Findings of the Association for Computational Linguistics:
  ACL 2022}, pages 3468--3473, Dublin, Ireland. Association for Computational
  Linguistics.

\bibitem[{Yu et~al.(2022)Yu, Kong, Zhang, Zhang, and
  Zhang}]{yu-etal-2022-actune}
Yue Yu, Lingkai Kong, Jieyu Zhang, Rongzhi Zhang, and Chao Zhang. 2022.
\newblock \href {https://doi.org/10.18653/v1/2022.naacl-main.102} {{A}c{T}une:
  Uncertainty-based active self-training for active fine-tuning of pretrained
  language models}.
\newblock In \emph{Proceedings of the 2022 Conference of the North American
  Chapter of the Association for Computational Linguistics: Human Language
  Technologies}, pages 1422--1436, Seattle, United States. Association for
  Computational Linguistics.

\bibitem[{Yu et~al.(2021)Yu, Zuo, Jiang, Ren, Zhao, and
  Zhang}]{yu-etal-2021-fine}
Yue Yu, Simiao Zuo, Haoming Jiang, Wendi Ren, Tuo Zhao, and Chao Zhang. 2021.
\newblock \href {https://doi.org/10.18653/v1/2021.naacl-main.84} {Fine-tuning
  pre-trained language model with weak supervision: A contrastive-regularized
  self-training approach}.
\newblock In \emph{Proceedings of the 2021 Conference of the North American
  Chapter of the Association for Computational Linguistics: Human Language
  Technologies}, pages 1063--1077, Online. Association for Computational
  Linguistics.

\bibitem[{Yuan et~al.(2020)Yuan, Lin, and Boyd-Graber}]{yuan-etal-2020-cold}
Michelle Yuan, Hsuan-Tien Lin, and Jordan Boyd-Graber. 2020.
\newblock \href {https://doi.org/10.18653/v1/2020.emnlp-main.637} {Cold-start
  active learning through self-supervised language modeling}.
\newblock In \emph{Proceedings of the 2020 Conference on Empirical Methods in
  Natural Language Processing (EMNLP)}, pages 7935--7948, Online. Association
  for Computational Linguistics.

\bibitem[{Zhang et~al.(2022{\natexlab{a}})Zhang, Hsieh, Yu, Zhang, and
  Ratner}]{zhang2022survey}
Jieyu Zhang, Cheng-Yu Hsieh, Yue Yu, Chao Zhang, and Alexander Ratner.
  2022{\natexlab{a}}.
\newblock A survey on programmatic weak supervision.
\newblock \emph{arXiv preprint arXiv:2202.05433}.

\bibitem[{Zhang and Plank(2021)}]{zhang-plank-2021-cartography-active}
Mike Zhang and Barbara Plank. 2021.
\newblock \href {https://doi.org/10.18653/v1/2021.findings-emnlp.36}
  {Cartography active learning}.
\newblock In \emph{Findings of the Association for Computational Linguistics:
  EMNLP 2021}, pages 395--406, Punta Cana, Dominican Republic. Association for
  Computational Linguistics.

\bibitem[{Zhang et~al.(2022{\natexlab{b}})Zhang, Li, Chen, Deng, Bi, Tan,
  Huang, and Chen}]{zhang2022differentiable}
Ningyu Zhang, Luoqiu Li, Xiang Chen, Shumin Deng, Zhen Bi, Chuanqi Tan, Fei
  Huang, and Huajun Chen. 2022{\natexlab{b}}.
\newblock \href {https://openreview.net/forum?id=ek9a0qIafW} {Differentiable
  prompt makes pre-trained language models better few-shot learners}.
\newblock In \emph{International Conference on Learning Representations}.

\bibitem[{Zhang et~al.(2022{\natexlab{c}})Zhang, Yu, Shetty, Song, and
  Zhang}]{zhang2022prompt}
Rongzhi Zhang, Yue Yu, Pranav Shetty, Le~Song, and Chao Zhang.
  2022{\natexlab{c}}.
\newblock \href {https://aclanthology.org/2022.acl-long.55} {Prompt-based rule
  discovery and boosting for interactive weakly-supervised learning}.
\newblock In \emph{Proceedings of the 60th Annual Meeting of the Association
  for Computational Linguistics (Volume 1: Long Papers)}, pages 745--758,
  Dublin, Ireland. Association for Computational Linguistics.

\bibitem[{Zhang et~al.(2020{\natexlab{a}})Zhang, Yu, and
  Zhang}]{zhang2020seqmix}
Rongzhi Zhang, Yue Yu, and Chao Zhang. 2020{\natexlab{a}}.
\newblock \href {https://doi.org/10.18653/v1/2020.emnlp-main.691} {{S}eq{M}ix:
  Augmenting active sequence labeling via sequence mixup}.
\newblock In \emph{Proceedings of the 2020 Conference on Empirical Methods in
  Natural Language Processing (EMNLP)}, pages 8566--8579, Online. Association
  for Computational Linguistics.

\bibitem[{Zhang et~al.(2020{\natexlab{b}})Zhang, Wu, Katiyar, Weinberger, and
  Artzi}]{zhang2020revisiting}
Tianyi Zhang, Felix Wu, Arzoo Katiyar, Kilian~Q Weinberger, and Yoav Artzi.
  2020{\natexlab{b}}.
\newblock Revisiting few-sample bert fine-tuning.
\newblock \emph{arXiv preprint arXiv:2006.05987}.

\bibitem[{Zhang et~al.(2015)Zhang, Zhao, and LeCun}]{zhang2015character}
Xiang Zhang, Junbo Zhao, and Yann LeCun. 2015.
\newblock Character-level convolutional networks for text classification.
\newblock \emph{Advances in neural information processing systems},
  28:649--657.

\bibitem[{Zhao et~al.(2021)Zhao, Wallace, Feng, Klein, and
  Singh}]{zhao2021calibrate}
Zihao Zhao, Eric Wallace, Shi Feng, Dan Klein, and Sameer Singh. 2021.
\newblock Calibrate before use: Improving few-shot performance of language
  models.
\newblock In \emph{International Conference on Machine Learning}, pages
  12697--12706. PMLR.

\end{thebibliography}
\bibliographystyle{acl_natbib}

\clearpage
\appendix
\begin{table*}[t!]
\centering
\label{tab:data}
\resizebox{\linewidth}{!}{
\begin{tabular}{llccccll}
\toprule
\textbf{Dataset} & \textbf{Domain} & \textbf{Classes $c$} & \textbf{$\#$Unlabeled} & \textbf{\#Test} & \textbf{Type} & \textbf{Template} & \textbf{Label words} \\ \midrule
IMDB& Movie Review & 2 & 25k & 25k & sentiment & $\langle S\rangle$. It was \texttt{[MASK]}. & terrible, great \\
Yelp-full & Restaurant Review & 2 & 560k & 38k & sentiment & $\langle S\rangle$. It was \texttt{[MASK]}. & terrible, bad, okay, good, great \\
AG News & News & 4 & 120k & 7.6k & News Topic & \texttt{[MASK]} News: $\langle S\rangle$ & World, Sports, Business, Tech \\
Yahoo! Answers & Web QA & 10 & 300k & 60k & QA Topic & [Category: \texttt{[MASK]}] $\langle S\rangle$ & Society, Science, Health, Education, Computer,  \\ 
& & & & & & & Sports, Business, Entertainment, Relationship, Politics\\
DBPedia & Wikipedia Text & 14 & 420k & 70k & Wikipedia Topic & $\langle T\rangle\langle S\rangle. \langle T\rangle$ is a  \texttt{[MASK]}]& Company, School, Artist, Athlete, Politics,  \\ 
& & & & & & & Transportation, Building, Mountain, Village,  \\
& & & & & & & Animal, Plant, Album, Film, Book\\

TREC & Web Text & 6 & 5k & 0.6k & Question Topic & $\langle S\rangle$. It was \texttt{[MASK]}. &  Expression, Entity, Description, Human, Location, Number \\
\bottomrule
\end{tabular}
}
\caption{Statistics, manual templates, and label words used in our experiments. $c$: number of classes.}
\label{app:dataset}
\end{table*}

\section{Datasets Details}

\subsection{Datasets for the Main Experiment}

The seven benchmarks in our experiments are all publicly available. Below are the links to downloadable versions of these datasets.


\noindent $\diamond$ \textbf{IMDB}: We use the datasets from \url{https://huggingface.co/datasets/imdb}.

\noindent $\diamond$ \textbf{Yelp-full}: Dataset is available at \url{https://github.com/yumeng5/WeSHClass/tree/master/yelp}.

\noindent $\diamond$ \textbf{AG News}: Dataset is available at \url{https://huggingface.co/datasets/ag_news}.

\noindent $\diamond$ \textbf{Yahoo! Answers}: Dataset is available at \url{https://huggingface.co/datasets/yahoo_answers_topics}.

\noindent $\diamond$ \textbf{DBPedia}: Dataset is available at \url{https://huggingface.co/datasets/dbpedia_14}.

\noindent $\diamond$ \textbf{TREC}: Dataset is available at \url{https://huggingface.co/datasets/trec}. Note that we only use the coarse-grained class labels.

\subsection{Train/Test Split}
For all the datasets, we use the original train/test split from the web. 
To keep the size of the development set small~\cite{bragg2021flex}, we randomly sample 32 data from the original training set as the development set, and regard the remaining as the unlabeled set $\cD_u$.

\begin{table}[t]
\centering 
\renewcommand\arraystretch{0.85}
\fontsize{7}{9}\selectfont 
\setlength{\tabcolsep}{0.5em}
\resizebox{0.99\linewidth}{!}{%
\begin{tabular}{l|c|c|c|c}
\toprule
\bf Datasets &Zero-shot Acc.  &Zero-shot Acc.  &NMI  &  ARI \\
& (in \%) & after UC. (in \%) & \\
\midrule
IMDB & 73.29 & 83.13 &0.249 & 0.319 \\
Yelp-full &  32.76&	38.62 & 0.079 &0.056 \\ 
AG News & 81.43& 80.66	& 0.443	&0.432 \\
Yahoo! Answers & 44.13 & 47.55 & 0.274 &0.193 \\
DBPedia  & 73.78& 81.13 & 0.717 &0.595 \\
TREC  & 35.69& 38.51 & 0.111 & 0.088 \\
\bottomrule
\end{tabular}
}
\caption{Quality of Prompts and SimCSE embeddings for six datasets used in our experiments. }
\label{app:quality}
\end{table}

\subsection{Datasets for OOD Evaluation}
\label{app:ood}
We use 3 datasets as OOD tasks for evaluating {\ours} and baselines. The details are listed as belows.

\noindent $\diamond$ \textbf{SST-2}~\cite{socher-etal-2013-recursive}\footnote{\url{https://huggingface.co/datasets/sst2}} is another movie review sentiment analysis dataset. 
The key difference between the SST-2 and IMDB datasets is that they consist of movie reviews with different lengths. We use the original validation set (containing 872 samples) for evaluation.

\noindent $\diamond$ \textbf{IMDB Contrast Set (IMDB-CS)}~\cite{gardner2020evaluating}\footnote{\url{https://github.com/allenai/contrast-sets/tree/main/IMDb}} and
\textbf{IMDB Counterfactually Augmented Dataset (IMDB-CAD)}~\cite{Kaushik2020Learning}\footnote{\url{https://github.com/acmi-lab/counterfactually-augmented-data/tree/master/sentiment}} are two challenging sentiment analysis datasets (both of them contain 488 examples) which can be used to evaluate a model’s true linguistic capabilities more accurately.
Specifically, for IMDB-CS, NLP researchers  creates contrast sets via manually change the ground-truth label of the test instances in a small but semantically meaningful way.
For IMDB-CAD, annotators are required to make minor changes to examples in the original IMDB dataset to flip the sentiment labels, without changing  the majority of contents.

\subsection{Prompt Format}
For these datasets, we directly use \emph{manual prompts} that have been used in previous works~\cite{schick-schutze-2021-exploiting,gao2021making,hu2022knowledgeable}. 
The details of the prompts used in our experiments is listed in Table \ref{app:dataset}.

\subsection{The Quality of Prompts and SimCSE Embeddings}
We list the quality of prompts as well as SimCSE embeddings in this part.
From prompts, we use the \emph{zero-shot accuracy} for the unlabeled data as the quality measure.
From embeddings, 
we perform clustering to evaluate the quality of the SimCSE embeddings.
We use K-Means as the clustering method, and use two metrics, namely Normalized Mutual Information (NMI), and Adjusted Rand Index (ARI)~\cite{ari} for evaluation. For these metrics, higher value indicates better quality. The results are shown in Table~\ref{app:quality}.
We observe that although the quality of these two terms are high for some tasks such as IMDB and AG News, for other tasks, the embeddings are less discriminative and the prompts are less accurate. These pose specific challenges for {\ours} to select most useful data with noisy prompt-based predictions with the imperfect embeddings.

\section{Experiment Setups}
\label{appendix:setup}
\subsection{Main Experiment Setups}
In experiments, both our method and baselines are run with 5 different random seed and the result is based on the average performance on them.  
We have show both the mean and the standard deviation of the performance in our experiment sections. 
\subsection{Experiment Setups for Prompt-based Few-shot Learning}
We mainly use the pipeline in LM-BFF~\cite{gao2021making} for prompt-based learning. For both {\ours} and baselines, we use the prompt defined in Table \ref{app:dataset} to fine-tune PLMs.  We use OpenPrompt toolkit \cite{openprompt} for implementation and use RoBERTa-base as the backbone for prompt-based learning.

\subsection{Experiment Setups for Semi-supervised Learning}
\label{app:ssl}
For semi-supervised learning,  we mainly adopt Unsupervised Data Augmentation (UDA)~\cite{xie2020unsupervised} and self-training~\cite{du-etal-2021-self} as two examples. 
The main idea of  \emph{UDA} is leveraging data augmentation techniques (TF-IDF word replacement or back translation) with the consistency-based loss for unlabeled data to improve the model performance. 
Since we do not have access to TPU service and need to use smaller amount
of unlabeled data, we implement UDA on our own. 
For \emph{self-training}, it  generates pseudo labels on unlabeled data, and encourages models to output confident predictions on these data.   
Please refer to the original papers for the details of these methods. 

\subsection{Experiment Setups for Standard Multi-round Active Learning}
\label{app:multi_round_al}

For standard multi-round active learning, we follow the standard multi-round active learning pipelines introduced in~\cite{margatina2021active,yuan-etal-2020-cold}, but in the beginning round, \emph{no} initial labeled data is given. 
In each round, we train the PLM \emph{from scratch} to avoid overfitting to the data collected in earlier rounds as observed by \citet{hu2018active}.

\section{Details on Implementations}
\label{appendix:impl}

\subsection{Computational Setups}
Overall we report the results of \textbf{\underline{3240}} BERT fine-tuning runs for main experiments (2 settings × 6 datasets × 3 labeling budgets × 9 methods × 10 repetitions). The computing infrastructure used for experiments are listed as follows. \\
\textbf{System}: Ubuntu 18.04.3 LTS; Python 3.8; Pytorch 1.10. \\
\textbf{CPU}: Intel(R) Core(TM) i7-5930K CPU @ 3.50GHz. \\
\textbf{GPU}: NVIDIA A5000. \\

\begin{table*}[t]
    \renewcommand\arraystretch{0.9}
	\begin{center}
		\begin{tabular}{c|c|c|c|c|c|c}
			\toprule 
			\bf Hyper-parameter &\bf IMDB & \bf  Yelp-full & \bf AG News & \bf Yahoo! & \bf DBPedia  & \bf TREC \\ \midrule 
			 Maximum Tokens  & 256 & 256 & 128 & 128 & 128 & 64  \\ \hline 
			 Learning Rate & 2e-5  & 2e-5  & 5e-5  & 5e-5 & 1e-5 & 2e-5     \\ \hline
             $k$ & \multicolumn{5}{c|}{1000} & 50 \\ \hline
             $\rho$ &0.05 & 0.1 & 0.1 & 0.1 & 0.1 & 0.1 \\ \hline
             $\gamma$ & 0.3 & 0.3 &  0.5 & 0.3 & 0.1 & 0.3  \\ \hline
             $\beta$ &0.5 & 5 & 0.5 & 1 & 5 & 5  \\ \hline
			 $m$ & \multicolumn{6}{c}{0.5} \\ 
			 \bottomrule
		\end{tabular}
	\end{center}
	\vspace{-1ex}
	\caption{Hyper-parameter configurations. Note that we only keep certain number of tokens.}
	\vspace{-1ex}
	\label{tab:hyperparameter}
\end{table*}

\subsection{Number of Parameters}
In our main experiments, {\ours} and all baselines use RoBERTa-base~\cite{liu2019roberta} with a task-specific classification head on the top as the backbone, which contains 125M trainable parameters. We do not introduce any other parameters in our experiments.


\subsection{Implementations of Baselines}
For \textbf{Random}, \textbf{Uncertainty}, \textbf{BERT-KM}, \textbf{Margin-KM}, we implement them by ourselves.
For other baselines, we run the experiments based on the implementations on the web. We list the link for the implementations as belows:

\noindent $\diamond$ \textbf{Coreset}:  \url{https://github.com/google/active-learning/tree/master/sampling_methods}.

\noindent $\diamond$ \textbf{ALPS}:  \url{https://github.com/forest-snow/alps}.

\noindent $\diamond$ \textbf{CAL}:  \url{https://github.com/mourga/contrastive-active-learning}.

\noindent $\diamond$ \textbf{TPC}:  \url{https://github.com/avihu111/TypiClust}.

\subsection{Hyper-parameters for Model Training}
\label{appendix:hyper}
We use AdamW \cite{loshchilov2018adamw} as the optimizer, and choose the learning rate from $\{1\times 10^{-5}, 2\times 10^{-5}, 5\times 10^{-5}\}$, the batch size from $\{4, 8, 16\}$, and set 
the number of training epochs to $15$ for both fine-tuning,  prompt-based few-shot learning, and multi-round active learning. 

For semi-supervised learning, we initialize the model with the RoBERTa-base fine-tuned on the acquired labeled data (based on different data selection strategies).
Then, we set the  batch size for unlabeled data to $32$, and choose the learning rate from $\{1\times 10^{-6}, 5\times 10^{-6}, 1\times 10^{-5}\}$ since we empirically find that smaller learning rates lead to the better training stability.  
We use the model with best performance on the development set to determine the best set of parameter for testing.

\subsection{Hyper-parameters for AL Implementation}
\label{appendix:ours}
{\ours} introduces several
hyper-parameters including $k$ in Eq.~\ref{eq:k_support}, $K$ for calculating $\cX_{\text{KNN}}(x)$ ,$K'$ for calculating $\cX_{\text{c-KNN}}(x)$, $\beta, \gamma, m$ in Eq.~\ref{eq:init}, $\rho$ in Eq.~\ref{eq:prop}, but most of them are keep fixed  during our experiments, thus it does not require heavy hyper-parameter tuning.

In our experiments, we keep $K'=10$, $K=50$,  $m=0.5$ for all datasets.
For other parameters, we use a grid search to find the optimal setting for each datasets.
We search $\rho$ from $\{0.01, 0.05, 0.1, 1\}$, $\beta$ from $\{0.5, 1, 5, 10\}$, $\gamma$ from $\{0.1, 0.3, 0.5\}$. 
All results are reported as the average over \textbf{ten runs}.
The number for hyper-parameters we use are shown in Table~\ref{tab:hyperparameter}. 

For other baselines, we follow the exact parameter tuning method mentioned in the original paper for hyperparameter tuning. 
For CAL~\cite{margatina2021active} and TPC~\cite{hacohen2022active}, we tune the number for KNN $k$ from $[5, 10, 20, 50]$ and report the best performance.

\section{Adapting {\ours} to Multi-round AL}
\label{app:pisa_multi}

When applying {\ours} to Multi-round AL, since there exists a warm-start model with a set of labeled data, we directly use the embedding from the warm-start model to generate features and leverage it for uncertainty estimation. 
After that, uncertainty propagation can be directly adopted for estimating the utility of training data. 
For the {\ptr} step, since we already have a smaller number of the labeled samples $\cD_l$, the Eq.~\ref{eq:cknn} can be refined as
\begin{equation}
\cX_{\text{c-KNN},i} =  \text{KNN}(q_i, \cQ \cup \cD_l), 
\label{eq:cknn_multi}
\end{equation}
as we don't want the selected samples to be too close to samples in $\cD_l$. The other steps of {\ptr} are remain unchanged.

\section{Time Complexity of {\ours}}
\begin{table}[t]
\centering
\renewcommand\arraystretch{0.9}
\begin{adjustbox}{max width=0.48\textwidth}
\begin{tabular}{l|c}
\toprule
\multicolumn{1}{c|}{\bf Method}    & \bf Time                     \\ \midrule
Random              & 0.1s  \\ \hline
Uncertainty     & {461s}                  \\ \hline
CAL         & {649s}          \\ \hline
BERT-KM   & {724s}      \\ \hline
Coreset   & {872s}                  \\ \hline
Margin-KM        & {1389s}                  \\ \hline
ALPS         & {682s}                  \\ \hline
TPC & {1448s}                  \\ \hline
{\ours}   &  {1480s}       \\ \bottomrule
\end{tabular}
\end{adjustbox}
\caption{The running time of {\ours} and different baselines on Yahoo! Answers dataset.}
	\label{tab:time}
\end{table}
\label{app:time}

The additional time introduced by {\ours} mainly comes from the KNN step in the uncertainty propagation as well as the K-Means partitioning. However, these operations have been efficiently supported via approximate nearest neighbor search (ANN)~\cite{FAISS}. As a result, {\ours} will not incur excessive computational overhead.

Table \ref{tab:time} exhibits the running time of {\ours} and baselines on the \emph{Yahoo! Answers} dataset for selecting 64 samples. 
Overall, compared with the recent baselines such as TPC~\cite{hacohen2022active} and Margin-KM~\cite{muller2022active}, the additional time introduced is small.
In particular, the uncertainty propagation takes 114 seconds, and the predict-then-propagate step only takes 5 seconds. This verifies that our key designs  do not takes much time and are scalable for large datasets.

\section{Additional Analysis}
In this section, we provide detailed comparison on different data selection strategies, aiming to better understand their relative advantages and disadvantages.
Specifically, we follow the method in \citet{ein-etal-2020-active} and focus on three types of metrics: \emph{class distribution}, \emph{feature diversity}, and  \emph{representativeness}.  
All of these metrics are calculated based on the results with 128 labels as the budget.

\subsection{Class Distribution of the Selected Data}
\label{app:label_dist}
We calculate the class distribution of the selected samples. 
Denote the number of samples selected from each class as $n_1, \ldots, n_c$ where $\sum_{i=1}^{c}n_i=|B|$ ($|B|=128$ in this case), we use two metrics, namely imbalance value and label distribution divergence value to measure the class distribution.
Specifically, imbalance value (IMB) is calculated as 
\begin{equation}
    \operatorname{IMB} = \frac{\max_{i=1,\ldots,c}(n_i)}{\min_{i=1,\ldots,c}(n_i)}. 
\end{equation}
The higher IMB value indicates the more imbalanced distribution. 
Note that when data from one or more classes are totally not sampled, the  IMB value will become \emph{infinity} (+inf). 

As the label distribution of some datasets are imbalanced, we introduce another metrics named label distribution divergence, to calculate the distance between the distribution of ground-truth labels and labels sampled by baselines or our method.
Specifically, denote $p_{i}$ as the frequency of label $i$. Then the label distribution divergence (LDD) is calculated as 
\begin{equation}
    \operatorname{LDD} = \cD_{\operatorname{KL}}\left(q||p\right) = -\sum_{i}q_i \log{(p_i/q_i)}. 
\end{equation}
where $q_i=n_i/|B|$ is equal to the frequency of class $i$ in the selected samples. 
The higher LDD value indicates the more biased sampled distribution from the original distribution. 

Table~\ref{tab:coverage} and \ref{tab:label_kl} show the IMB and LDD value for all methods on six datasets. 
From the results, we find that for uncertainty-based approaches, the corresponding values for these two metrics are very high. 
This indicates that the selected samples are highly imbalanced. 
As there does not exist any startup labels for cold-start data selection, fine-tuning PLMs on such imbalanced data leads to the biased predictions.   
These results explain why the performance of such uncertainty-based methods are extremely poor under cold-start scenarios.
\begin{table*}[!t]
\centering 
\renewcommand\arraystretch{0.9}
\fontsize{7.5}{9.5}\selectfont \setlength{\tabcolsep}{0.5em}
\resizebox{0.95\linewidth}{!}{%
\begin{tabular}{c|c|c|cc|cc|ccc|c}
\toprule
\bf    Task     & $c$ & \bf  Random & \bf Uncertainty & \bf CAL  & \bf BERT-KM &\bf  Coreset & \bf Margin-KM & \bf ALPS & \bf TPC & \bf {\ours} \\ \midrule
{IMDB} & {2} 
&  1.207 &	6.111	&7.000&	1.286&	1.000	&1.133&	1.783&	2.765&	1.286	\\
{Yelp-F} & {5} 
&  1.778 &	3.800	&13.500	&2.000&	6.000&	1.600&	2.833&	5.200&	2.250\\ 
{AG News} & {4} 
&  1.462 &	28.000 &	2.000 &	1.500 &	2.000 &	2.625 &	1.667 &	1.818 & 1.500\\
{Yahoo! Ans.} & {10} 
&  3.000 &	12.000 &	+inf &	2.250 &	7.000 &	10.000 &	5.500 &	3.333 &	5.500 \\
{DBPedia} & {14} 
& 3.500	& +inf&	+inf	&3.500	&9.000&	12.000&	9.000	&9.000&	2.333 \\
{TREC} & {6} 
& 8.000	& 16.000 & 	+inf	& 10.500 & 	+inf & 	18.000 & 	9.500 & 	21.000	& 15.000 \\
\bottomrule
\end{tabular}
}
\caption{The label imbalance value (IMB) of different data selection approaches. The lower value indicates more balanced sampling over classes.}
\label{tab:coverage}
\end{table*}

\begin{table*}[!t]
\centering 
\renewcommand\arraystretch{0.9}
\fontsize{7.5}{9.5}\selectfont \setlength{\tabcolsep}{0.5em}
\resizebox{0.95\linewidth}{!}{%
\begin{tabular}{c|c|c|cc|cc|ccc|c}
\toprule
\bf    Task     & $c$ & \bf  Random & \bf Uncertainty & \bf CAL  & \bf BERT-KM &\bf  Coreset & \bf Margin-KM & \bf ALPS & \bf TPC & \bf {\ours} \\ \midrule
{IMDB} & {2} 
&  0.004 &	0.287&	0.410&	0.008&	0.000 &	0.002 &	0.040 &	0.114 &	0.008 \\
{Yelp-F} & {5} 
&  0.021 &	0.094 &	0.323 &	0.030 &	0.147 &	0.014 &	0.046 &	0.137 &	0.051\\ 
{AG News} & {4} 
&  0.010 &	0.253 &	0.027 &	0.011 &	0.030 &	0.054 &	0.016 &	0.027 &	0.012\\
{Yahoo! Ans.} & {10} 
&  0.039 &	0.172 &	1.223 &	0.046 &	0.170 &	0.150 &	0.101 &	0.098 &	0.090 \\
{DBPedia} & {14} 
&  0.067 &	1.074 &	2.639 &	0.049 &	0.120 &	0.468 &	0.117 &	0.117 &	0.041\\
{TREC} & {6} 
&  0.015 &	0.081 &	1.598 &	0.070 &	0.078 &	0.085 &	0.030 & 0.212 &	0.063\\
\bottomrule
\end{tabular}
}
\caption{The label divergence value (LDD) of different data selection approaches. The lower value indicates more balanced sampling over classes.}
\label{tab:label_kl}
\end{table*}

\begin{table*}[!t]
\centering 
\renewcommand\arraystretch{0.9}
\fontsize{7.5}{9.5}\selectfont \setlength{\tabcolsep}{0.5em}
\resizebox{0.95\linewidth}{!}{%
\begin{tabular}{c|c|c|cc|cc|ccc|cc}
\toprule
\bf    Task     & $c$ & \bf  Random & \bf Uncertainty & \bf CAL  & \bf BERT-KM &\bf  Coreset & \bf Margin-KM & \bf ALPS & \bf TPC & \bf {\ours} w/o {\ptr} & \bf {\ours} \\ \midrule
{IMDB} & {2} 
&  0.646 &	0.647 &	0.603 &	0.687&	0.643&	0.642&	0.647&	0.648 &	0.670 & 0.684\\
{Yelp-F} & {5} 
& 0.645	& 0.626	& 0.587	& 0.685	& 0.456	& 0.626	& 0.680	& 0.677	& 0.681	& 0.685 \\ 
{AG News} & {4} 
& 0.354	& 0.295	& 0.339	& 0.436	& 0.340	& 0.328	& 0.385	& 0.376	& 0.420	& 0.423 \\
{Yahoo! Ans.} & {10} 
& 0.430	& 0.375	& 0.338	& 0.470	& 0.400	& 0.388	& 0.441	& 0.438	& 0.481	& 0.486 \\
{DBPedia} & {14} 
& 0.402	& 0.316	& 0.244	& 0.461	& 0.381	& 0.361	& 0.420	& 0.399	& 0.456	& 0.459 \\
{TREC} & {6} 
& 0.301	& 0.298	& 0.267	& 0.337	& 0.298	& 0.307	& 0.339	& 0.326	& 0.337	& 0.338 \\
\bottomrule
\end{tabular}
}
\caption{The diversity value of different data selection approaches. The higher value indicates higher diversity.}
\label{tab:diversity}
\end{table*}

\begin{table*}[!t]
\centering 
\renewcommand\arraystretch{0.9}
\fontsize{7.5}{9.5}\selectfont \setlength{\tabcolsep}{0.5em}
\resizebox{0.95\linewidth}{!}{%
\begin{tabular}{c|c|c|cc|cc|ccc|cc}
\toprule
\bf    Task     & $c$ & \bf  Random & \bf Uncertainty & \bf CAL  & \bf BERT-KM &\bf  Coreset & \bf Margin-KM & \bf ALPS & \bf TPC & \bf {\ours} w/o {\ptr} & \bf {\ours} \\ \midrule
{IMDB} & {2} 
& 0.742	& 0.749	& 0.685	& 0.759	& 0.735	& 0.717	& 0.731	& 0.764	& 0.802	& 0.806 \\
{Yelp-F} & {5} 
& 0.731	& 0.711	& 0.702	& 0.825	& 0.504	& 0.701	& 0.823	& 0.827	& 0.825	& 0.824 \\ 
{AG News} & {4} 
& 0.656	& 0.601	& 0.683	& 0.733	& 0.646	& 0.624	& 0.716	& 0.816	& 0.742	& 0.749 \\
{Yahoo! Ans.} & {10} 
& 0.667	& 0.614	& 0.670	& 0.680	& 0.621	& 0.605	& 0.678	& 0.784	& 0.782	& 0.787 \\
{DBPedia} & {14} 
& 0.678	& 0.610	& 0.568	& 0.698	& 0.666	& 0.597	& 0.696	& 0.802	& 0.736	& 0.735 \\
{TREC} & {6} 
& 0.435	& 0.435	& 0.424	& 0.518	& 0.442	& 0.442	& 0.520	& 0.553	& 0.509	& 0.512 \\
\bottomrule
\end{tabular}
}
\caption{The representativeness value of different data selection approaches. The higher value indicates better representativeness.}
\label{tab:representativeness}
\end{table*}

\subsection{Feature Diversity of the Selected Data}
\label{app:diversity}
Apart from the categorical-level statistics, we aim to measure the diversity from the feature space. 
For each sample $x$, we use the SimCSE embeddings (used in Section~\ref{sec:uncertainty_estimation}) to obtain its embeddings. 
Then, we follow the method in \cite{ein-etal-2020-active} to calculate the diversity over the samples within the batch $\cQ$ as
\begin{equation}
D(\cQ)=\left(\frac{1}{|U|} \sum_{x_i \in U} \min _{x_j \in \cQ} d\left(x_i, x_j\right)\right)^{-1},
\end{equation}
where $d(x_i, x_j)$ is the Euclidean distance between $x_i$ and $x_j$.

Table \ref{tab:diversity} shows the diversity of different data selection methods. Overall, BERT-KM achieves the best sample diversity, as its objective mainly focuses on promoting the sample diversity. 
In contrast, Coreset method cannot improve the sample diversity for all datasets, as it aims to sample data that are farthest from the already selected instances, which can often be outliers. 
Compared with the other hybrid methods such as ALPS and TPC, {\ours} overall has a better sample diversity. Moreover, {\ptr} strategy further improve the sample diversity on 5 of 6 datasets. This indicates that {\ptr} fulfills the purpose of improving the diversity of the selected examples. 

\subsection{Representativeness of the Selected Data}
\label{app:repre}
The representativeness of samples are defined as their density, which is quantified by the average distance between the example in question and its $10$ most
similar examples based on the \texttt{[CLS]} representations~\cite{ein-etal-2020-active} as 
\begin{equation}
R(x)=\frac{\sum_{x_i \in \text{kNN}(x)} \cos \left(x, x_i\right)}{K}.
\end{equation}

Table \ref{tab:representativeness} shows the score for different methods. {\ours} also achieves comparable performance to the baselines. 

To sum up, the results in above sections indicate that {\ours} strikes a balance between these metrics --- it achieves competitive performance on both diversity and representativeness, which lead to overall better performance under cold-start scenarios.

\section{Additional Experimental Results}

\subsection{The Result with F1 Score for the TREC Dataset}
\label{app:f1}
\begin{table*}[t]
\centering 
\renewcommand\arraystretch{0.9}
\fontsize{7.5}{9.5}\selectfont \setlength{\tabcolsep}{0.5em}
\resizebox{0.97\linewidth}{!}{%
\begin{tabular}{c|cc|c|cc|cc|ccc|c}
\toprule
\bf    Task     & $c$ & $|B|$  & \bf  Random & \bf Uncertainty & \bf CAL  & \bf BERT-KM &\bf  Coreset & \bf Margin-KM & \bf ALPS & \bf TPC & \bf {\ours} \\ \midrule
\multirow{3}{*}{TREC} & \multirow{3}{*}{6}  
& 32 &   42.7\std{1.6} &	34.7\std{1.7}&	13.0\std{4.0} &	\underline{45.4\std{1.8}} &	42.4\std{1.6}&	30.5\std{2.6}&	46.7\std{0.9} & 29.1\std{2.2} &	\bf 48.4\std{1.0} \\
& & 64 & 53.5\std{1.2}&	52.1\std{2.0}& 15.5\std{3.2} &	\underline{64.5\std{1.4}} &	55.5\std{2.0}&	40.3\std{2.3}&	57.1\std{2.4}&	55.6\std{2.0}&	\bf 66.0\std{1.1} \\
& & 128 & 77.4\std{2.0}& 	62.3\std{1.8}& 	44.5\std{2.9}	& \underline{85.6\std{1.1}} & 	74.4\std{1.7} & 	70.3\std{1.0} & 	84.0\std{1.6} & 	67.9\std{2.3} & 	\bf 89.8\std{0.8} \\
\bottomrule
\end{tabular}
}
\caption{The F1 score of the main experiments (few-shot PLM fine-tuning) on the TREC dataset.}
\label{tab:main_f1}
\vspace{-0.1ex}
\end{table*}

\begin{table*}[t]
\centering 
\renewcommand\arraystretch{0.9}
\fontsize{7.5}{9.5}\selectfont \setlength{\tabcolsep}{0.5em}
\resizebox{0.97\linewidth}{!}{%
\begin{tabular}{c|cc|c|cc|cc|ccc|c}
\toprule
\bf    Task     & $c$ & $|B|$  & \bf  Random & \bf Uncertainty & \bf CAL  & \bf BERT-KM &\bf  Coreset & \bf Margin-KM & \bf ALPS & \bf TPC & \bf {\ours} \\ \midrule
\multirow{3}{*}{TREC} & \multirow{3}{*}{6}  
& 32 &  
62.3\std{1.7} &	57.0\std{1.2} &	29.8\std{1.3} &	51.5\std{2.0} &	56.6\std{1.4} &	58.9\std{1.3} &	\underline{62.6\std{1.4}} &	50.1\std{1.2} &	\bf 67.6\std{0.8} \\
& & 64 &  69.6\std{1.1} &	62.7\std{1.4} &	33.8\std{1.7} &	73.0\std{1.2} &	69.2\std{1.5} &	63.5\std{2.0} & 	\bf 75.1\std{1.1} &	66.8\std{1.3} &	\underline{74.2\std{1.4}} \\
& & 128 & 77.3\std{2.4}	& 67.7\std{1.5} & 	55.6\std{4.0} & 	80.8\std{1.6} & 	74.7\std{3.0} & 	66.4\std{2.0} & \underline{83.6\std{2.3}} & 	70.6\std{1.6} & \bf	86.7\std{1.4}
\\
\bottomrule
\end{tabular}
}
\caption{The F1 score of the prompt-based experiments on the TREC dataset.}
\label{tab:main_f1_prompt}
\vspace{-0.1ex}
\end{table*}

The result of the TREC dataset with F1 score as the metric is shown in Table \ref{tab:main_f1} and \ref{tab:main_f1_prompt}. 
In most of the cases, {\ours} still outperforms all the baselines.

\begin{figure*}[!t]
    \centering
         \subfigure[Effect of $\rho$.]{
            \includegraphics[width=0.32\textwidth]{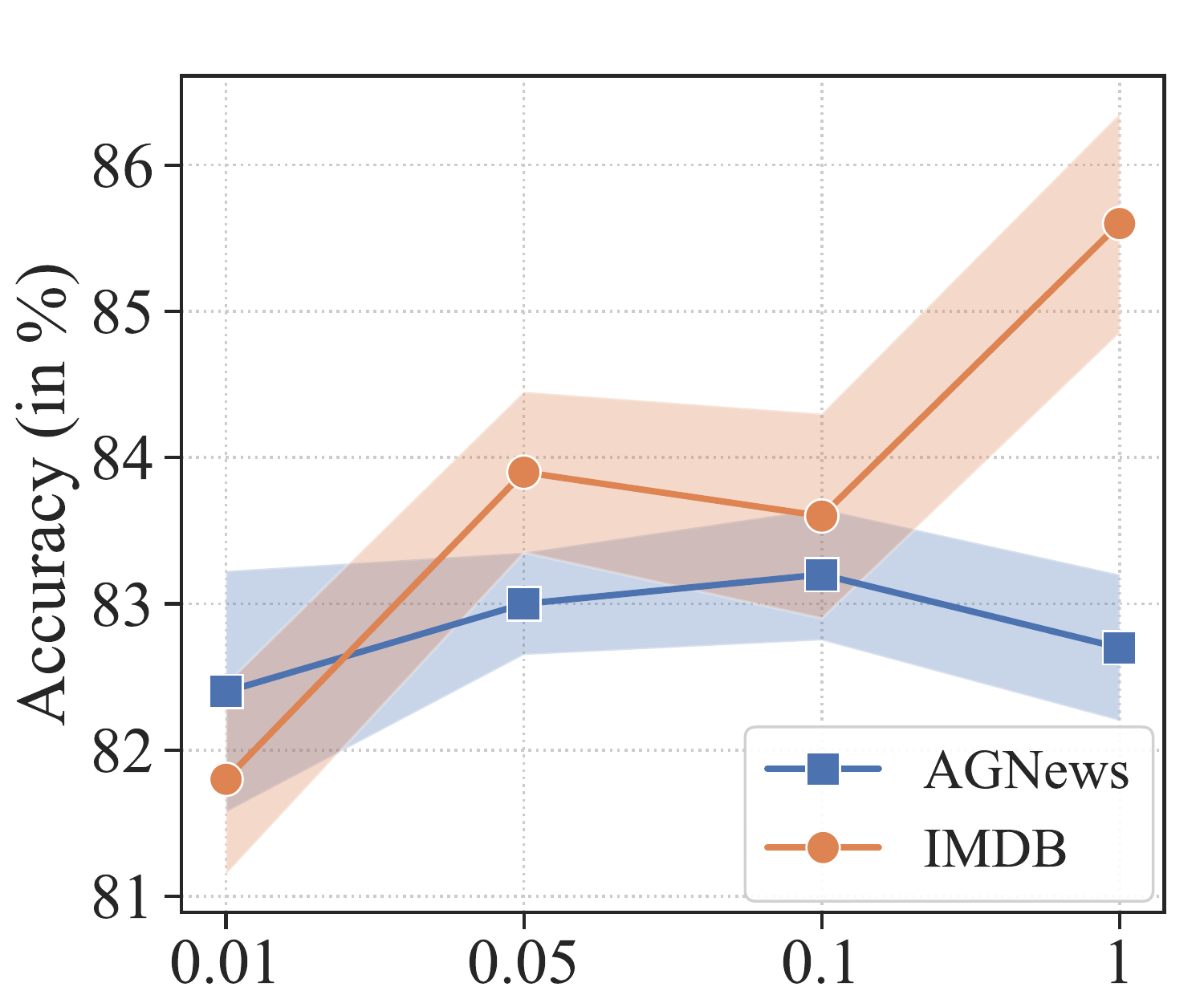}
            \label{fig:rho_agnews}
        }\hspace{-2mm}
        \subfigure[Effect of $\rho$.]{
            \includegraphics[width=0.32\textwidth]{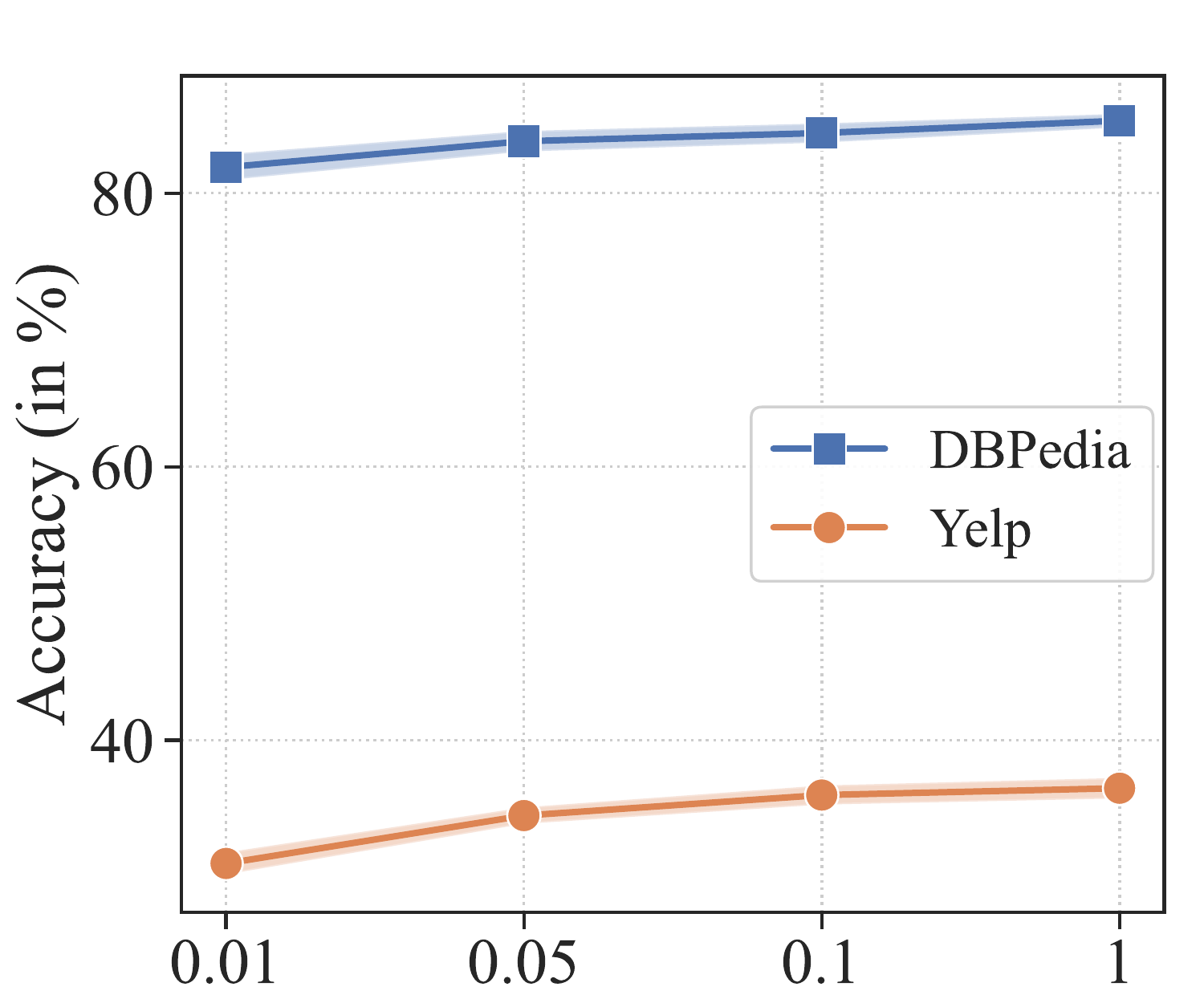}
            \label{fig:rho_yelp}
        }
        \hspace{-2mm}
        \subfigure[Effect of $\beta$.]{
            \includegraphics[width=0.32\textwidth]{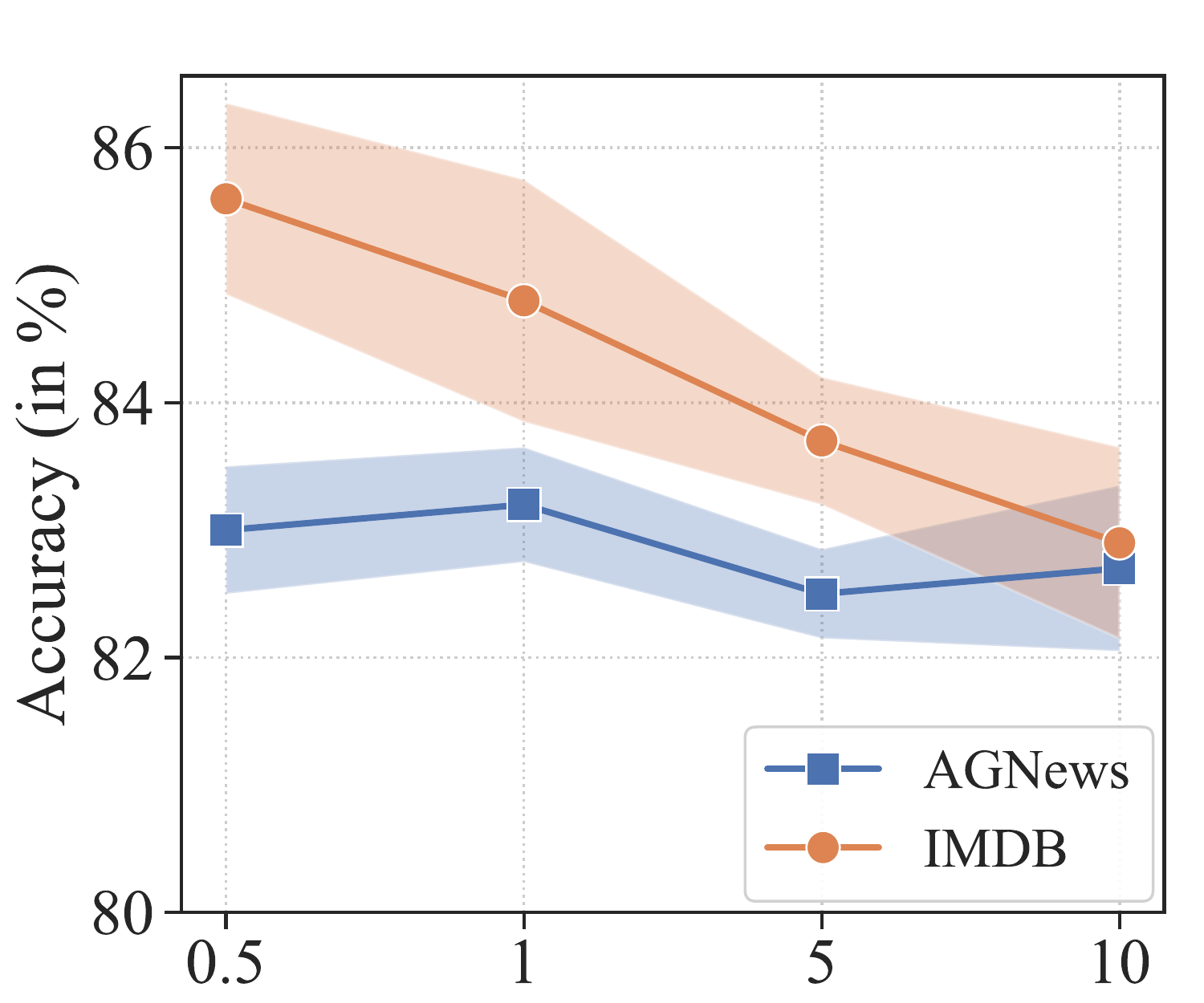}
            \label{fig:beta_imdb}
        }
        \hspace{-2mm}
        \subfigure[Effect of $\beta$.]{
            \includegraphics[width=0.32\textwidth]{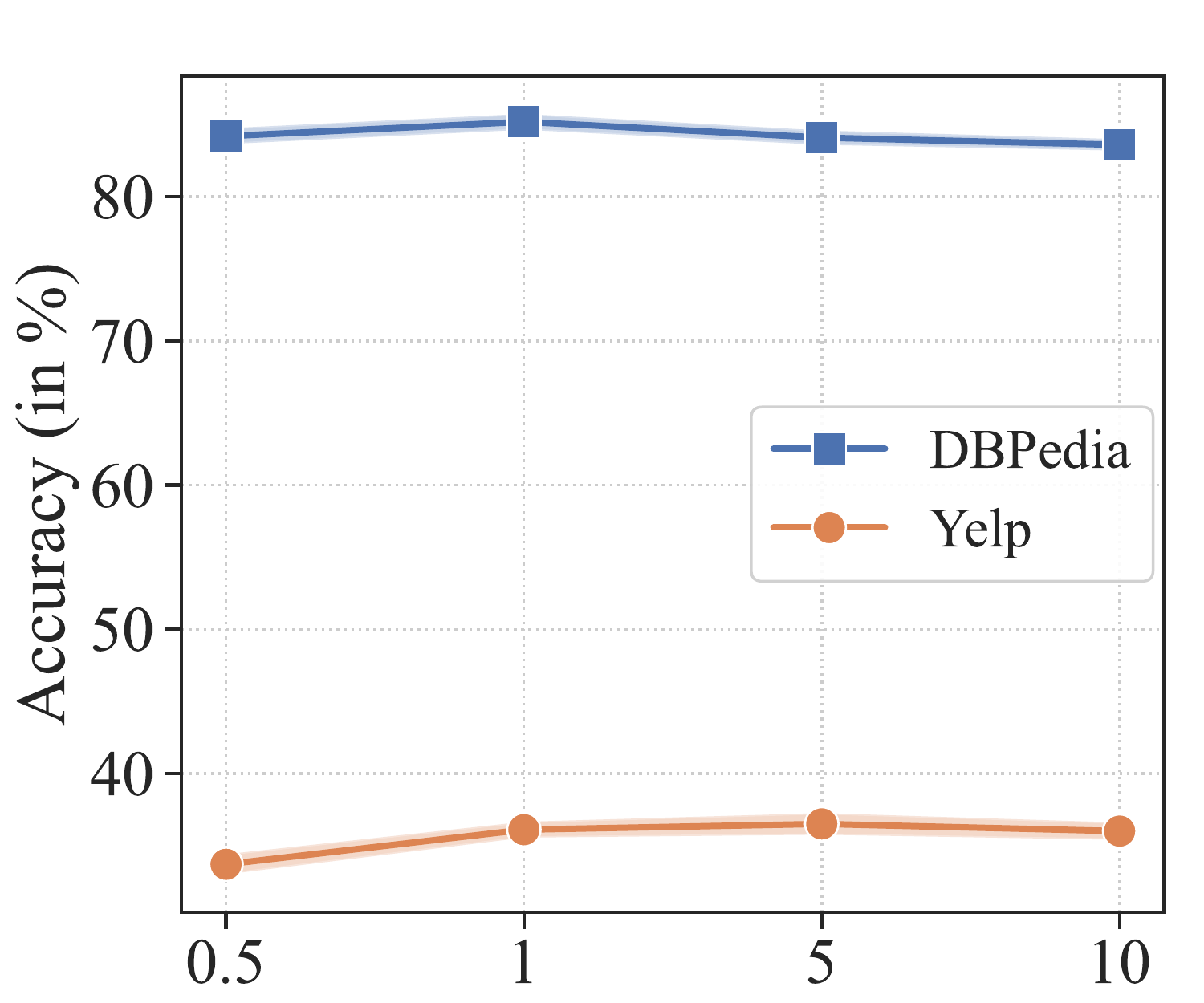}
            \label{fig:beta_yelp}
        }
         \hspace{-2mm}
        \subfigure[Effect of $\gamma$.]{
            \includegraphics[width=0.32\textwidth]{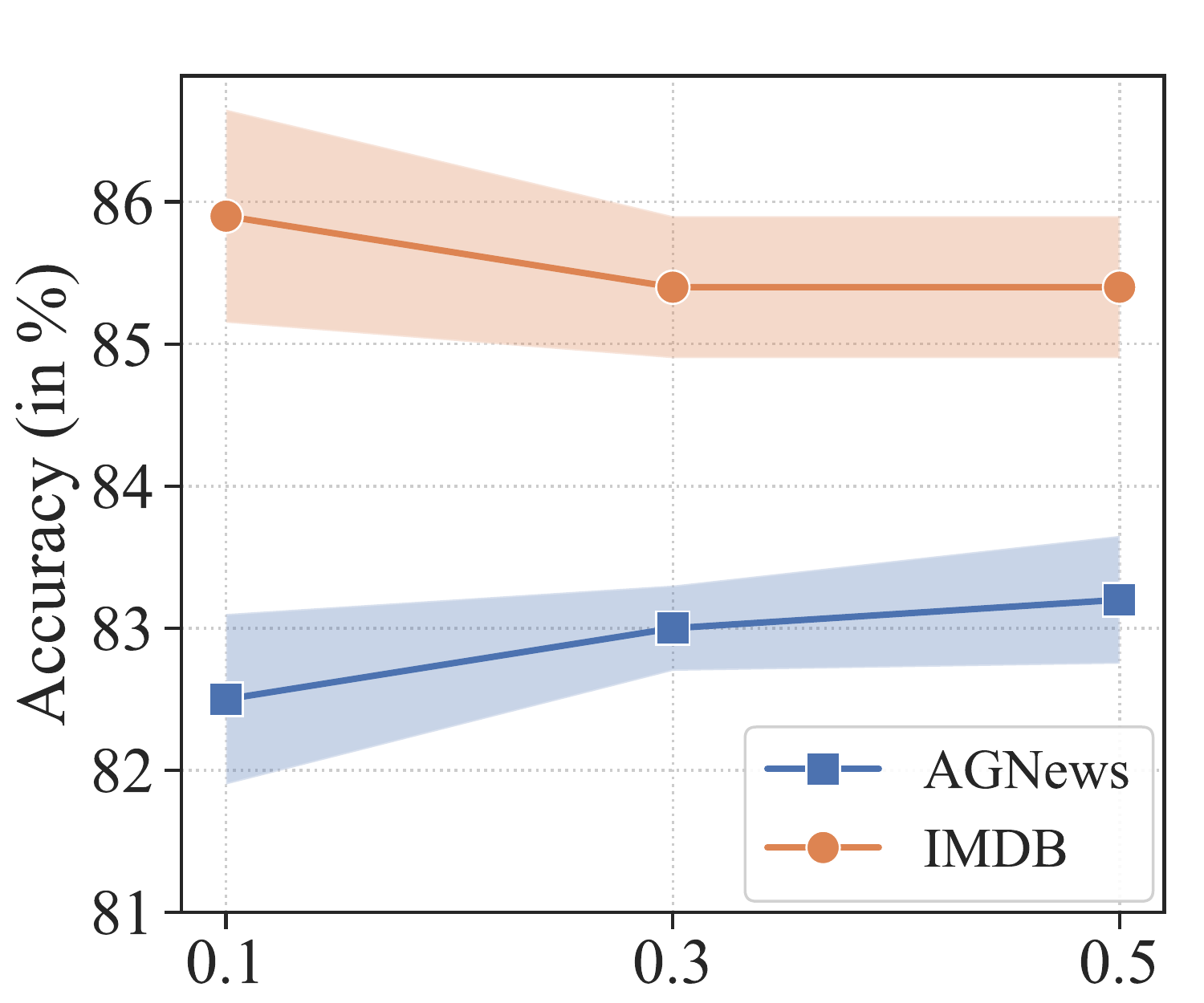}
            \label{fig:gamma_imdb}
        }
        \hspace{-2mm}
        \subfigure[Effect of $\gamma$.]{
            \includegraphics[width=0.32\textwidth]{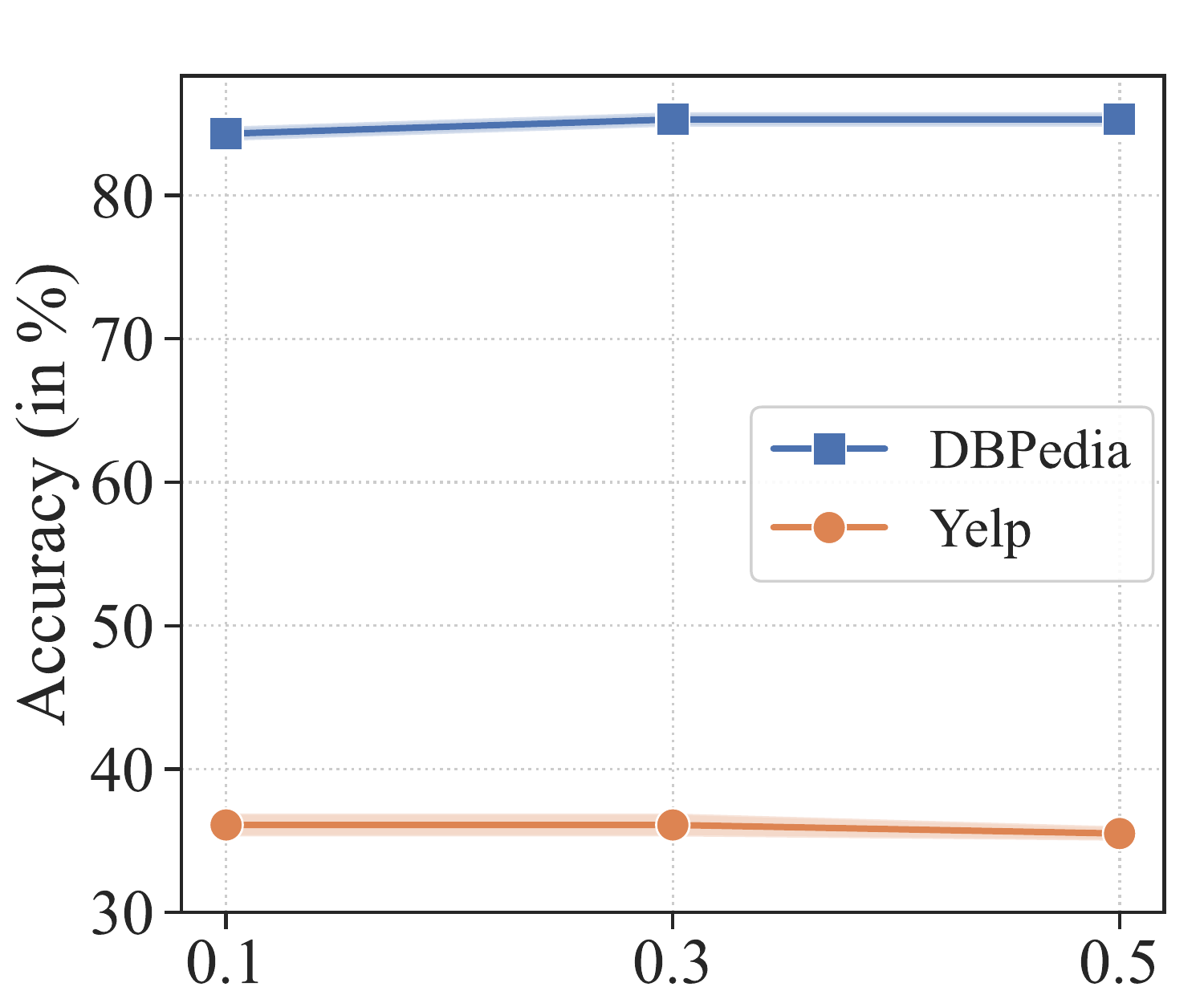}
            \label{fig:gamma_yelp}
        }
        \caption{The additional hyperparameter study on the other datasets.}
        \label{fig:hyper_app}
\end{figure*}

\begin{figure*}[!t]
    \centering
         \subfigure[AG News]{
            \includegraphics[width=0.32\textwidth]{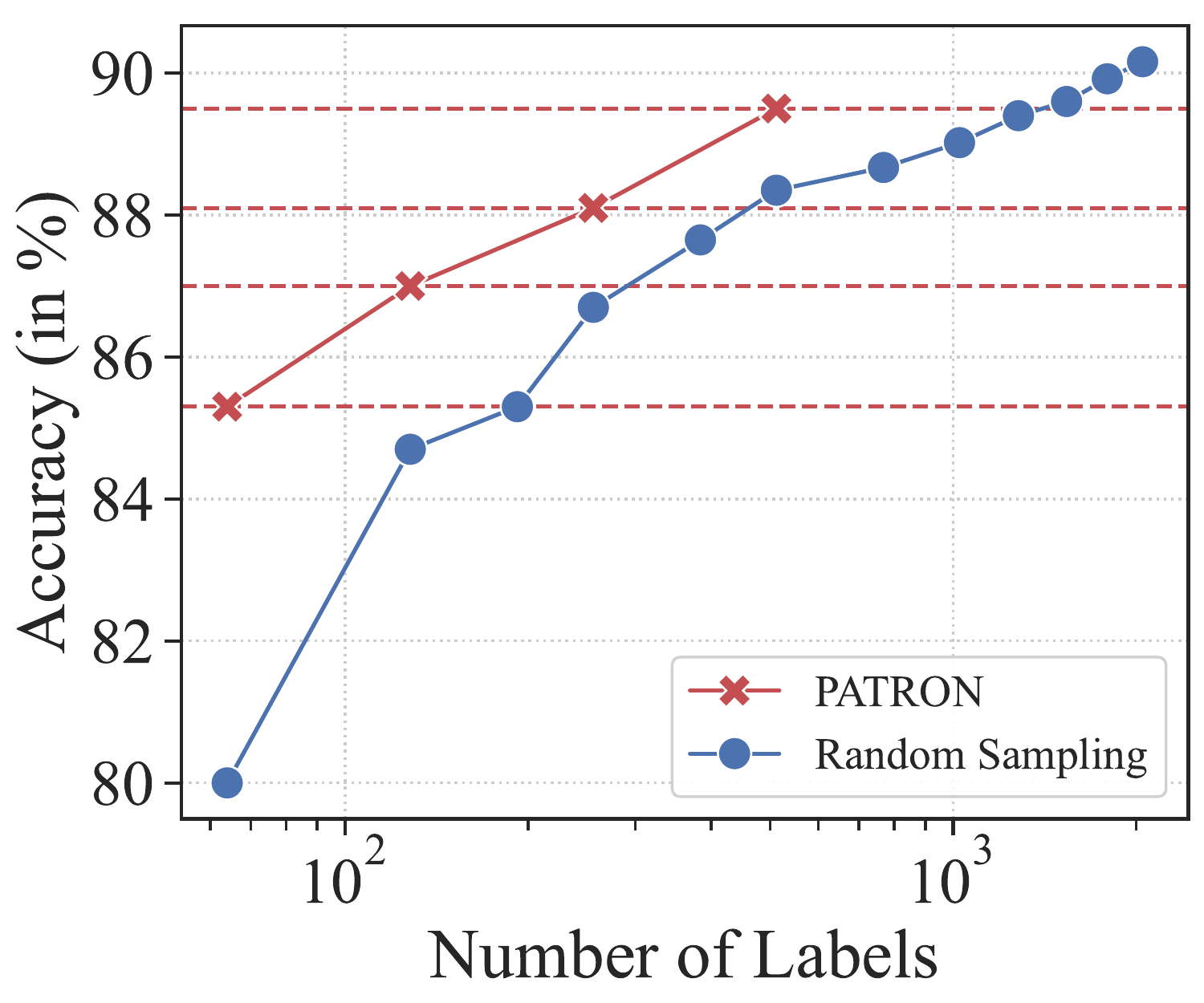}
            \label{fig:log_agnews}
        }\hspace{-2mm}
        \subfigure[Yelp]{
            \includegraphics[width=0.32\textwidth]{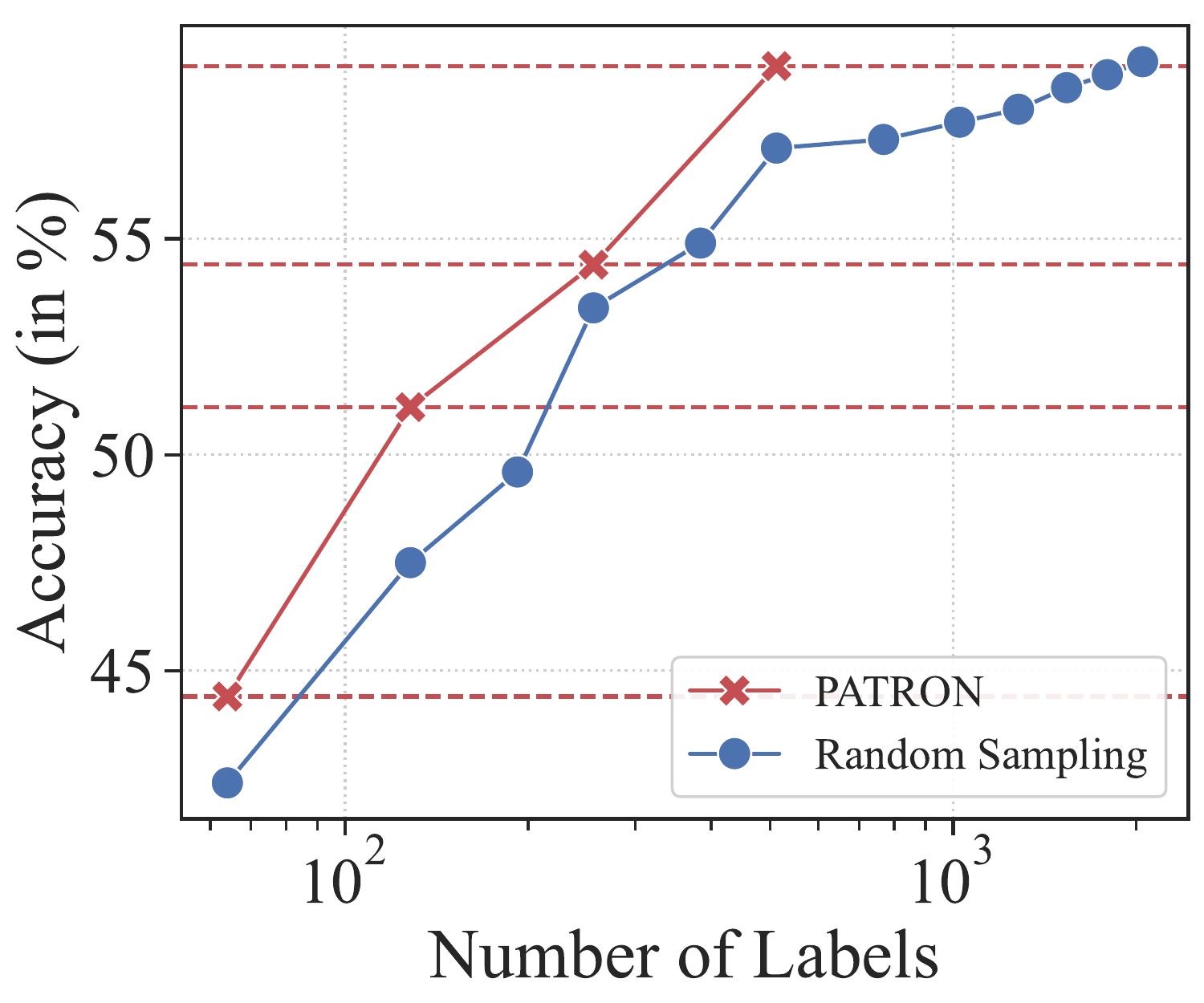}
            \label{fig:log_yelp}
        }
        \hspace{-2mm}
        \subfigure[IMDB]{
            \includegraphics[width=0.32\textwidth]{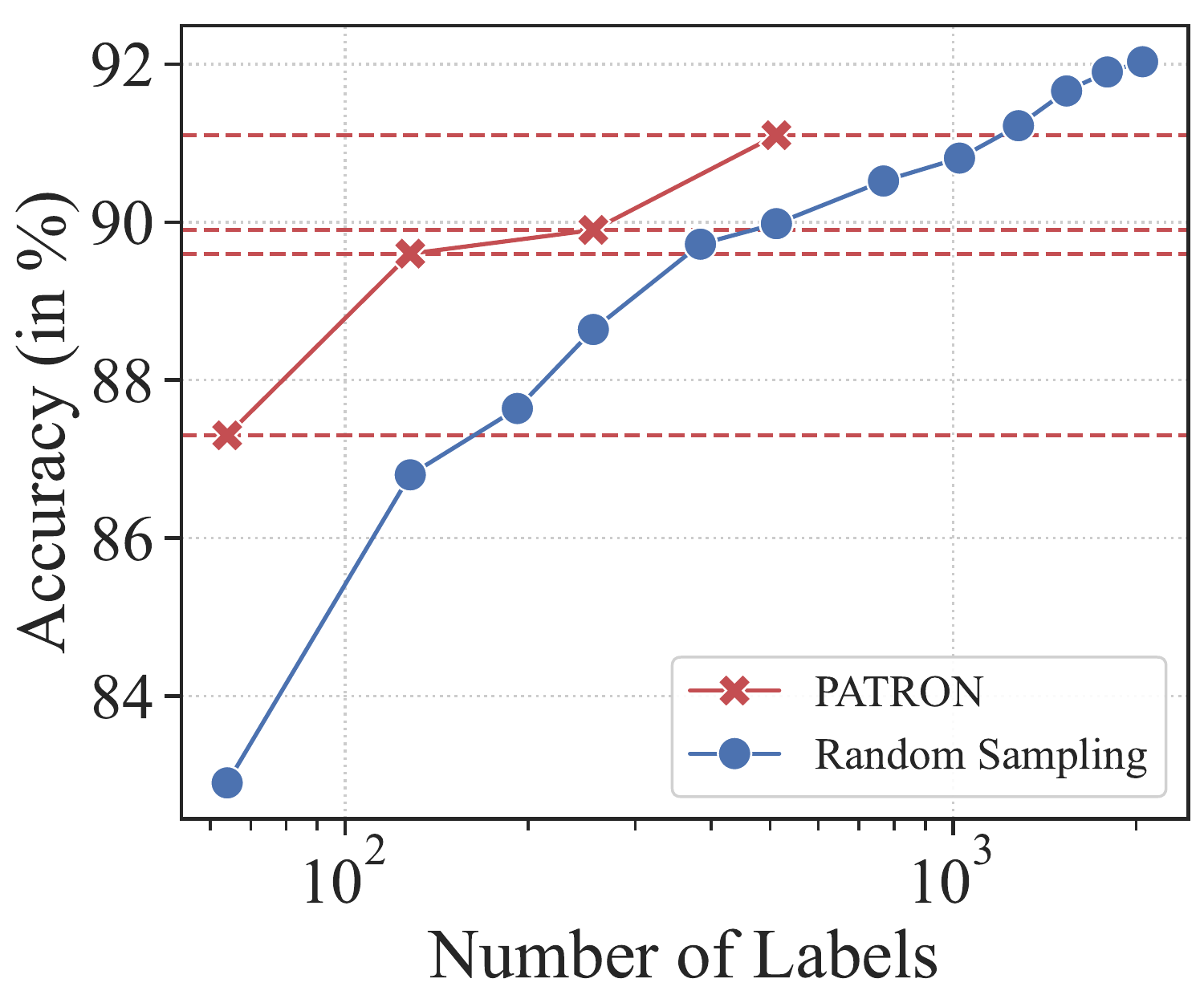}
            \label{fig:log_imdb}
        }
        \hspace{-2mm}
        \subfigure[Yahoo!]{
            \includegraphics[width=0.32\textwidth]{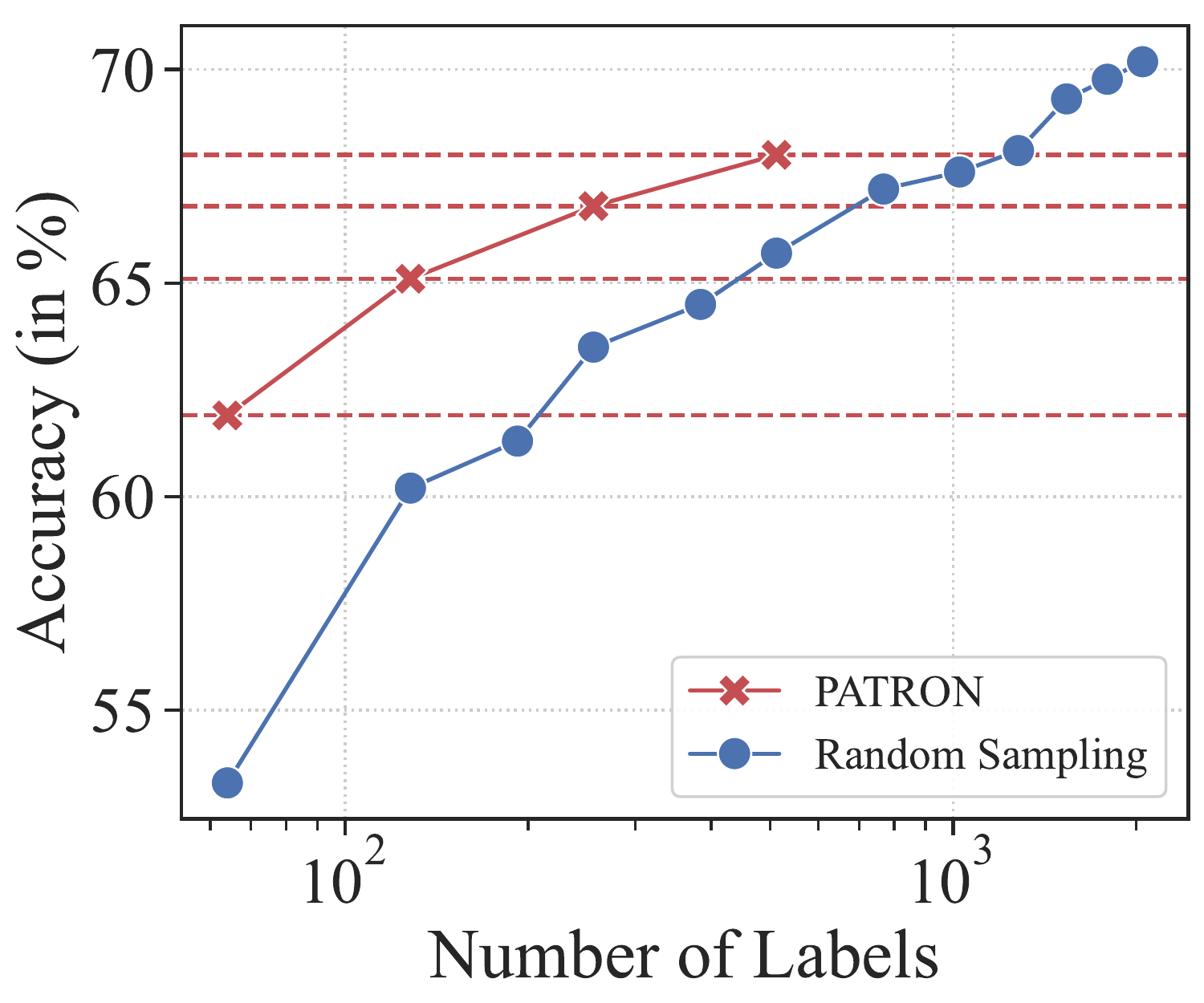}
            \label{fig:pisa_after}
        }
        \hspace{-2mm}
        \subfigure[DBPedia.]{
            \includegraphics[width=0.32\textwidth]{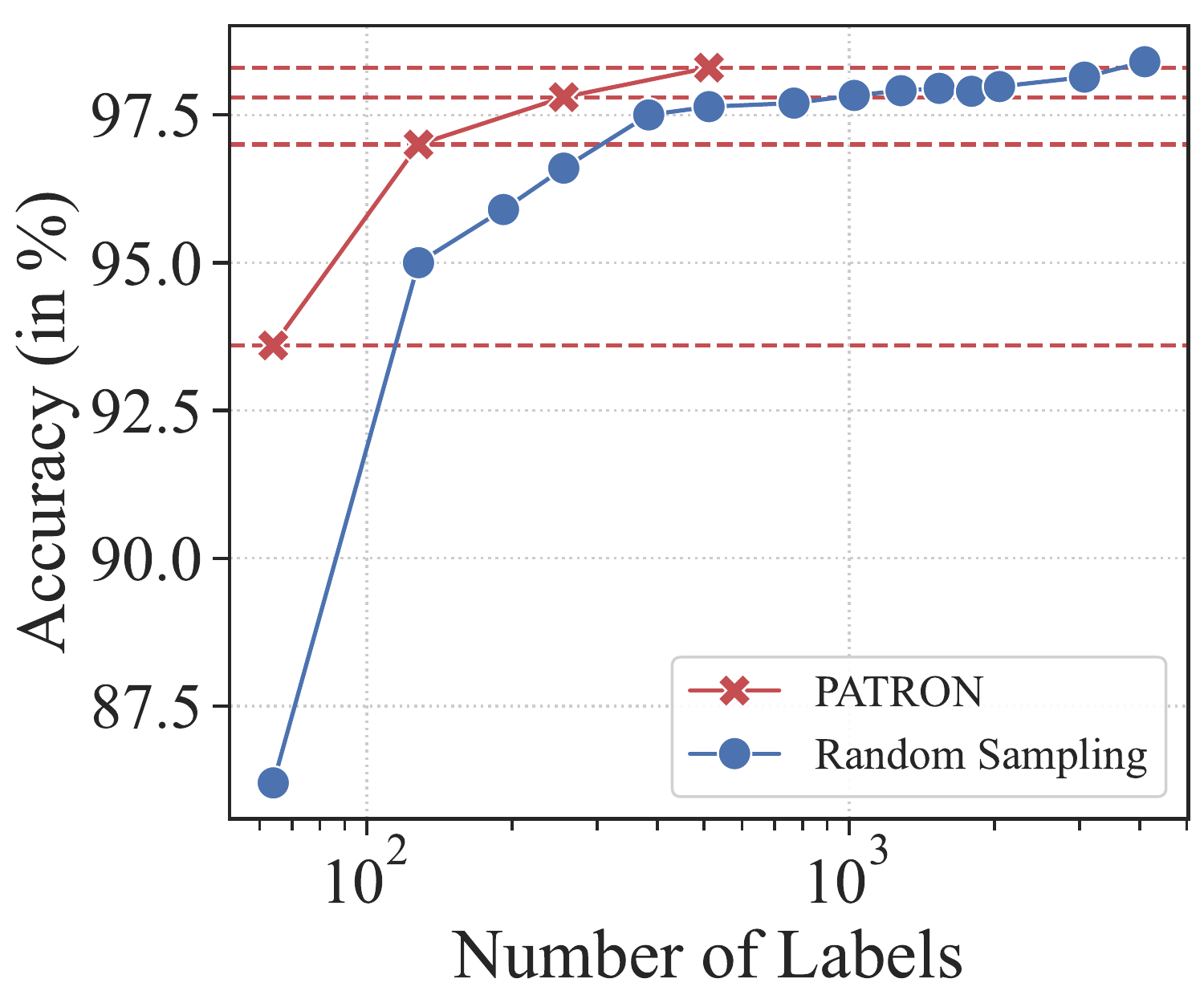}
            \label{fig:log_dbpedia}
        }
        \hspace{-2mm}
        \subfigure[TREC.]{
            \includegraphics[width=0.32\textwidth]{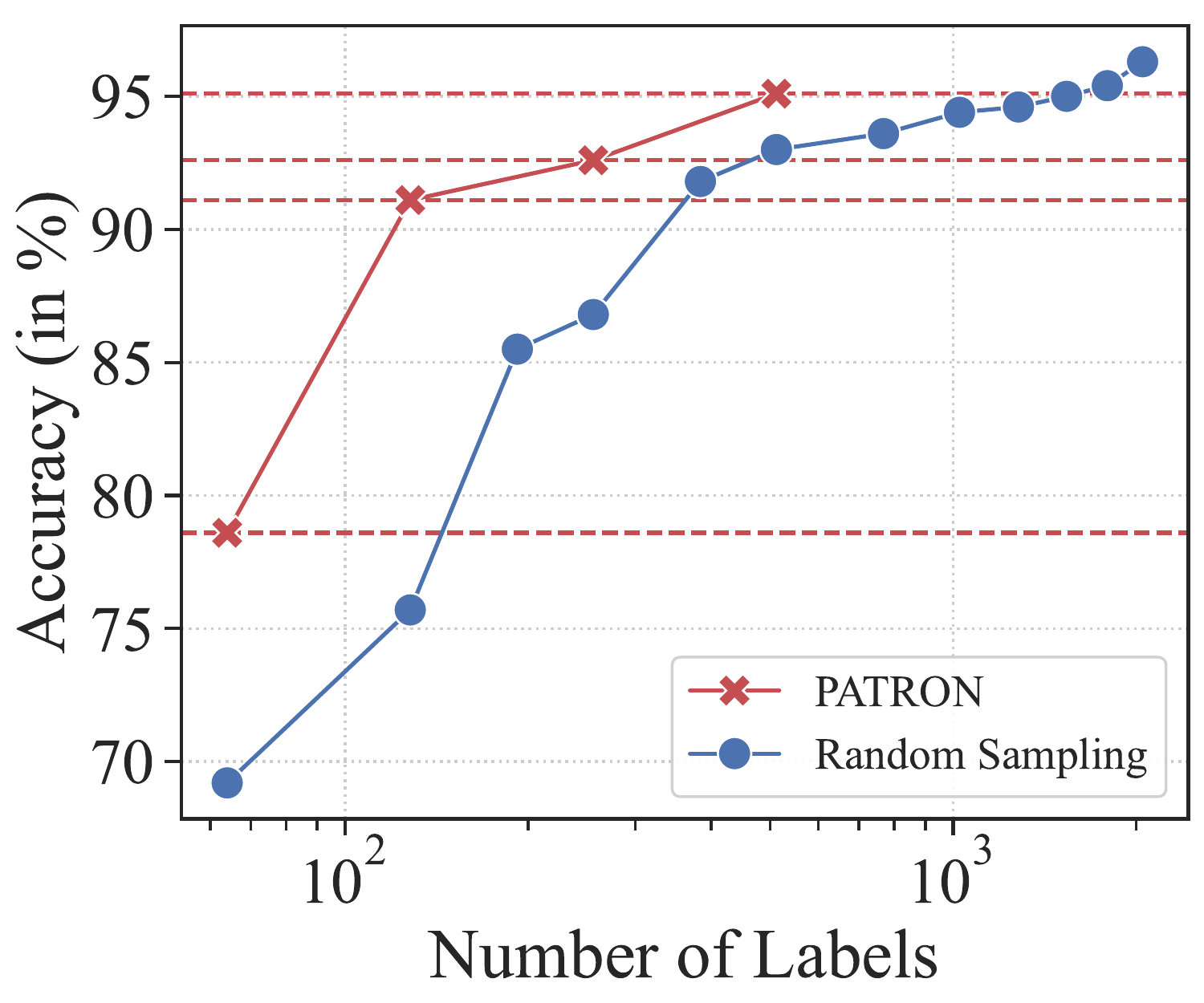}
            \label{fig:log_trec}
        }
        \caption{Illustration of label efficiency on six datasets.}
        \label{fig:efficiency_app}
\end{figure*}

\subsection{Additional Hyperparameter Study}
\label{app:param}
We exhibit the additional hyperparameter study on the other four datasets in Figure \ref{fig:hyper_app}.
Overall, the performance of {\ours} is stable across a broad range of  hyperparameters on all datasets.

\subsection{Additional Label Efficiency Study}
We provide the label efficiency studies for each dataset in detail, shown in Figure \ref{fig:efficiency_app}. 
From the figure, we estimate the approximate number of labels required (via random sampling) to achieve the same performance as {\ours} with 512 labels (Figure~\ref{fig:multi_round_al}) as follows: Yahoo: 1280 (2.5X), TREC: 1024 (2X), AG News: 1536 (3X), IMDB: 1024 (2X), DBPedia: 2304 (4.5X),  Yelp: 1792 (3.5X).
The results indicate that {\ours} can improve the label efficiency for all datasets significantly.

\begin{figure*}[t!]
    \centering
         \subfigure[{\ours} before {\ptr}.]{
            \includegraphics[width=0.48\textwidth]{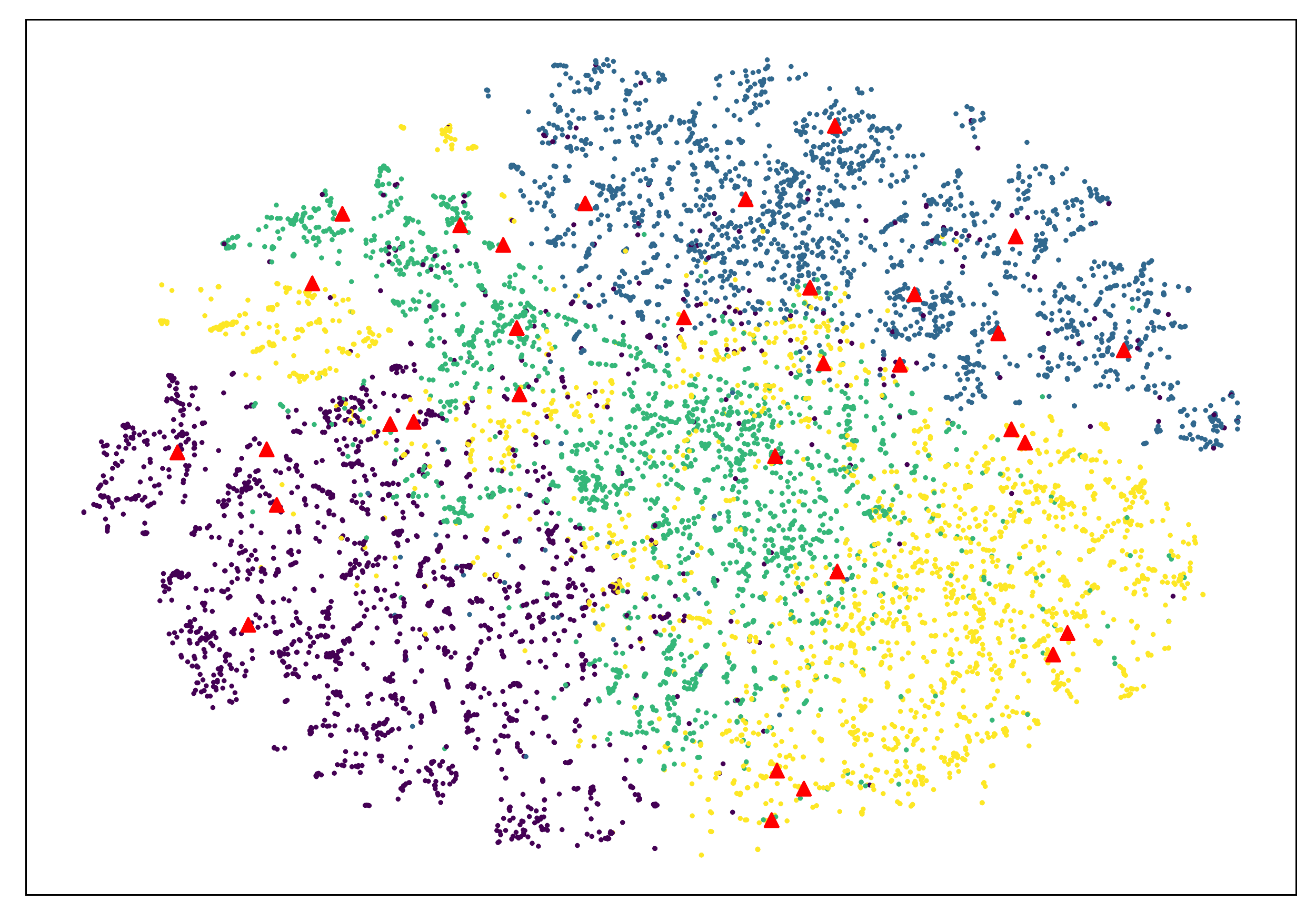}
            \label{fig:pisa_before}
        }\hspace{-2mm}
        \subfigure[{\ours} after {\ptr}.]{
            \includegraphics[width=0.48\textwidth]{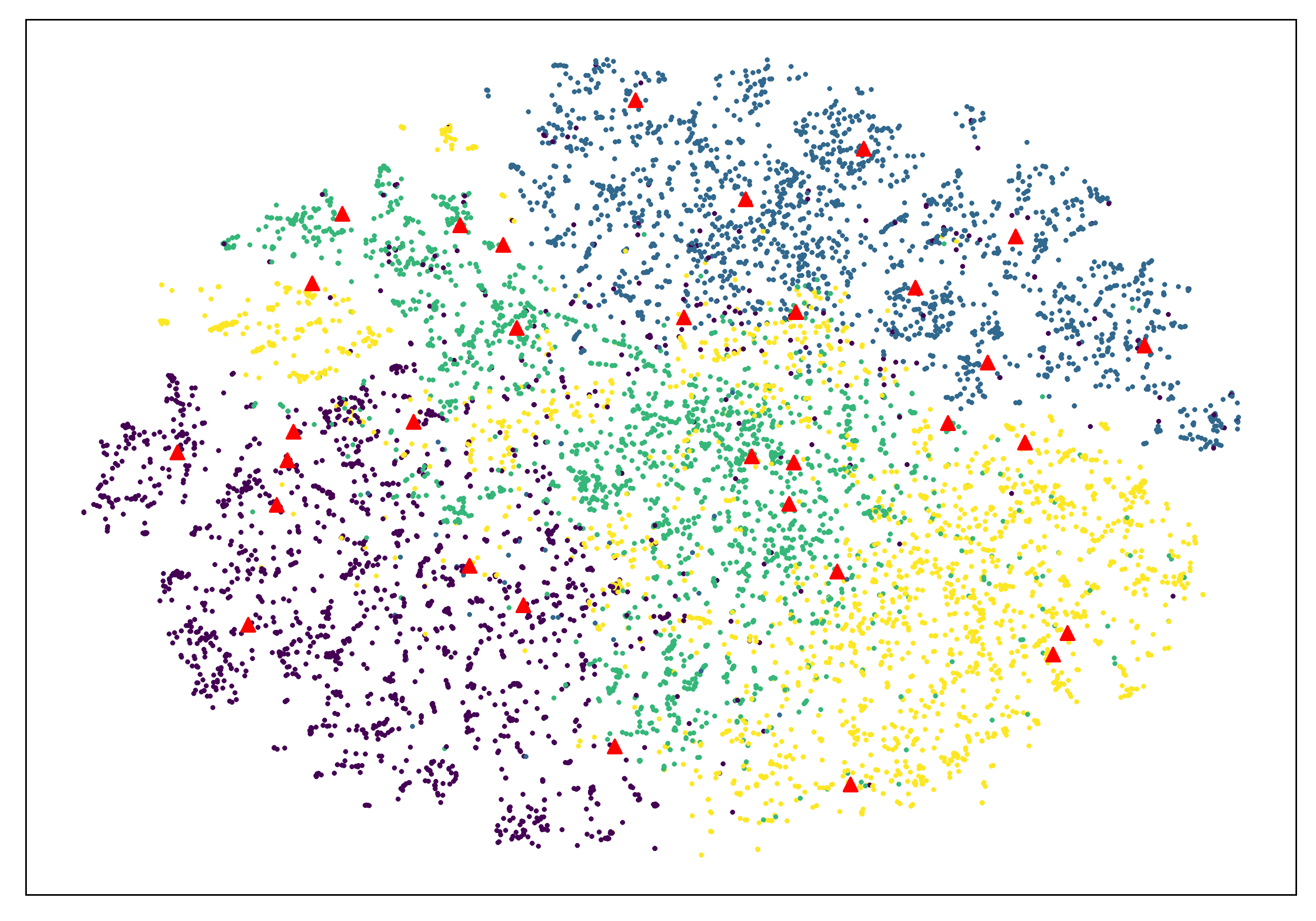}
            \label{fig:pisa_after}
        }
        \vspace{-1ex}
        \caption{Illustration of {\ours} on AG News Dataset. Different colors stands for different classes. Our selected samples are denoted as red triangles.}
        \label{fig:case_study}
\end{figure*}
\section{Case Study}

Figure~\ref{fig:case_study} gives an example on the selected samples of {\ours} on AG News dataset. 
We can see that the initialized solution after Eq.~\ref{eq:init} still
suffers from the issue of limited coverage, and some of the samples are very close. 
Fortunately, after  the {\ptr} step, the diversity of selected samples is much improved. This result suggests the {\ptr} has successfully fulfilled its purpose for diversity-promoting selection.


\end{document}